%% file: main.tex
\documentclass{article}

\PassOptionsToPackage{round, authoryear}{natbib}

 \usepackage[main, final]{neurips_2026}

\usepackage{macros}

\theoremstyle{plain}
\newtheorem{theorem}{Theorem}[section]
\newtheorem{proposition}[theorem]{Proposition}
\newtheorem{lemma}[theorem]{Lemma}
\newtheorem{corollary}[theorem]{Corollary}
\theoremstyle{definition}
\newtheorem{definition}[theorem]{Definition}
\newtheorem{assumption}[theorem]{Assumption}
\crefname{assumption}{assumption}{assumptions}
\Crefname{assumption}{Assumption}{Assumptions}
\theoremstyle{remark}

\title{Differentiable Exact Learning}

\author{%
  Hristo Papazov\thanks{Corresponding author:
    \texttt{hristo.papazov@epfl.ch}} \qquad Francesco D'Angelo \qquad Nicolas Flammarion \\[7.5pt]
    Department of Computer Science, EPFL
}

\begin{document}

\maketitle

\begin{abstract}
We introduce a framework for differentiable exact learning that couples an unbounded external environment to a state-tracking neural controller, which we train on policy-trajectory observations (PTOs): expert behavior traces that expose observations and actions while hiding internal states. In addition to the learning framework, we introduce Differentiable Finite-State Transducers (DFSTs) which use multilinear updates, support log-parallel training, and emulate arbitrary finite-state controllers, yielding computational universality through environmental interaction. From tiny training datasets, compact DFSTs achieve extreme length generalization on arithmetic and universal-simulation tasks. We replicate the success of DFSTs on RNNs and GRUs, thereby supporting a broader learning recipe: \textbf{State-Tracking Controller + Unbounded External Environment + PTOs = Extreme Length Generalization}. Beyond length generalization, our metric-extraction protocol recovers interpretable discrete controllers from trained DFSTs, which we prove equivalent to the target arithmetic and universal-simulation algorithms. Furthermore, for our Softmax DFST models, simply taking $\argmax$ over the transition probabilities yields discrete controllers which provably exactly perform universal Turing-machine simulation and binary arithmetic. Proving exact learning for the neural controllers requires assumptions, which we empirically support. We make our code, training recipes, model weights, and proof checkers \href{https://github.com/fanagor/DEL}{publicly  available}.
\end{abstract}

\begin{figure}[!h]
\centering
\includegraphics[width=0.95\textwidth]{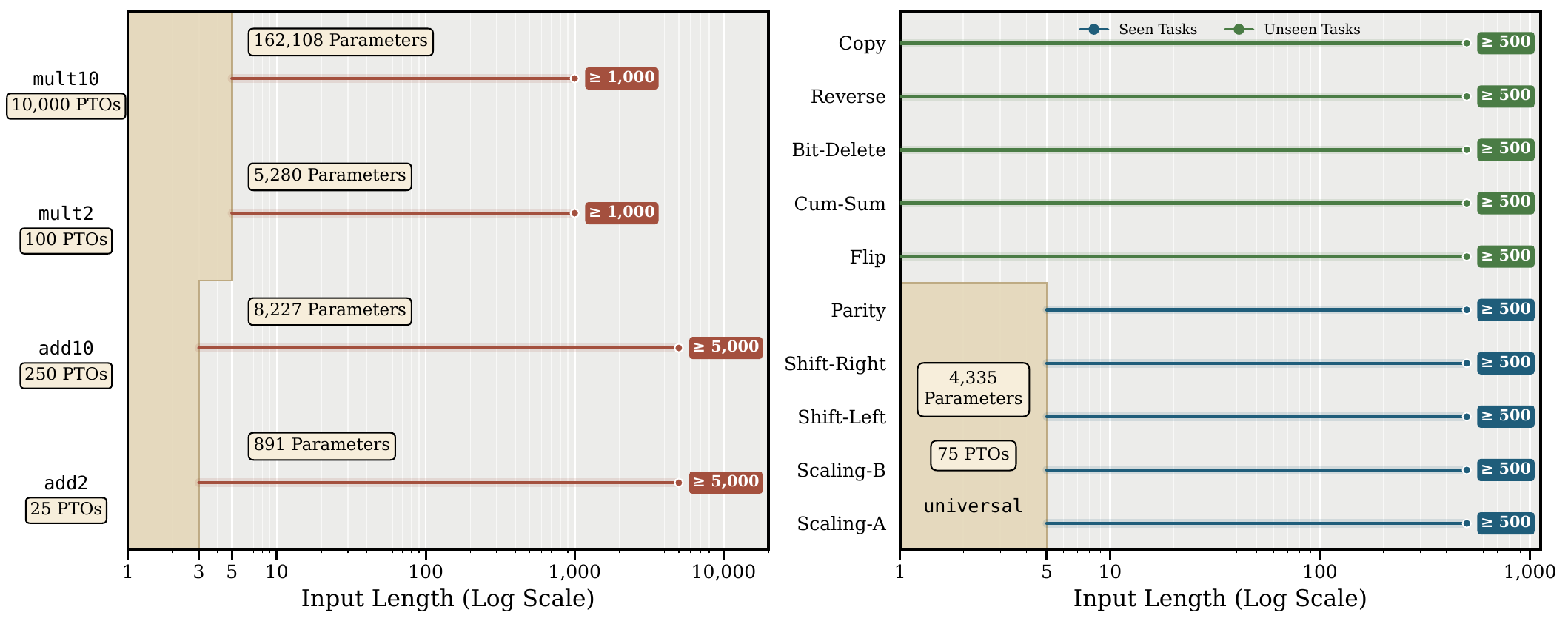}
\caption{\textbf{Extreme Length Generalization from Small PTO Datasets and Compact Controllers.} Length generalization by DFST models trained from scratch on arithmetic and universal-simulation tasks. Shading marks the length of inputs seen during training. Our models see computation traces of inputs with $\leq 3$ digits for addition and $\leq 5$ digits for multiplication. On the universal task, DFSTs see simulation traces of 5 Turing machines on inputs of up-to 5 symbols and generalize to 5 unseen Turing machines. Endpoints mark successful length-generalization reaching a pre-defined testing cap.}
\label{fig:train-vs-gen}
\end{figure}
\clearpage


\input{Sections/Introduction}
\input{Sections/Framework}
\input{Sections/Experiments}
\input{Sections/Exact_Learning}
\input{Sections/Conclusion}

\begin{ack}
    This work was partially funded by an unrestricted gift from Coefficient Giving, and the grant number 212111 from the Swiss National Science Foundation. Francesco D'Angelo was supported by the Google PhD Fellowship.
\end{ack}

\bibliography{references}


\newpage
\appendix
\section*{Organization of the Appendix}
\begin{itemize}[topsep=3pt, itemsep=2pt, left=0pt]
    \item \textbf{\Cref{app:related}} positions our work within the broader literature on neural program induction, arithmetic length generalization, and learning-theoretic supervision hierarchies.
    \item \textbf{\Cref{app:dfsts}} proves the universality theorem for DFSTs (\Cref{thm-universal}) and discusses non-uniqueness of DFST parametrizations.
    \item \textbf{\Cref{app:grid-agents}} collects the extended notation for grid-agent execution, the hardness results motivating our learning paradigm, the pedagogical interpretation of PTOs, and the full catalog of expert agents.
    \item \textbf{\Cref{app:experimental-details}} details optimizer settings, the learning-rate schedule, per-task DFST recipes, the recurrent-controller comparisons, and the capped evaluation protocol.
    \item \textbf{\Cref{app:rnn-sweeps}} presents additional uniform and robust generalization results, Softmax DFST sweeps, and training dynamics for DFST, RNN, and GRU controllers on binary arithmetic.
    \item \textbf{\Cref{app:universal}} describes the teleportation rune mechanics, the Turing-machine encoding, and the 10 simulated tasks used in the universal-agent experiment.
    \item \textbf{\Cref{app:extraction}} develops metric extraction and softmax annealing, and separates empirical evidence from the assumptions needed for neural-model equivalence.
    \item \textbf{\Cref{app:exact}} proves the verification obstruction and conditional faithfulness theorem, then presents eight mathematical equivalence certificates grouped by task, including the required transition maps.
    \item \textbf{\Cref{app:spurious}} constructs small arithmetic controllers that match every short-input PTO yet fail beyond the training lengths, and relates these counterexamples to the Softmax DFST capacity sweeps.
\end{itemize}
\input{Appendix/Related_Work}
\input{Appendix/DFSTs}
\input{Appendix/Grid_Agents_PTOs}
\input{Appendix/Experimental_Details}
\input{Appendix/additional_experiments}
\clearpage
\input{Appendix/Universal}
\clearpage
\input{Appendix/Extraction}
\clearpage
\input{Appendix/Proving_Exact_Learning}
\clearpage
\input{Appendix/Spurious_Controllers}


\newpage
\input{checklist.tex}

\end{document}

%% file: Sections/Introduction.tex
\section{Introduction}\label{sec:intro}

For decades, researchers have sought to make neural networks learn algorithmic procedures from data, arguing that genuine machine intelligence requires exact recovery of underlying rules rather than matching surface-level statistics \citep{solomonoff2006machine, gyorgy2025beyond}.

Yet arithmetic, the canonical testbed for algorithmic learning, continues to expose a persistent gap. Transformers and RNNs trained on input-output data generalize only marginally to longer inputs \citep{deletang2022neural}. Even an extensive arsenal of scratchpad protocols, positional-encoding schemes, and input-formatting tricks brings Transformer length generalization\footnote{Defined in the referenced papers as achieving above 95\% exact-match accuracy (EMA).} to $2$--$7 \times$ the training operand length, with exact-match accuracy swiftly decaying as inputs lengthen \citep{hou2024universal, zhou2024transformers, cho2024position}. The Neural GPU \citep{kaiser2015neural} stands out by reaching $100 \times$ length generalization on random binary arithmetic instances, although later diagnostics expose long-carry and symmetric-input failures \citep{price2016extensions} (see \Cref{tab:comparison}). Such finite-test success leaves exactness on arbitrarily long inputs unresolved.

A relevant exception comes from the recursive Neural Programmer-Interpreter (NPI) \citep{reed2015neural, cai2017making}, which provably recovers decimal addition and several sorting algorithms. However, the training uses a stronger form of supervision through recursive call/return annotations that goes beyond merely observing expert actions in an external environment.

In contrast, real learners routinely succeed under exactly this restriction. Indeed, children acquire complex procedural skills by observing adults perform goal-directed actions, with no access to the adults' internal neural processes. From such behavior traces alone, infants infer the underlying procedural structure and reproduce the target behavior through practice \citep{meltzoff1995infants}.

Motivated by the success of natural learners, this paper entertains the idea of gradient-based exact algorithmic learning from behavior observations and provides a simple recipe for achieving extreme length generalization with compact controllers trained from scratch on small datasets (see \Cref{fig:train-vs-gen}).
To our knowledge, this work provides the first differentiable exact recovery of a fixed \emph{universal interpreter} with correctness proofs covering arbitrary simulations of Turing machines with unbounded input lengths and unbounded execution horizons. We similarly achieve exact differentiable recovery on addition and multiplication tasks. Our training withholds expert control states and the learned interpreter requires no retraining when exchanging input programs.
We just need two ingredients: sufficiently informative observations and a sufficiently expressive learner.

\myparagraph{Sufficiently Informative Observations.} The choice of training data determines the possibility of exact learning. \citet{gold1967language} proved that no algorithm can identify the class of recursive functions in the limit from input–output examples alone \citep[Theorem I.5]{gold1967language}. Moreover, the impossibility persists in the active-learning setting \citep[Theorem I.3]{gold1967language}. Hence, supervised learning from $(z,f(z))$ pairs and reinforcement learning rewarded only for correct output without a helpful curriculum cannot identify every recursive function in general. To navigate between this general identification barrier and the additional (strong and unrealistic) program-structure supervision of NPIs, we adopt policy-trajectory observations (PTOs) as our training signal. PTOs record how the environment evolves under the expert's policy without exposing the expert's internal state transitions.\footnote{Recovering an algorithm with PTOs mirrors more realistic learning-from-demonstration settings \citep{argall2009survey}.} This middle ground carries two advantages: (i) PTOs avoid Gold's impossibility barrier \citep[Theorem 16]{papazov2025learning}; (ii) PTOs admit a natural interpretation as behavior observations -- the learner only observes how the world changes under the expert's actions.

\myparagraph{Sufficiently Expressive Learner.} Since we target exact learning of arbitrary algorithms, the learner must belong to a Turing-complete model class. Prior work establishes Turing-completeness for Transformers and RNNs through assumptions that preclude effective optimization: unbounded arithmetic precision \citep{siegelmann1992computational, perez2018on}, unbounded context windows \citep{dehghani2018universal, bhattamishra2020computational, merrill2023expresssive}, or non-differentiable environmental interaction \citep{chung2021turing, schuurmans2024autoregressive}. We propose a simpler workaround that preserves finite precision and fixed-size contexts, allowing for effective gradient-based optimization. The key idea couples a state-tracking\footnote{We define state-tracking as capable of simulating arbitrary finite-state automata given sufficient model capacity.} controller -- such as an RNN, LSTM, or GRU -- to a dynamically resizing external environment in which the controller can move, read, and write. Because the environment grows as needed, computation no longer bottlenecks on the controller's fixed internal memory. \citet{zaremba2015reinforcement} explored a related architecture pairing an LSTM controller with external tapes, but trained via reinforcement learning on input–output rewards alone, which Gold's theorem renders insufficient for exact learning in general. We hope to revive and strengthen this architectural idea through the richer PTO supervision.

Thus, our learning recipe distills into a simple equation:
\begin{center}
\fbox{\resizebox{0.95\textwidth}{!}{
  State-Tracking Controller + Unbounded External Environment + PTOs = Extreme Length Generalization.
}}
\end{center}
In fact, we go a step further and introduce a simple extraction procedure, which recovers discrete controllers from our trained neural models. We prove that the extracted controllers exactly follow the target behavior and hence certify the following method for differentiable exact learning of algorithms:

\noindent\begin{minipage}{\textwidth}
\centering
\input{Figures/learning_pipeline}
\captionsetup{hypcap=false}
\label{fig:learning-pipeline}
\end{minipage}
\par

\myparagraph{Main Contributions.}

\begin{enumerate}[topsep=4pt, left=0pt, itemsep=3pt, labelsep=5pt]

\item \textbf{A Conceptual Framework for Differentiable Exact Learning.} We revive the idea of achieving Turing-complete differentiable computation through a state-tracking neural controller coupled to an unbounded external environment. Training on a small number of PTOs yields extreme length generalization. Subsequent extraction establishes exact learning for the recovered discrete controllers across the studied tasks and for the neural controllers subject to stability assumptions.

\item \textbf{Differentiable Finite-State Transducers (DFSTs).} We introduce a (i) minimalist, (ii) multilinear, (iii) differentiable, (iv) state-tracking, and (v) parallelizable model family whose members serve as controllers in our experiments. DFSTs use no nonlinear activations, admit log-parallel training via prefix scans \citep{blelloch1990prefix}, and offer a transparent structure for studying the loss landscape of algorithmic learning and the useful implicit bias of gradient descent.

\item \textbf{Extreme Generalization from Small PTO Datasets.} We train on small PTO datasets (25 traces for \texttt{add2}, 100 for \texttt{mult2}, 250 for \texttt{add10}, and 10000 for \texttt{mult10}) and reach robust testing caps of 5000 digits for addition and 1000 digits for multiplication. The training operands contain at most 3 and 5 digits, respectively. We replicate the extreme length generalization on binary arithmetic using RNN and GRU controllers thereby supporting our generalization recipe.

\item \textbf{Exact Learning of a Universal Controller.} A 4335-parameter DFST trained from scratch on just 75 PTOs from 5 program simulations over inputs of at most 5 symbols reaches the 500-symbol evaluation cap across 10 tasks, including 5 held-out programs. From trained DFST models, our extraction procedures recover controllers provably equivalent to the expert universal interpreter and capable of simulating arbitrary programs on any input lengths and execution horizons. 

\item \textbf{Extraction and Verification of Exact Learning.} A lightweight extraction procedure applied to trained DFSTs recovers compact FSTs. Eight proofs establish exact expert-policy equivalence across binary and decimal arithmetic and universal simulation. Additional empirically verified assumptions prove the equivalence of the trained neural controllers to the experts.

\end{enumerate}

\myparagraph{Remark.} An earlier workshop version introduced DFSTs and reported considerable arithmetic length generalization \citep{papazov2025exact}. The present work simplifies training; pushes length generalization to the extreme; and adds learned universal simulation, Softmax DFSTs, recurrent-controller comparisons, and exact-equivalence certificates for extracted controllers.

\myparagraph{Paper Organization.} \Cref{sec:framework} formalizes PTO-learning, DFSTs, and the studied algorithmic tasks. \Cref{sec:exp} expands on our arithmetic and universal experiments, with a short passage on the implicit bias of gradient descent. \Cref{sec:exact-learning} explains FST extraction and verification of correctness. We refer the reader to \Cref{app:related} for a thorough discussion of related work.

%% file: Figures/learning_pipeline.tex
\begingroup
\definecolor{PipelineBlue}{HTML}{226780}
\definecolor{PipelineTerra}{HTML}{AF5039}
\definecolor{PipelineGreen}{HTML}{4D8046}
\definecolor{PipelineGray}{HTML}{F0F0EE}
\begin{tikzpicture}[x=1pt,y=1pt,
    every node/.style={inner sep=0pt,font=\fontsize{8}{9}\selectfont},
    flow/.style={-{Stealth[length=3pt,width=3pt]},line width=.8pt}]
  \foreach \left/\right in {0/123,137/260,274/397}
    \fill[PipelineGray,rounded corners=4pt]
      (\left,-14) rectangle (\right,74);
  \node[text=PipelineBlue,font=\small\bfseries] at (61.5,63)
    {Observe PTOs};
  \node[text=PipelineTerra,font=\small\bfseries] at (198.5,63)
    {Learn Differentiably};
  \node[text=PipelineGreen,font=\small\bfseries] at (335.5,63)
    {Extract};

  \foreach \i/\symbol in {0/1,1/0,2/1,3/+,4/1,5/1} {
    \draw[fill=white,draw=black!20]
      ({15+16*\i},34) rectangle ({29+16*\i},49);
    \node[text=PipelineBlue,font=\small]
      at ({22+16*\i},41.5) {\symbol};
  }
  \draw[PipelineBlue,line width=1pt] (15,34) rectangle (29,49);
  \node at (61.5,15) {Read, Write, Move};
  \node[text=PipelineBlue] at (61.5,-7) {Expert States Hidden};

  \node at (198.5,49) {Read};
  \draw[flow,PipelineTerra] (198.5,44) -- (198.5,39);
  \node[draw=PipelineTerra,fill=white,rounded corners=3pt,
        line width=.8pt,minimum width=103pt,minimum height=23pt,
        align=center,text=PipelineTerra] at (198.5,27)
        {State-Tracking\\Neural Controller};
  \draw[flow,PipelineTerra] (198.5,15) -- (198.5,11);
  \node at (198.5,6) {Predict Write and Move};
  \node[text=PipelineTerra] at (198.5,-7)
    {Extreme Length Generalization};

  \foreach \x in {302,335.5,369}
    \draw[PipelineGreen,fill=white,line width=1pt]
      (\x,37) circle (7pt);
  \draw[flow,PipelineGreen] (310,37) -- (327.5,37);
  \draw[flow,PipelineGreen] (343.5,37) -- (361,37);
  \draw[flow,PipelineGreen] (331,43)
    .. controls (325,56) and (346,56) .. (340,43);
  \node at (335.5,15) {Discrete Controller};
  \node[text=PipelineGreen] at (335.5,-7)
    {Provably Equivalent to Expert};

  \draw[flow] (125,37) -- (135,37);
  \draw[flow] (262,37) -- (272,37);
\end{tikzpicture}%
\endgroup

%% file: Sections/Framework.tex
\section{Preliminaries} \label{sec:framework}

As discussed in the introduction, our recipe for exact learning couples a state-tracking controller to an unbounded external environment.

\subsection{Policy-Trajectory Observations}\label{subsec:ptos}

We model the external environment as a symbolic grid.

\begin{definition}[Symbolic Grid]\label{def:grid}
Given a finite alphabet $\Sigma$ containing the empty symbol $\l$, a \emph{grid configuration} $w : \Z^2 \to \Sigma$ assigns a symbol $w(p) \in \Sigma$ to each cell $p \in \Z^2$ such that only finitely many cells differ from $\l$. The symbolic grid $\mathcal{G} = \mathcal{G}(\Z^2, \Sigma)$ denotes the set of all such configurations.
\end{definition}

The finiteness requirement captures the idea that the environment grows as needed but never stores an infinite amount of information at any point in time. We now formalize the agents that inhabit the grid.

\begin{definition}[Grid Agent]\label{def:agent}
A \emph{grid agent} over the symbolic grid $\mathcal{G}(\Z^2, \Sigma)$ stands for a finite-state transducer $T = (Q, \Sigma, \Gamma, \d, \g, q_0, q_f)$, where $Q$ denotes a finite set of states, $\Sigma$ serves as both the read alphabet and the write alphabet, $\Gamma = \Sigma \times \M$ with $\M = \{\texttt{L}, \texttt{R}, \texttt{U}, \texttt{D}, \texttt{S}\}$ forms the output alphabet pairing a grid symbol with a motion direction, $\d : Q \times \Sigma \to Q$ specifies the state-transition function, $\g : Q \times \Sigma \to \Gamma$ specifies the output function, $q_0 \in Q$ denotes the initial state, and $q_f \in Q$ denotes the halting state. The agent always starts at $p_0=(0,0)$.
\end{definition}

\begin{figure*}[t]
    \centering
    \input{Figures/dfst_setup}
    \caption{\textbf{One Step of the DFST Agent.}
    \emph{Left:} the grid $w_{t-1}$ before the action; the agent at position $p_{t-1}$ reads the symbol in the \textcolor[RGB]{190,45,55}{red} cell as observation $x_t$.
    \emph{Center:} the \textsc{DFST}$_\theta$ controller updates the hidden state via $h_t\!=\!A[x_t]\,h_{t-1}$ and produces the discrete outputs $\sigma_t\!=\!\arg\max(B[x_t]\,h_{t-1})\!\in\!\Sigma$ (\textcolor[RGB]{210,115,15}{write}) and $\mu_t\!=\!\arg\max(C[x_t]\,h_{t-1})\!\in\!\M$ (\textcolor[RGB]{20,140,120}{move}).
    \emph{Right:} the resulting grid $w_t$; the \textcolor[RGB]{210,115,15}{orange} cell shows the symbol $\sigma_t$ written at $p_{t-1}$, and the agent has shifted to $p_t$ in the \textcolor[RGB]{20,140,120}{teal} cell. We provide an \href{https://github.com/fanagor/DEL\#interactive-tape-and-grid-viewer}{interactive viewer} that animates expert and trained neural controllers.}
    \label{fig:setup}
\end{figure*}

The agent occupies a single cell $p_t \in \Z^2$ at each time step $t$, reads the symbol $x_{t+1} = w_{t}(p_{t})$, produces an output $\g(q_t, x_{t+1}) = (s_{t+1}, m_{t+1})$ that writes $s_{t+1}$ at position $p_t$ and shifts the agent's position by $m_{t+1}$, and transitions to state $\d(q_t, x_{t+1})$. We adopt the convention that every transition containing the stop motion $\texttt{S}$ leads to the halting state $q_f$, and all transitions from $q_f$ satisfy $\d(q_f,x)=q_f$ and $\g(q_f,x)=(x,\texttt{S})$ for every $x\in\Sigma$, rendering $q_f$ a dead state that leaves the grid unchanged.

Note that a grid agent perceives only one cell at a time. Future work could explore richer perception mechanisms encoding the contents of multiple cells and agents interacting with a larger motion vocabulary. Since standard Turing machines arise as a special case of grid agents restricted to a single row, the class $\A$ of all grid agents achieves computational universality when interacting with $\G$.

\myparagraph{PTOs.}
At each time step $t$, an external learner observing agent $T$ perceives only the written symbol $s_t \in \Sigma$, the motion $m_t \in \M$, and the observation $x_t$ that triggered the action. Importantly, the internal state $q_t$ remains hidden. The resulting sequence $(x_t, s_t, m_t)_{t=1}^T$ constitutes a \emph{policy-trajectory observation} (PTO), matching the formalism of \citet{papazov2025learning}. Crucially, \citet[Theorem~16]{papazov2025learning} proved that PTOs enable exact identification of any computable function in the limit -- a guarantee unachievable through input--output supervision alone \citep{gold1967language}.

\myparagraph{Hardness.}
Each algorithmic task $\mathcal{P}$ comes with a set of valid inputs $S_\mathcal{P} \subseteq \SS$. Because the motion token outputted at each step constrains the next observation, $S_\mathcal{P}$ may never cover all reachable transducer transitions, thereby making the learning problem fundamentally hard. \citet[Theorem~17]{papazov2025learning} proved that no \textit{polynomial-mass} characteristic set exists for this restricted-input regime, ruling out any algorithm that learns the policy of every grid agent from polynomially many polynomially long PTOs (see \Cref{app:hardness} for details).


\subsection{State-Tracking Differentiable Controllers}

While \citet[Theorem~16]{papazov2025learning} showed that a learning-by-enumeration strategy provably identifies any grid agent from PTOs, such an approach takes time and memory exponential in the size of the target expert agent. We instead pursue gradient-based learning by introducing a simple differentiable model that can serve as a grid-agent controller.

\begin{definition}[DFST] \label{def:dfst}
    A \emph{Differentiable Finite-State Transducer} (DFST) of dimension $d$ over the symbolic grid $\G(\Z^2, \Sigma)$ with motion vocabulary $\M$ comprises three trainable weight tensors
    \begin{equation*}
        A \in \R^{|\Sigma| \times d \times d}, \qquad B \in \R^{|\Sigma| \times |\Sigma| \times d}, \qquad C \in \R^{|\Sigma| \times |\M| \times d},
    \end{equation*}
    and a trainable initial hidden state $h_0 \in \R^d$. At each time step $t \geq 1$, the controller reads the one-hot encoding $x_t \in \R^{|\Sigma|}$ of the symbol at the current grid cell $w_{t-1}(p_{t-1})$ and computes
    \begin{equation*}
        h_t = A[x_t]\, h_{t-1}, \qquad \hat{s}_t = B[x_t]\, h_{t-1}, \qquad \hat{m}_t = C[x_t]\, h_{t-1}.
    \end{equation*}
    $A$ drives the hidden-state recurrence, $B$ produces scores over symbol tokens, and $C$ produces scores over motion tokens. At inference time, the controller emits $\s_t = \argmax(\hat{s}_t) \in \Sigma$ and $\mu_t = \argmax(\hat{m}_t) \in \M$, writing $\s_t$ at position $p_{t-1}$ and moving $\mu_t$ to $p_t$ (see \Cref{fig:setup}).
\end{definition}

The DFST recurrence and readouts use only input-indexed matrix-vector products, without nonlinear activations. This multilinear structure enables log-parallel training via prefix scans \citep{blelloch1990prefix} and yields a compact parametrization. A $d$-dimensional DFST over an alphabet of size $k = |\Sigma|$ with $r = |\M|$ motion tokens carries $kd(k + d + r) + d$ trainable parameters, scaling as $O(d^2)$.

\begin{definition}[Softmax DFST] \label{def:softmax-dfst}
A Softmax DFST has the same recurrence and readout mechanism as a \hyperref[def:dfst]{DFST}, with trainable logits $\Theta^A,\Theta^B,\Theta^C,\theta^h$ parametrizing
\begin{equation*}
 P[x]_{:j}=\operatorname{softmax}(\Theta^P[x]_{:j}),\quad P\in\{A,B,C\},
 \qquad h_0=\operatorname{softmax}(\theta^h).
\end{equation*}
Each column sums to one. Thus, the hidden state remains in the probability simplex and the readouts give distributions over writes and motions. Execution uses the same argmax decoding as a DFST.
\end{definition}

This parametrization makes extraction transparent: each column assigns probabilities to successors or actions, and concentration near a simplex vertex suggests a deterministic transition. The recurrence retains the associative matrix scan. \Cref{sec:exact-learning} develops differentiable sharpening and the conditions for faithful extraction.

Despite the simplicity of these architectures, both families support state tracking:

\begin{restatable}[DFST-Grid Universality]{theorem}{universalitydfst} \label{thm-universal}
    Any $d$-state grid agent admits exact emulation by a $d$-dimensional DFST at any finite arithmetic precision, and by a $(d+1)$-dimensional Softmax DFST with finite logits in exact real arithmetic. A $d$-dimensional Softmax DFST also suffices under rounded arithmetic whose softmax attains simplex vertices and whose recurrence preserves vertex operations.
\end{restatable}

Consequently, both the DFST and Softmax DFST families achieve Turing-completeness on the symbolic grid under the respective arithmetic conditions. \Cref{app:dfsts} gives the constructive proof and discusses non-uniqueness of DFST parametrizations and the relationship between hidden dimension and expert state count.

More broadly, any sequential model capable of emulating a finite-state transducer can represent the policy of a grid agent and thereby compute any computable function on the grid. This class includes RNNs, LSTMs, and GRUs \citep{alon1991efficient, indyk1995optimal, siegelmann1996recurrent, weiss2018practical, merrill2019sequential, svete2023recurrent} whereas standard Transformers and diagonal SSMs face state-tracking limitations as demonstrated by \citet{merrill2023parallelism, merrill2024illusion}. We focus on DFSTs due to the multilinear structure and simplicity of the model class, which we hope will inspire theoretical analysis of the implicit bias of gradient descent toward generalizing solutions. The DFST, RNN, and GRU experiments in \Cref{sec:exp} support this broader state-tracking perspective.

\subsection{Experimental Setup}\label{subsec:setup}

We study grid agents for binary and decimal arithmetic (\texttt{add2}, \texttt{add10}, \texttt{mult2}, \texttt{mult10}) and a universal grid agent $U$ that simulates Turing machines over the $\{\lambda, 0, 1\}$ alphabet. \Cref{tab:agents} summarizes expert state counts, alphabets, and runtimes for all implemented grid agents and Turing machines.

\paragraph{Arithmetic Tasks.}\label{subsec:arithmetic}

Each arithmetic agent receives $z = a \circ b$, with $\circ \in \{+, \times\}$, written horizontally on an otherwise empty grid. The agents execute the schoolbook algorithms for addition and multiplication, halting when only the result remains on the grid.

\myparagraph{Data Generation.} Each arithmetic PTO dataset includes all $R^2$ pairs of single-digit operands, then samples distinct pairs with independent uniform operand lengths in $1,\ldots,q$ and digits in $0,\ldots,R-1$. Leading zeros remain part of the representation. Ordinary DFST datasets always use:
\texttt{add2}: $q = 3$, $N = 25$;\enspace
\texttt{add10}: $q = 3$, $N = 250$;\enspace
\texttt{mult2}: $q = 5$, $N = 100$;\enspace
\texttt{mult10}: $q = 5$, $N = 10\text{k}$.

\myparagraph{Length-Generalization Metrics.} Uniform Length Generalization (ULG) records the operand length returned by a capped search requiring perfect next-action exact-match accuracy on ten random pairs of exactly that length. Robust Length Generalization (RLG) additionally requires perfect accuracy on structured repdigit probes and cannot exceed ULG. Every RLG evaluation tests all $R^2$ repdigit pairs $(a^m,b^m)$ with $a,b\in\{0,\ldots,R-1\}$. Every timestep must match both the expert write and motion. The Neural GPU motivates this distinction: strong generalization on random binary inputs \citep{kaiser2015neural} can coexist with failure on repdigit instances \citep{price2016extensions}. \Cref{app:bp-eval} specifies the search, numerical execution modes, and limits of finite testing.

\paragraph{Learning a Universal Interpreter.}\label{subsec:universal-learning}

The universal grid agent $U$ uses 15 states and alphabet $\{0, 1, \l, X, Y, Z, \mathtt{L}, \mathtt{R}, \mathtt{S}\}$. Given an encoding $\langle M,z\rangle$ of a Turing machine $M$ and input $z\in\bits^+$, $U$ simulates $M$ on $z$, halting with $M(z)$ in a designated grid region whenever $M$ halts.

$U$ uses 3 teleportation runes $\{X,Y,Z\}$ to mark and jump directly between grid locations, shortening simulation trajectories. The grid stores an encoding of the currently simulated Turing machine $M$ together with the input. From PTOs of the universal expert, a differentiable controller learns how to read and execute this program, without access to the internal states or transition map of $U$. The \href{https://github.com/fanagor/DEL}{public repository} provides model weights and animated executions.

\myparagraph{Data Generation and Evaluation.} Universal datasets contain $N\in\{75,150,225\}$ PTOs covering 5 programs, with binary input lengths at most 5. After sampling a random string for each program, the rest of the dataset comes from random program--input pairs. Evaluation covers 10 tasks, including 5 held-out programs and scaling tasks that increase both input length and program size. \Cref{app:universal} details the rune mechanics, program/input encoding, and training and evaluation tasks.

\paragraph{Controller Training.}
For each task, a manually designed expert supplies PTOs. We train controllers via next-action prediction with the MSE loss
\begin{equation*}
\frac{1}{2T}\sum_{t=1}^T \left[\|\hat{s}_t-\mathbf{1}_{s_t}\|_2^2+\|\hat{m}_t-\mathbf{1}_{m_t}\|_2^2\right],
\end{equation*}
where $\mathbf{1}_{s_t}$ and $\mathbf{1}_{m_t}$ denote the one-hot encodings of the expert write and motion at time $t$. We train DFST controllers on all tasks, study Softmax DFSTs on binary arithmetic and universal simulation, and compare RNN and GRU controllers on binary arithmetic. Training always uses Adam \citep{kingma2014adam}, batch size 32, and \texttt{float32} precision.

\myparagraph{DFST Training.} Unless otherwise specified, DFST hidden dimensions match the expert state counts. We always use a simple deterministic initialization. Transition matrices start at the identity, output matrices at zero, and the trainable initial state at $\mathbf{1}_d$. All ordinary DFSTs runs use an exponential warm-up and cool-down learning-rate scheduler(\Cref{app:dfst-init,app:opt-sched}). \Cref{subsec:hidden} examines successful learning with smaller and larger hidden dimensions.

\myparagraph{Softmax DFST Training.} Softmax DFSTs apply the same initialization recipe to the logits before softmax normalization. The sweeps use one, two, or three times the expert state count and vary dataset size. Training uses 100k updates at constant learning rate $0.1$ (and optional annealing towards approximately one-hot transition probabilities). \Cref{app:softmax-training} gives the full training settings.

%% file: Figures/dfst_setup.tex
\begin{tikzpicture}[
    >={Stealth[length=2.0mm, width=1.6mm]},
    font=\small,
    gridcell/.style    = {draw=black!35, line width=0.3pt, minimum size=0.45cm, inner sep=0pt},
    gridsym/.style     = {font=\scriptsize, text=black!75},
    dfstframe/.style   = {draw=accentBlue!70, line width=0.7pt, rounded corners=4pt,
                         fill=accentBlue!4, inner sep=8pt},
    state/.style       = {circle, draw=accentBlue!75!black, line width=0.7pt,
                         fill=accentBlue!15, minimum size=0.72cm, inner sep=0pt,
                         font=\footnotesize},
    tensor/.style      = {rectangle, draw=accentBlue!60!black, line width=0.5pt,
                         rounded corners=2pt, fill=white, inner sep=3pt,
                         font=\scriptsize},
    flow/.style        = {->, line width=0.5pt, shorten >=1pt, shorten <=1pt,
                         draw=black!70},
    lbl/.style         = {font=\scriptsize\itshape, inner sep=1pt},
]

\definecolor{accentBlue}{RGB}{33, 90, 160}
\definecolor{accentRed}{RGB}{190, 45, 55}
\definecolor{colWrite}{RGB}{210, 115, 15}   
\definecolor{colMove}{RGB}{20, 140, 120}    

\begin{scope}[shift={(0,0)}]
    \foreach \x in {0,...,4}{
        \foreach \y in {0,...,4}{
            \node[gridcell] (L\x\y) at (\x*0.45,\y*0.45) {};
        }
    }
    \node[gridsym] at (L12) {$1$};
    \node[gridsym] at (L32) {$1$};
    \node[gridsym] at (L42) {$0$};
    \node[gridsym] at (L11) {$0$};
    \node[gridsym] at (L21) {$1$};
    \node[gridsym] at (L31) {$1$};
    \node[gridsym] at (L13) {};
    \node[gridsym] at (L33) {};

    \node[gridcell, draw=accentRed, line width=1.0pt, fill=accentRed!12]
        (Lagent) at (L22) {};
    \node[gridsym, font=\scriptsize\bfseries, text=accentRed!80!black]
        at (L22) {$1$};

    \node[circle, fill=accentRed!80, draw=accentRed!85, line width=0.35pt,
          minimum size=3pt, inner sep=0pt]
        (Lhead) at ($(L24.north)+(0,0.28)$) {};
    \draw[accentRed!70, line width=0.45pt] (Lagent.north) -- (Lhead.south);

    \node[lbl, text=accentRed!80!black, above=0pt of Lhead] {$p_{t\!-\!1}$};
    \node[font=\footnotesize, anchor=south] at ($(L24.north)+(0,0.50)$)
        {$w_{t-1}$};
\end{scope}

\begin{scope}[shift={(4.3,0.9)}]

    \node[state] (ht)     at (0,0)     {$h_{t\!-\!1}$};
    \node[state] (htnext) at (3.0,0)   {$h_{t}$};

    \node[tensor] (Atensor) at ($(ht)!0.5!(htnext)+(0,0.78)$)
        {$A[{\color{accentRed}x_t}]\,h_{t\!-\!1}$};
    \draw[flow] (ht.north east) to[bend left=18] (Atensor.west);
    \draw[flow] (Atensor.east)  to[bend left=18] (htnext.north west);
    \node[lbl, text=accentBlue!70!black, above=0pt of Atensor] {state update};

    \node[tensor] (Btensor) at (1.0,-1.10)  {$B[{\color{accentRed}x_t}]\,h_{t\!-\!1}$};
    \node[font=\scriptsize, text=black!70, fill=white, inner sep=3pt]
        (argmaxS) at (2.35,-1.10) {argmax};
    \node[rectangle, draw=colWrite!70!black, line width=0.5pt,
          rounded corners=2pt, fill=colWrite!8,
          inner sep=3pt, font=\scriptsize]
        (sigma) at (3.65,-1.10) {$\sigma_t \!\in\! \Sigma$};
    \draw[flow, shorten >=4pt] (ht.south) |- (Btensor.west);
    \draw[line width=0.5pt, draw=black!70, shorten <=2pt, shorten >=1pt]
        (Btensor.east) -- (argmaxS.west);
    \draw[flow, shorten <=1pt] (argmaxS.east) -- (sigma.west);
    \node[lbl, text=colWrite!85!black, below=1pt of sigma] {write};

    \node[tensor] (Ctensor) at (1.0,-2.00)  {$C[{\color{accentRed}x_t}]\,h_{t\!-\!1}$};
    \node[font=\scriptsize, text=black!70, fill=white, inner sep=3pt]
        (argmaxM) at (2.35,-2.0) {argmax};
    \node[rectangle, draw=colMove!70!black, line width=0.5pt,
          rounded corners=2pt, fill=colMove!8,
          inner sep=3pt, font=\scriptsize]
        (mu) at (3.65,-2.00) {$\mu_t \!\in\! \M$};
    \draw[flow, shorten >=4pt] (ht.south) |- (Ctensor.west);
    \draw[line width=0.5pt, draw=black!70, shorten <=2pt, shorten >=1pt]
        (Ctensor.east) -- (argmaxM.west);
    \draw[flow, shorten <=1pt] (argmaxM.east) -- (mu.west);
    \node[lbl, text=colMove!85!black, below=1pt of mu] {move};

    \begin{pgfonlayer}{background}
        \node[dfstframe,
              fit=(ht)(htnext)(Atensor)(Btensor)(Ctensor)(argmaxS)(argmaxM)(sigma)(mu)]
            (dfstbox) {};
    \end{pgfonlayer}
    \node[font=\footnotesize\bfseries, text=accentBlue!80!black,
          anchor=south west, inner sep=2pt]
        at ($(dfstbox.north west)+(0.08,0.04)$) {\textsc{DFST}$_\theta$};
\end{scope}

\begin{scope}[shift={(10.2,0)}]
    \foreach \x in {0,...,4}{
        \foreach \y in {0,...,4}{
            \node[gridcell] (R\x\y) at (\x*0.45,\y*0.45) {};
        }
    }
    \node[gridsym] at (R12) {$1$};
    \node[gridsym] at (R42) {$0$};
    \node[gridsym] at (R11) {$0$};
    \node[gridsym] at (R21) {$1$};
    \node[gridsym] at (R31) {$1$};
    \node[gridsym] at (R13) {};
    \node[gridsym] at (R33) {};

    \node[gridcell, draw=colWrite, line width=1.0pt, fill=colWrite!12]
        (Rwritten) at (R22) {};
    \node[gridsym, font=\scriptsize\bfseries, text=colWrite!80!black]
        at (R22) {$\sigma_t$};

    \node[gridcell, draw=colMove, line width=1.0pt, fill=colMove!10]
        (Rtarget) at (R32) {};
    \node[gridsym, text=colMove!75!black] at (R32) {$1$};

    \node[circle, fill=colMove!70, draw=colMove!75, line width=0.35pt,
          minimum size=3pt, inner sep=0pt]
        (Rhead) at ($(R34.north)+(0,0.28)$) {};
    \draw[colMove!65, line width=0.45pt] (Rtarget.north) -- (Rhead.south);

    \node[lbl, text=colMove!75!black, above=0pt of Rhead] {$p_{t}$};
    \node[font=\footnotesize, anchor=south] at ($(R24.north)+(0,0.50)$)
        {$w_{t}$};
\end{scope}


\coordinate (obsL) at ($(L42.east)+(0.10,0)$);
\draw[->, line width=0.8pt, draw=accentRed!75!black,
      shorten >=2pt, shorten <=2pt]
    (obsL) .. controls +(0.55,0.12) and +(-0.55,0.12) ..
    ($(dfstbox.west)+(0,0.25)$)
    node[pos=0.5, above=3pt, xshift=-2pt, font=\scriptsize, text=accentRed!80!black,
         fill=white, inner sep=1pt]
        {observation $x_t$};

\coordinate (wrR) at ($(R02.west)+(-0.10, 0.10)$);
\draw[->, line width=0.8pt, draw=colWrite!80!black,
      shorten >=2pt, shorten <=2pt]
    ($(dfstbox.east)+(0,0.12)$)
    .. controls +(0.55,0.10) and +(-0.55,0.10) ..
    (wrR)
    node[pos=0.5, above=3pt, font=\scriptsize, text=colWrite!85!black,
         fill=white, inner sep=1pt]
        {write $\sigma_t$};

\coordinate (mvR) at ($(R02.west)+(-0.10,-0.15)$);
\draw[->, line width=0.8pt, draw=colMove!80!black,
      shorten >=2pt, shorten <=2pt]
    ($(dfstbox.east)+(0,-0.18)$)
    .. controls +(0.55,-0.10) and +(-0.55,-0.10) ..
    (mvR)
    node[pos=0.5, below=3pt, font=\scriptsize, text=colMove!85!black,
         fill=white, inner sep=1pt]
        {move $\mu_t$};

\end{tikzpicture}

%% file: Sections/Experiments.tex
\section{Experimental Results and Discussion} \label{sec:exp}

In this section, we discuss length-generalization results solely due to differentiable training.

\subsection{Length Generalization on Arithmetic}\label{sec:arithmetic-results}

\begin{figure}[t]
\centering
\includegraphics[width=0.95\linewidth]{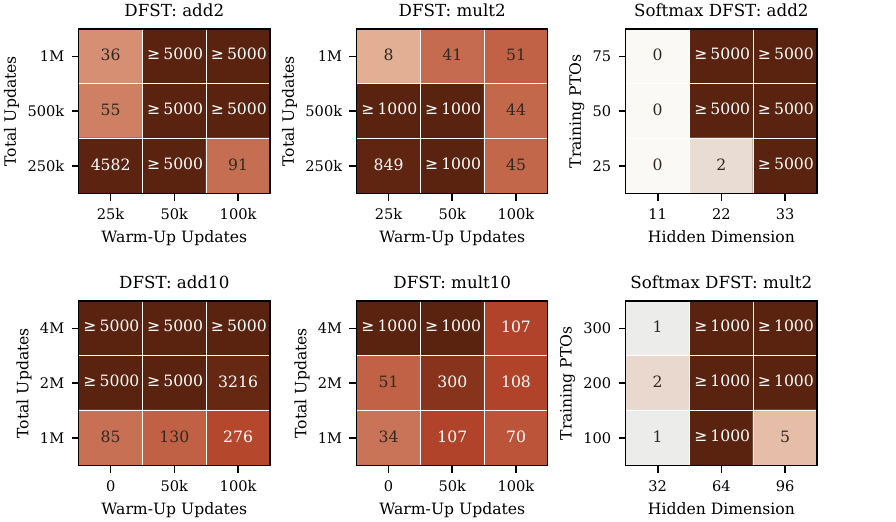}
\caption{\textbf{Robust Generalization on Arithmetic from Differentiable Training.} The first two columns show ordinary DFST learning-rate schedule sweeps. The last column shows Softmax DFST dimension/data sweeps after 100k updates of constant learning-rate training. Cells report RLG in operand digits. \Cref{app:rnn-sweeps} provides the paired ULG/RLG plots.}
\label{fig:heatmaps-arithmetic}
\label{fig:softmax-generalization}
\end{figure}

\Cref{fig:heatmaps-arithmetic} compares arithmetic generalization before extraction. All DFST and Softmax DFST models train from scratch with the same simple deterministic initialization recipe.

Several DFST runs reach both the ULG and RLG testing caps: 5000 digits per operand for addition and 1000 for multiplication, in binary and decimal radix, after training on operands of at most 3 and 5 digits, respectively. Success varies across training schedules, and additional updates need not improve generalization.

After 100k constant learning-rate updates, many Softmax DFST configurations reach the robust  binary-addition and multiplication caps. The Softmax DFST runs hold the advantage of providing interpretable transition probabilities while also supporting extreme generalization. However, Softmax DFST optimization proves more challenging, and we could not fit even small training sets on the decimal tasks. \Cref{app:optimization-limits} discusses these limitations.

\myparagraph{Tens of Millions of Consecutive Correct Actions.}
Operand length understates the scale of this extrapolation. Adding two 5000-digit all-9 numbers requires 50\,070\,010 PTO steps; multiplying two 1000-digit all-9 numbers requires 15\,012\,991. Each step requires the correct write and motion, including the final stopping action. The corresponding binary all-1 probes require exactly the same numbers of steps (\Cref{app:pto-lengths}). Successful decimal DFSTs reproduce the all-9 trajectories, while multiple binary DFST and Softmax DFST configurations reproduce the equally long all-1 trajectories. The binary controllers learn from only 25--75 addition PTOs or 100--300 multiplication PTOs on operands of at most 3 or 5 digits, respectively. Such training supports faithful execution over tens of millions of steps on these structured probes, far beyond the short training traces.

\myparagraph{Small-Data Learning and Implicit Bias.}
These long executions raise a question beyond representational capacity: why does gradient-based training favor rules that continue working so far beyond the training lengths? Matching the training PTOs alone does not force such behavior. We construct spurious binary addition and multiplication controllers that reproduce every expert PTO within the training-length bounds, yet stop prematurely on longer inputs. Thus, even exhaustive supervision over the entire 3 or 5-digit input domains cannot distinguish the expert algorithms from these degenerate alternatives. Furthermore, the spurious controllers need only 14 and 38 states (\Cref{app:spurious-add2,app:spurious-mult2}), compared to the 11 and 32 states for the binary arithmetic experts.

The larger Softmax DFSTs in \Cref{fig:heatmaps-arithmetic} have enough capacity to express these spurious solutions: addition uses dimensions 22 and 33, and multiplication uses 64 and 96 (\Cref{app:spurious-capacity}). Nevertheless, five of six configurations in each larger sweep reach the robust testing cap. These successful configurations therefore avoid the constructed spurious rules despite sufficient capacity to represent those rules. Gradient-based optimization with the chosen parametrization and initialization repeatedly favors behavior that survives long execution horizons despite the training set failing to distinguish other spurious targets. \textbf{These results support the existence of a useful implicit bias towards generalizing rules.} Characterizing the source and reliability of this bias remains an open direction.

\subsection{Length Generalization on the Universal Task}\label{sec:universal-results}

\begin{figure}[t]
\centering
\includegraphics[width=0.95\linewidth]{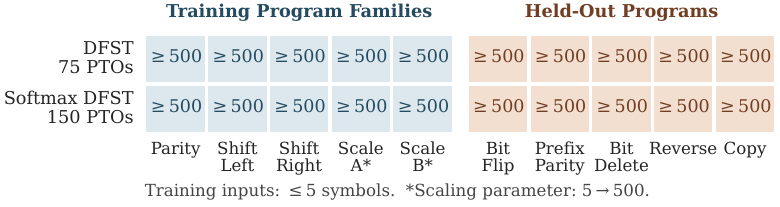}
\caption{\textbf{Generalization on Universal Simulations from Differentiable Training.} Cells report successful tested input lengths under teacher-forced exact action matching. Both controllers reach the 500-symbol cap after training on inputs of at most 5 symbols. \Cref{fig:universal-sweeps} gives the full sweeps.}
\label{fig:universal-main}
\end{figure}

A 15-dimensional DFST with 4,335 parameters, trained on 75 PTOs from 5 programs on inputs of at most 5 symbols, reaches the 500-symbol cap on all 10 universal-simulation tasks (\Cref{fig:universal-main}). A 30-dimensional Softmax DFST with 12,720 parameters, trained on 150 PTOs, also reaches every task cap. The longest sampled trajectory requires 15,316,822 consecutive expert-action matches (\Cref{tab:pto-horizons}).

These evaluations test 3 distinct forms of generalization: longer inputs, transfer to 5 unseen programs, and larger encoded Turing machines. Each controller keeps the same learned weights across all evaluations. Thus, a fixed model can execute novel computations without retraining, simply by changing the input Turing machine.

\subsection{Controller Families and Training Dynamics}\label{sec:controller-dynamics}

\begin{figure}[t]
\centering
\includegraphics[width=0.95\linewidth]{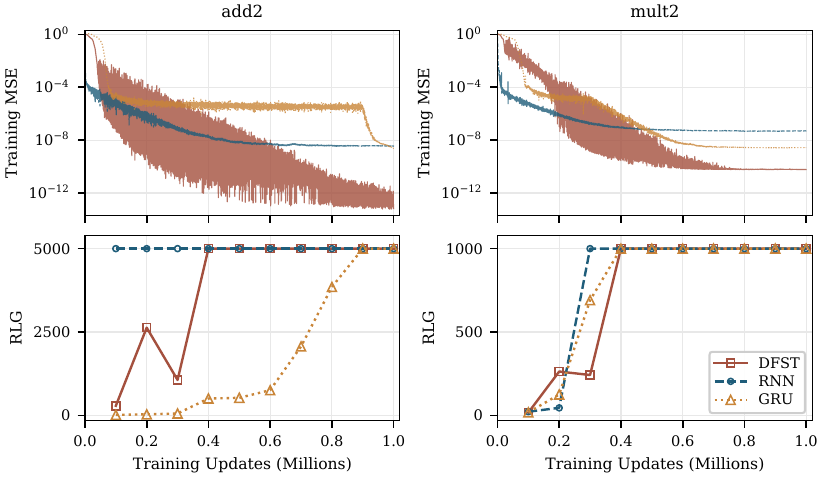}
\caption{\textbf{Learning Dynamics across Controller Families.} All six runs reach the 5000/1000-digit caps within one million updates. Runs use different dimensions, seeds, and schedules.}
\label{fig:dynamics-main}
\end{figure}

All six DFST, RNN, and GRU runs reach both binary-task caps (\Cref{fig:dynamics-main}). Each controller learns from 25 addition PTOs or 100 multiplication PTOs. These results support the same learning recipe across three distinct controller families: state tracking, unbounded external memory, and PTO supervision.

In \Cref{fig:dynamics-main}, the loss curves track fitting of the short training PTOs, while the RLG curves measure generalization to long structured inputs. All later recorded checkpoints retain the caps. Intermediate generalization need not improve monotonically, and low training loss does not certify algorithmic correctness. \Cref{app:training-dynamics} and \Cref{tab:dynamics-config} provide the full comparison and complete training settings.

%% file: Sections/Exact_Learning.tex
\section{Verification of Exact Learning} \label{sec:exact-learning}

Beyond the finite length-generalization tests in \Cref{sec:exp}, we extract and verify discrete controllers, then identify conditions for transferring correctness to neural execution (\Cref{fig:learning-pipeline}). Even with discrete controllers, the unbounded grid prevents a general sufficient testing horizon:

\begin{restatable}[Undecidability of Finite-Horizon Verification]{lemma}{undecidabilitylemma} \label{lem:undecidable}
No algorithm can take descriptions of two grid agents $T,M$ and return a finite input-length bound $L$ such that agreement of the PTOs $T[z]=M[z]$ for every $z\in\Sigma^{\leq L}$ implies agreement on all inputs.
\end{restatable}

\Cref{app:finite-precision} proves the claim by reduction from the halting problem. Thus, in general, no algorithm can guarantee that length-generalizing up to a certain cap proves exact learning. To certify exact learning, we will have to consider the specific controller structure. And since finite-precision neural controllers represent prohibitively large FSTs, we first need to extract something manageable. 

\subsection{Recovering Explicit Controllers} \label{subsec:extraction}

\myparagraph{Metric Extraction.}
We feed random inputs to a trained DFST controller and unroll the hidden-state trajectories from the executions. We connect hidden states within $\eps$ Euclidean distance and cluster the hidden states into connected components which become the initial discrete states. Each discrete state points to other discrete states through symbol transitions inherited from the hidden states inside the connected component. Thus, we have obtained a non-deterministic FST. We begin to merge the discrete states until either we get a deterministic FST and return, or until we get some inconsistency. In case of inconsistency, we halve $\eps$ and start anew. For some small enough $\eps$, we will obtain a consistent deterministic transition graph. The resulting graph reproduces the collected trajectories, leaving unobserved transitions undefined. Interestingly, we empirically verify that after enough sampled hidden-state trajectories, the extracted graph stabilizes. Note that this type of extraction uses no expert hidden-state labels. See \Cref{app:metric-extraction} for more details.

\myparagraph{Softmax Annealing and Direct Readout.}
A trained Softmax DFST assigns probabilities to successor states and actions. We continue training with a $G(p)=1-\|p\|_2^2$ penalty which sharpens the distributions. Progressively increasing (annealing) the penalty weight concentrates the probabilities into approximately one-hot vectors. Column-wise $\argmax$ then selects deterministic discrete transitions, thereby extracting an FST. \Cref{app:softmax-extraction} gives the full objective and schedule.

\begin{figure}[t]
\centering
\includegraphics[width=\linewidth]{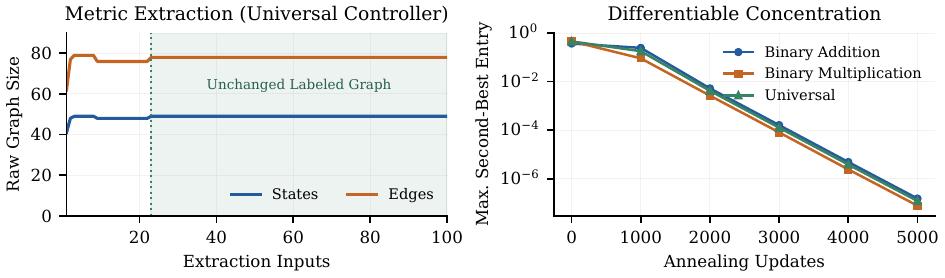}
\caption{\textbf{From Trained Controllers to Interpretable Graphs.} \textbf{Left:} the extracted graph stabilizes after 23 sampled trajectories, providing finite evidence for graph stability. \textbf{Right:} runner-up probabilities approach $10^{-7}$ during annealing; longer annealing with larger regularization coefficients reaches the smallest positive subnormal probabilities, which remain distinct from exact zeros.}
\label{fig:extraction-summary}
\end{figure}

Both extraction routes produce candidate discrete controllers. \Cref{fig:extraction-summary} shows empirical graph stability and approximate one-hot probability concentration.

\subsection{Proving Target Equivalence for the Recovered Algorithms} \label{subsec:equiv-results}

We relate extracted states and grid configurations to phases of the expert algorithm. The relation holds initially, and we prove that each permitted transition produces matching actions and preserves the relation. Algorithm-specific invariants show that the phases cover every valid execution. Induction then establishes agreement for arbitrary execution lengths, including simultaneous stopping. Different internal representations and state counts can therefore implement the same algorithm (\Cref{app:simulation-principle}).

\myparagraph{Eight Certificates.}
We provide five metric-extraction and three annealing certificates. \Cref{thm:add2-equiv,thm:mult2-equiv,thm:add10-equiv,thm:mult10-equiv} cover binary and decimal arithmetic on arbitrary nonempty operands, including unequal lengths and leading zeros. The universal certificates extend far beyond the five training programs. \Cref{thm:universal-equiv} proves that the recovered discrete controllers implement the interpreter's rules for arbitrary correctly encoded machines and input lengths, yielding the following consequence.

\begin{corollary}[Exact Learning of a Universal Controller]\label{cor:learned-universal}
The universal controllers obtained by metric extraction and softmax annealing from differentiably trained DFSTs and Softmax DFSTs correctly simulate every binary Turing machine on every nonempty binary input.
\end{corollary}

GPT-6 Astra generated the proofs. We verified the arguments. The \href{https://github.com/fanagor/DEL/tree/main/analyses}{repository checkers} mechanically verify finite transition obligations, while the induction arguments establish unbounded correctness.

\subsection{Transferring Guarantees to Neural Controllers} \label{subsec:stability}

Both routes transfer the recovered FST certificates to neural execution under explicit assumptions.

\myparagraph{Graph Stability for Metric Extraction.}
Graph stability requires every further neural hidden-state trajectory to preserve the graph extracted from trajectories $S$ at threshold $\eps$, up to state renaming.

\myparagraph{Uniformly Bounded Leakage for Softmax DFSTs.}
We assume $\|h_t-e_{q_t}\|_1\leq\varepsilon$ uniformly over valid inputs and execution lengths, where $q_t$ tracks the unminimized argmax controller driven by the same observations. This bounds off-target leakage and drift from the intended one-hot state. At a one-hot hidden state, let $\mu$ denote the smallest score advantage of the prescribed write or motion over any competing choice, across all valid executions. Hidden-state leakage can reduce this advantage by at most $\varepsilon$. Rounding can reduce the advantage by another $2\rho$, since each of the two compared scores can change by up to $\rho$. Thus, $\mu>\varepsilon+2\rho$ ensures that both readouts still select the prescribed actions.

\begin{restatable}[Behavioral Faithfulness]{theorem}{faithfulnesstheorem} \label{thm:faithfulness}
Under \Cref{ass:stability}, a deterministic trace-consistent metric extraction reproduces every PTO of the neural controller on every valid input. Under \Cref{ass:softmax-leakage}, the Softmax DFST and the argmax controller likewise produce identical PTOs. In either case, trace-preserving quotient reduction retains this equivalence, so an expert-equivalence certificate for the extracted controller transfers to the neural controller on the same domain.
\end{restatable}

\Cref{app:faithfulness-proof} proves both cases. For Softmax DFSTs, the readout margin exceeds the combined leakage and rounding perturbations, so induction preserves every action and environment update.

%% file: Sections/Conclusion.tex
\section{Conclusion}\label{sec:conclusion}
Our framework places differentiable parameter search inside a discrete algorithmic shell: train a neural controller, extract a discrete model through a fixed rule, and verify exact agreement with the target algorithm. Since extraction follows a fixed simple procedure (especially with $\argmax$ readout for Softmax DFSTs), we view this combination of differentiable search and discrete readout as \emph{differentiable exact learning}. The certificates concern the recovered discrete models, while the graph-stability or bounded-leakage assumptions additionally transfer correctness to neural execution.

%% file: Appendix/Related_Work.tex
\section{Related Work} \label{app:related}

Our work builds on a long line of research on \emph{neural program induction}, which studies the problem of learning algorithmic procedures using neural architectures. Within this paradigm, prior work has explored a wide variety of learning signals and model classes. Our framework combines the following three desiderata:
\begin{enumerate}[label = (\arabic*), topsep=2.5pt, left=0pt, itemsep=0pt, labelsep=5pt]
    \item Learning from expert behavior observations;
    \item Computational universality of the learning framework;
    \item Continuous optimization in a fully differentiable setting.
\end{enumerate}
This work investigates exact learning at the intersection of these three principles, with equivalence proofs for extracted controllers over all valid inputs. Specifically, we focus on learning scenarios where intelligent systems infer algorithmic processes solely from externally observable behavior, as in learning-from-demonstration settings \citep{argall2009survey}. Crucially, behavior observations circumvent the general identification barrier for input--output supervision \citep{gold1967language, papazov2025learning}, without access to the expert's internal state. Furthermore, we require computational universality and differentiability from our framework in order to study learning as a general-purpose mechanism, avoiding task-specific instruction templates and hand-crafted search procedures during training.

Prior universality results differ in the computational resources and learning guarantees provided:
\begin{itemize}[label = $\bullet$, topsep=2.5pt, left=0pt, itemsep=0pt, labelsep=5pt]
    \item \textbf{Precision and Context Requirements:} RNNs and Neural GPUs with unbounded-precision internal arithmetic \citep{siegelmann1992computational, perez2018on}; Transformers with growing context windows or unbounded-precision internal arithmetic \citep{perez2018on, dehghani2018universal, bhattamishra2020computational, merrill2023expresssive}.
    \item \textbf{External-Memory Constructions:} Bounded-precision recurrent controllers coupled to growing memory \citep{chung2021turing}; language models prompted with a complete production-rule system \citep{schuurmans2024autoregressive}.
    \item \textbf{Horizon-Dependent Architectures:} Statistical approximation guarantees for bounded-time computation with model sizes depending on the execution horizon \citep{wei2022statistically}.
\end{itemize}
These results establish computational expressivity or statistical approximation rather than recovery of a persistent controller from behavior observations. Finite-precision universality alone therefore does not distinguish our framework. Our universal controller instead learns from 75 PTOs without supplied control states or a transition table, transfers to five programs absent from training, and yields extracted-controller certificates covering arbitrary valid machine encodings (\Cref{app:universal,app:exact}). The certified controllers retain the same parameters across unbounded input lengths and execution horizons.

We now compare the supervision and verification guarantees of related learning frameworks.

\textbf{External-Memory Frameworks.} The work of \citet{zaremba2016learning} comes, perhaps, closest to our setting. The authors consider an RNN-based controller which operates over a read-only input grid and a write-only unidirectional output tape by issuing action tokens containing move and write commands. \citet{zaremba2016learning} investigate whether the neural controller can learn to execute simple algorithms (e.g., copying, reversing, addition) when trained via reinforcement learning for correct outputs or with supervision from expert policies. Unlike in our experiments, training uses nontrivial curricula, and off-the-shelf Q-learning fails on most tasks without additional modifications. The task suite includes single-digit multiplication and demonstrates substantial extrapolation on selected tasks. Our experiments additionally cover unrestricted two-operand multiplication and universal simulation using an unbounded read--write workspace.

Related neurosymbolic frameworks include Higher-Order RNNs \citep{giles1989higher}, Memory Networks \citep{weston2014memory}, Neural Turing Machines \citep{graves2014neural}, Neural GPUs \citep{kaiser2015neural}, Neural Random-Access Machines \citep{kurach2015neural}, Neural Programmers \citep{neelakantan2015neural}, Pointer Networks \citep{vinyals2015pointer}, Stack-Augmented RNNs \citep{joulin2015inferring}, Differentiable Neural Computers \citep{graves2016hybrid}, Hierarchical Attentive Memories \citep{andrychowicz2016learning}, TerpreT \citep{gaunt2016terpret}, R3NNs \citep{parisotto2016neuro}, Differentiable Forth Interpreters \citep{bovsnjak2017programming}, and Neural-Symbolic Stack Machines \citep{chen2020compositional}. External storage and even unbounded differentiable memory already appear in this literature \citep{grefenstette2015learning}. Our distinction concerns learning latent controller state from PTOs and certifying the extracted computation, rather than external memory alone.

\textbf{Input-Output Observations.} Another closely related approach appears in \citet{zaremba2015reinforcement}, which considers an LSTM controller operating over a read-only input tape, a read--write memory tape, and a write-only unidirectional output tape. The addition of a read--write memory renders the resulting model computationally universal. Nevertheless, learning proceeds solely via RL with rewards for correct output. \citet[Theorems I.3 and I.5]{gold1967language} rule out identification of the full class of recursive functions from input--output observations; these results do not prohibit learning a particular arithmetic algorithm. PTOs avoid this general identification barrier \citep{papazov2025learning}, while the present experiments investigate finite-data optimization and verification.

\textbf{Program-Structure Observations.} A particularly relevant line of work stems from Neural Programmer-Interpreters (NPIs) \citep{reed2015neural}. NPIs use an LSTM controller as a program executor: At each step, the controller consumes a local environment snapshot together with the active subprogram embedding and current arguments, then outputs (i) a distribution over the next subprogram in a fixed library, (ii) the corresponding arguments, and (iii) a halt/return decision. Training traces explicitly supervise this call--return structure. Each program invocation starts a fresh recurrent state; return resumes the caller's state. These annotations expose program structure, without supplying the numerical recurrent hidden vectors.

This additional program supervision supports data-efficient learning\footnote{The addition experiments train on 200 execution traces with operands of at most three digits.} and in special cases permits formal guarantees of exactness. \citet{cai2017making} observe that when the expert program follows a recursive template in which each subprogram admits only finitely many base configurations and reductions, one can formally certify that the trained LSTM controller recovers the ground-truth algorithm. The experiments already reach 5000-digit addition, with all-input proofs for addition and several sorting tasks. Our distinction concerns recovering persistent controller state without prescribed subprogram identities or recursive boundaries, and extending the certificates to full binary and decimal multiplication and universal simulation. \citet{pierrot2019learning} further demonstrate reinforcement learning through hierarchical program specifications, pre/postcondition checks, and recursive tree search; an induction argument certifies unbounded Tower of Hanoi execution for agents satisfying specified base-case and recursive-trace conditions. Our training instead uses PTO next-action losses without a supplied recursive decomposition or program library.

\textbf{Exact Instruction Learning and Logic Induction.}
\citet{backdeluca2025learning} prove high-probability exact instruction execution in the infinite-width neural tangent kernel regime using explicit instruction templates, but \emph{require retraining whenever the memory budget increases}; neural training experiments cover permutations, while arithmetic training remains untested because of ensemble cost. Our controllers retain fixed parameters across input lengths; the extracted-controller proofs impose no memory or execution-horizon bound.

\citet{payani2019inductive} recover explicit recursive programs from positive/negative facts and supplied background predicates, including addition for multiplication, without NPI-style program-structure supervision; the empirical studies use finite domains and provide neither length-generalization measurements nor equivalence proofs over all valid inputs. Our multiplication controllers learn digit-level actions and carry propagation without a supplied addition subroutine, and admit all-input proofs after extraction.

\textbf{Arithmetic Length Generalization.}
Finally, we situate our contribution in the arithmetic generalization literature.

Recent empirical studies \citep{joulin2015inferring, deletang2022neural, liu2022transformers, zhang2022paradox, shen2023positional, kazemnejad2023impact, jelassi2023length} demonstrate that Transformers, modified RNNs, and LSTMs length-generalize poorly on a variety of simple algorithmic tasks when trained on input-output data. With additional data-formatting strategies (such as task-specific index hints, optimized positional encodings, input reversing, position coupling, and scratchpad guidance) multiple works on arithmetic generalization in Transformers \citep{anil2022exploring, jelassi2023length, zhou2023algorithms, lee2023teaching, zhou2024transformers, hou2024universal, cho2024position, cho2024arithmetic} achieve substantial finite-length extrapolation. Most notably, for decimal addition, \citet{zhou2024transformers} report above 98\% exact-match accuracy (EMA) on 100-digit summands after training on decimal addition with up to 40-digit numbers (2.5$\times$ LG); and \citet{cho2024position} report above 95\% EMA on roughly 6--7$\times$ longer operands, with a separate scale-up experiment reaching 500 digits after training on operands of up to 160 digits. The clearest quantitative distinction concerns \emph{both operands} in decimal multiplication: \citet{cho2024arithmetic} train on operands of at most $10\times10$ digits and attain at least 78.55\% median EMA through $20\times15$ digits, counting the entire response including the scratchpad, whereas \citet{hou2024universal} extrapolate from $50\times1$ or $50\times3$ digits to $100\times1$ or $100\times3$ digits with 97\% accuracy. Our selected decimal-multiplication controller trains on operands of at most $5\times5$ digits and passes every action-level probe at $1000\times1000$ digits before extraction; the extracted controller additionally admits a proof over all valid inputs.

Interestingly, similar to us, \citet{hou2024universal} train on scratchpad data consisting of the computational traces of a Turing machine (TM) designed to solve the specific algorithmic task. However, whereas our agent only observes the evolution of the environment, \citet{hou2024universal} supervise explicit machine states, head positions, and tape contents. The random-machine experiments evaluate one-step prediction on longer tapes, without establishing correct execution over unbounded trajectories. The accompanying RASP construction guarantees $\exp(n)$ simulated steps with a program of size $O(n)$ under distinct-$n$-gram and in-memory execution assumptions; this construction does not certify experimentally trained weights. Our PTOs withhold expert control states, and the extracted-controller proofs impose no input-length or execution-horizon bound.

We conclude by comparing our results in \Cref{tab:comparison} against the Neural GPU \citep{kaiser2015neural, price2016extensions}, which achieves strong arithmetic generalization from input-output observations.
\citet{kaiser2015neural} introduce a Neural Cellular Automaton, called the Neural GPU, which achieves remarkable random-test generalization when trained solely on input-output examples (and an upper bound on the computation length) of binary addition and multiplication. Error-free sampled binary multiplication already reaches 2000 bits, exceeding our 1000-digit testing cap. However, \citet{price2016extensions} report sensitivity to the random seed and document long-carry and symmetric-input failures, despite greater than 99\% whole-answer accuracy for decimal multiplication at 100 digits. Our selected models pass all repdigit cross-pairs at the testing caps in both bases, including decimal multiplication, and the extracted controllers admit proofs of exact learning. The single-seed DFST schedule sweeps do not establish superior robustness across random seeds. Furthermore, our DFST agents train on small PTO datasets using Adam, gradient clipping, and a learning-rate schedule, while Neural GPU training combines curricula, parameter-sharing relaxation, dropout, and gradient noise. As shown in \Cref{tab:comparison}, observable intermediate steps support extreme generalization with far fewer training examples, although PTOs provide richer supervision than input-output pairs. RNN and GRU runs also reach both binary-task caps, supporting the broader state-tracking recipe beyond DFSTs (\Cref{app:training-dynamics}).

\begin{table*}[h!]
\centering
\begin{threeparttable}
\caption{Neural GPU vs. DFST on Arithmetic Tasks. DFST entries report the selected runs before extraction; lengths count bits in binary and digits in decimal.}
\label{tab:comparison}
\begin{tabular}{clcccc}
\toprule
Framework & Metric & \texttt{add2} & \texttt{add10} & \texttt{mult2} & \texttt{mult10} \\
\midrule
\multirow{4}{*}{\shortstack{Neural GPU\\with\\Input--Output Data}} & \# Samples & $\sim200$k & -- & $\sim200$k & -- \\
& \# Parameters & 10368 & -- & 10368 & -- \\
& Training Digits & 20 & -- & 20 & -- \\
& RLG & $\leq 20$ & -- & $\leq 8$ & -- \\
& ULG & 2000 & -- & 2000 & -- \\
\midrule
\multirow{2}{*}{\textbf{DFST (ours)}} & \# Samples & 25 & 250 & 100 & 10000 \\
& \# Parameters & 891 & 8227 & 5280 & 162108 \\
with & Training Digits & 3 & 3 & 5 & 5 \\
\multirow{2}{*}{PTOs}
& RLG & $\geq$ 5000 & $\geq$ 5000 & $\geq$ 1000 & $\geq$ 1000 \\
& ULG & $\geq$ 5000 & $\geq$ 5000 & $\geq$ 1000 & $\geq$ 1000 \\
\bottomrule
\end{tabular}
\end{threeparttable}
\end{table*}

\textbf{Summary.}
Our contribution combines latent-state discovery from behavior, extreme extrapolation in both arithmetic operands, and exact verification within one universal framework. Eight certificates cover extracted controllers across all five trained tasks. Graph stability or a suitable invariant of the annealed dynamics additionally transfers these guarantees to the neural controllers (\Cref{app:extraction,app:exact}).

%% file: Appendix/DFSTs.tex
\section{Differentiable Finite-State Transducers} \label{app:dfsts}

This appendix collects the formal details of DFST expressivity deferred from \Cref{sec:framework}. We prove \Cref{thm-universal}, discuss the non-uniqueness of emulating parametrizations, and relate the required hidden-state dimension to the number of states of the target agent.

\subsection{Proof of Universality}

We restate the theorem for convenience.

\universalitydfst*

\begin{proof}
Let $T = (Q, \Sigma, \Gamma, \d, \g, q_0, q_f) \in \A$ denote a grid agent with $d$ states $q_0, \dots, q_{d-1} = q_f$. Write $k = |\Sigma|$ for the alphabet size with symbols $s_0, \dots, s_{k-1}$, and $r = |\M|$ for the number of motion tokens $m_0, \dots, m_{r-1}$. We encode states, symbols, and motions as standard basis vectors: $q_i \leftrightarrow e_{i+1} \in \R^d$, $s_j \leftrightarrow e_{j+1} \in \R^k$, and $m_\ell \leftrightarrow e_{\ell+1} \in \R^r$.

Define the DFST weights $(A, B, C, h_0)$ as follows. Set $h_0 = e_1 = (1, 0, \dots, 0)^\top \in \R^d$. For each transition $\d(q_i, s_j) = q_{i'}$ with $\g(q_i, s_j) = (s_{j'}, m_\ell)$, assign
\begin{equation*}
    A[j,\, i',\, i] = 1, \qquad B[j,\, j',\, i] = 1, \qquad C[j,\, \ell,\, i] = 1,
\end{equation*}
and set all remaining tensor entries to $0$.

Under this assignment, if $h_{t-1} = e_{i+1}$ encodes state $q_i$ and the input $x_t = e_{j+1}$ encodes symbol $s_j$, then $A[x_t]\, h_{t-1} = e_{i'+1}$, $B[x_t]\, h_{t-1} = e_{j'+1}$, and $C[x_t]\, h_{t-1} = e_{\ell+1}$. Since $h_0 = e_1$ encodes the initial state $q_0$, induction on $t$ shows that the DFST$(A, B, C, h_0)$ exactly reproduces the output sequence of $T$ on every input. This construction uses only $\{0, 1\}$-valued weights and hence works at any finite arithmetic precision $p \geq 1$.

\textbf{Softmax DFSTs in Exact Real Arithmetic.}
Let $D_x\in\{0,1\}^{d\times d}$ denote the deterministic transition matrix for read symbol $x$, and let $O_x^s\in\{0,1\}^{k\times d}$ and $O_x^m\in\{0,1\}^{r\times d}$ denote the corresponding one-hot output matrices. Set $n=d+1$ and $u=\mathbf 1_n/n$. Define
\begin{equation*}
 V=\begin{pmatrix}I_d\\-\mathbf 1_d^\top\end{pmatrix},
 \qquad W=\begin{pmatrix}I_d&0\end{pmatrix}-\mathbf 1_d u^\top.
\end{equation*}
Then $WV=I_d$, $Wu=0$, and $\mathbf 1_n^\top V=0$. Choose $0<\alpha<1$ and $\eta>0$ sufficiently small that every entry of
\begin{equation*}
 A[x]=u\mathbf 1_n^\top+\alpha V D_xW,
 \qquad h_0=u+\eta V e_{q_0}
\end{equation*}
remains strictly positive. Such choices exist because the alphabet and state set contain finitely many elements. The columns of $A[x]$ and the vector $h_0$ sum to one. Induction gives
\begin{equation*}
 h_t=u+\eta\alpha^t V e_{q_t},
 \qquad q_t=\delta(q_{t-1},x_t).
\end{equation*}
For either output alphabet of size $a\in\{k,r\}$, write $u_a=\mathbf 1_a/a$, $V_a=I_a-u_a\mathbf 1_a^\top$, and $O_x\in\{O_x^s,O_x^m\}$. Choose $\beta>0$ sufficiently small that the column-stochastic matrix
\begin{equation*}
 P[x]=u_a\mathbf 1_n^\top+\beta V_aO_xW
\end{equation*}
has strictly positive entries, and use $P=B$ or $P=C$ accordingly. If $y_t$ denotes the expert output on $(q_{t-1},x_t)$, the readout satisfies
\begin{equation*}
 P[x_t]h_{t-1}=u_a+\beta\eta\alpha^{t-1}(e_{y_t}-u_a).
\end{equation*}
The coordinate $y_t$ therefore attains the unique maximum whenever $a>1$; a singleton output alphabet needs no comparison. Every strictly positive probability vector admits finite logits given by the coordinatewise logarithm. Thus these tensors and initialization define a Softmax DFST reproducing every expert action at every finite time. The margins shrink with $t$, so this exact-real construction alone supplies no finite-precision robustness guarantee.

\textbf{Softmax DFSTs under Rounded Arithmetic.}
For the $d$-dimensional one-hot construction above, assign logit $0$ to each selected coordinate and $-L$ to every other coordinate. A stable softmax whose exponential rounds $\exp(-L)$ to zero produces exactly the required one-hot column; apply the same construction to the initial state. For example, correctly rounded binary32 exponentiation with round-to-nearest underflows $\exp(-128)$ to zero. Once the implementation produces these vertices and preserves the corresponding matrix--vector operations, the original induction applies unchanged for every execution length. The saturation condition concerns the effective probabilities, rather than large finite logits alone.

For Turing-completeness, observe that standard Turing machines arise as grid agents restricted to a single row of the grid. Each construction emulates every such agent under the stated arithmetic conditions, so both controller families can simulate any Turing machine.
\end{proof}

\subsection{Non-Uniqueness of Emulating Parametrizations}

The one-hot construction in the proof above produces one particular DFST that emulates a given grid agent $T$, but many other parametrizations achieve the same behavior. In exact arithmetic, any invertible matrix $R\in\Q^{d\times d}$ defines an equivalent controller through
\begin{align*}
 A'[x]&=R A[x]R^{-1}, & B'[x]&=B[x]R^{-1},\\
 C'[x]&=C[x]R^{-1}, & h'_0&=Rh_0.
\end{align*}
Induction gives $h'_t=Rh_t$ for every read sequence, while $B'[x_t]h'_{t-1}=B[x_t]h_{t-1}$ and $C'[x_t]h'_{t-1}=C[x_t]h_{t-1}$ preserve both readouts. Thus the transformed controller reproduces every action and PTO. Choosing $R=cI_d$ for distinct nonzero rational values $c$ already yields infinitely many distinct exact-arithmetic parametrizations, since $h_0\neq0$. These transformations preserve the PTO loss and therefore generate infinitely many zero-loss realizations of the expert's one-hot targets.

A fixed finite arithmetic format admits only finitely many parameter configurations. Arbitrary rational changes of basis need not preserve representability or execution under rounding, so the infinitude claim does not extend to fixed finite precision. Permutations of the one-hot construction nevertheless preserve the binary weights and vertex operations, yielding equivalent finite-precision realizations whenever those operations remain exact. Separately, adding unreachable states yields infinitely many discrete grid agents with the same observable policy.

\subsection{Hidden-State Dimension and State Complexity} \label{subsec:hidden}

\Cref{thm-universal} guarantees that an $n$-dimensional ordinary DFST suffices to emulate any $n$-state grid agent. Most ordinary-DFST experiments therefore match the hidden dimension to the expert state count, but successful learning does not require an exact match. For the 11-state \texttt{add2} expert, a sweep over hidden dimensions $d\in\{5,8,11,14,17\}$ reaches the 5000-digit RLG cap at every tested dimension except $d=5$. Thus both a smaller controller and controllers with excess capacity achieve extreme tested generalization. These finite probes do not establish exact simulation on arbitrary inputs or identify the minimum sufficient dimension.

The Softmax DFST sweeps provide complementary evidence for the benefits of additional capacity (\Cref{fig:softmax-arithmetic,fig:universal-sweeps}). Binary addition reaches both caps with dimensions 22 and 33, compared with 11 expert states; binary multiplication reaches both caps with dimensions 64 and 96, compared with 32 expert states. Universal simulation reaches every task cap with dimensions 30 and 45, compared with 15 expert states. Across these tasks, successful configurations therefore accommodate two or three times the expert state count. Larger dimensions do not improve every configuration, but these results support choosing excess capacity when the expert state count remains unknown: precise knowledge of that count need not precede successful training.

Precisely characterizing the minimum hidden dimension required for a DFST to simulate a given $n$-state FST remains an open direction. The one-hot construction supplies an upper bound of $n$ for ordinary DFSTs, while smaller representations may exploit the structure of the target transitions and outputs. Such a characterization should distinguish agreement on all input strings from agreement only on realizable environment traces, and account for arithmetic precision and readout margins. A separate question concerns whether gradient-based training can reliably discover representations near the minimum dimension.

%% file: Appendix/Grid_Agents_PTOs.tex
\section{Grid Agents, PTOs, and Task Catalog} \label{app:grid-agents}

This appendix collects the pedagogical interpretation of PTOs, the formal notation for grid-agent execution, the hardness results that motivate our learning paradigm, and the full catalog of expert agents used in our experiments.

\subsection{Pedagogical Interpretation of PTOs} \label{app:pto-interpretation}

Intelligent agents rarely acquire novel algorithmic skills from input–output pairs alone. Rather, learning typically proceeds from observing the evolution of an underlying computational process and inferring the target function from intermediate steps. The PTO formalism admits a concrete pedagogical interpretation: A grid agent models a teacher navigating a blackboard, with $q_t$ corresponding to the teacher's mental state, $p_t$ marking the chalk's position, and $x_{t+1}$ denoting the symbol written at that position. Students observe the chalk's movements $m_{t+1}$ and the evolving blackboard contents $s_{t+1}$ but lack access to the teacher's neuronal activity $q_t \to q_{t+1}$. PTOs thus capture a rich but reasonable chain of thought -- not unlike the algorithmic scratchpad data employed to boost length generalization in Transformers \citep{lee2023teaching}, yet grounded in the natural structure of agentic computation rather than requiring internal-state annotations \citep{hou2024universal, cai2017making}.

\subsection{Extended Notation for Grid-Agent Execution} \label{app:grid-notation}

We extend $\d$ and $\g$ recursively to $Q \times \SS$ in the standard way and denote by $T(x) = T(x_1 \dots x_n)$ the output sequence $\g(q_0, x_1),\, \g(\d(q_0, x_1), x_2),\, \dots,\, \g(\d(q_0, x_1 \dots x_{n-1}), x_n)$.

Each algorithmic task $\mathcal{P}$ comes with a set of well-specified inputs $S_\mathcal{P} \subseteq \SS$. For arithmetic tasks, each input $x \in S_\mathcal{P}$ occupies cells $(0,0),\dots,(|x|-1,0)$ on an otherwise empty grid. Universal inputs use the program-and-tape encoding in \Cref{app:encoding}. We also denote the initial world-state by $x$ (overloading notation). Running the expert grid agent $T$ on $x$ produces a PTO $T[x]$. We write $T\{x\}$ for the \emph{input trace} -- the sequence of symbols observed by $T$ during execution on $x$ -- so that $T(T\{x\})$ gives the corresponding output trace.

For a training set $S \subseteq S_\mathcal{P}$, we write $T[S]$ for the PTOs induced by $S$ and $T\{S\}$ for the input traces due to $S$. Let $I = T\{S_\mathcal{P}\}$ denote the set of all realizable input traces.

\subsection{Hardness of Learning from Restricted Input Traces} \label{app:hardness}

Since $I$ constitutes a restricted subset of $\SS$, a training set $S$ -- no matter how large -- may leave some transitions of $T$ unexplored. This incompleteness makes the learning problem fundamentally hard: Many irreducible FSTs can exhibit identical behavior on $I$.

Learning FSTs from complete data constitutes a classical problem in grammatical inference, enjoying polynomial-size characteristic sets \citep{gold1978complexity, oncina1992identifying, de1997characteristic, parekh2001learning} and efficient active-learning algorithms \citep{angluin1987learning, vaandrager2022new}. However, restricting all observable inputs to the unknown set $I$ dramatically increases difficulty. In particular, \citet[Theorem~17]{papazov2025learning} proved that no polynomial-size characteristic set exists for this restricted problem, implying that no algorithm can learn the policy of every grid agent from polynomially many PTOs with lengths polynomial in the number of states of the expert agent.

\subsection{Expert Agent Catalog} \label{app:agent-catalog}

\Cref{tab:agents} summarizes the state counts, alphabets, and runtimes of all grid agents and Turing machines used in our experiments. For arithmetic tasks, $m$ and $n$ denote the lengths in digits of the two operands; for all other tasks, $n$ denotes the input length.

For universal simulation, $Q_M$ denotes the state set of the simulated Turing machine $M$, including the halting state, and $|M[z]|$ counts the simulated machine's actions on input $z$, including the stopping action. The runtime bound counts each teleportation as one action.

The public \href{https://github.com/fanagor/DEL/blob/main/notebooks/controller_walkthrough.ipynb}{controller walkthrough} provides animated expert executions and instructions for inspecting individual computation steps.

\begin{table}[h!]
\centering
\caption{\centering Grid Agents and Turing Machines Used in Experiments.}
\label{tab:agents}
\setlength{\tabcolsep}{3pt}
\renewcommand{\arraystretch}{1.1}
\begin{tabular}{lccc}
\toprule
Task & \# States $|Q|$ & Alphabet $\Sigma$ & Runtime \\
\midrule
\multicolumn{4}{l}{\textit{Grid Agents}} \\
\midrule
\texttt{add2}   & $11$   & $\{0,1,\l,+\}$          & $O((m+n)^2)$ \\
\texttt{add10}  & $19$  & $\{0,\dots,9,\l,+\}$     & $O((m+n)^2)$ \\
\texttt{mult2}  & $32$  & $\{0,1,\l,\times\}$      & $O((m+n)^2)$ \\
\texttt{mult10} & $108$ & $\{0,\dots,9,\l,\times\}$& $O((m+n)^2)$ \\
\midrule
\multicolumn{4}{l}{\textit{Turing Machines (simulated by $U$)}} \\
\midrule
\texttt{parity}   & $3$   & $\{0,1,\l\}$           & $\T(n)$ \\
\texttt{flip}  & $2$  & $\{0,1,\l\}$               & $\T(n)$ \\
\texttt{cum-sum}  & $3$  & $\{0,1,\l\}$            & $\T(n)$ \\
\texttt{shift-left}  & $5$  & $\{0,1,\l\}$            & $\T(n)$ \\
\texttt{shift-right}  & $7$  & $\{0,1,\l\}$            & $\T(n)$ \\
\texttt{scaling-A}$(s)$ & $s+2$ & $\{0,1,\l\}$     & $\T(s)$ \\
\texttt{scaling-B}$(s)$   & $4s+1$   & $\{0,1,\l\}$& $\T(s)$ \\
\texttt{bit-delete}  & $5$  & $\{0,1,\l\}$         & $\T(n)$ \\
\texttt{reverse}  & $16$  & $\{0,1,\l\}$           & $\T(n^2)$ \\
\texttt{copy} & $10$ & $\{0,1,\l\}$                & $\T(n^2)$ \\
\midrule
\multicolumn{4}{l}{\textit{Universal Agent}} \\
\midrule
$U(\langle M, z\rangle)$ & $15$ & $\{0,1,\l, X, Y, Z, \mathtt{L}, \mathtt{R}, \mathtt{S}\}$ & $O(|Q_M|\,|M[z]|)$ \\
\bottomrule
\end{tabular}
\end{table}

%% file: Appendix/Experimental_Details.tex
\section{Implementation and Experimental Details} \label{app:experimental-details}

We specify the data, optimization, execution semantics, and evaluation protocol for the experiments. \Cref{app:related} compares these results with prior work.

The public repository provides \href{https://github.com/fanagor/DEL/tree/main/configs}{training recipes}, \href{https://github.com/fanagor/DEL/tree/main/runs}{saved model weights and run configurations}, and \href{https://github.com/fanagor/DEL/tree/main/paper}{commands for reproducing the figures and analyses}.

\subsection{Data and Controller Initialization} \label{app:dfst-init}

Arithmetic training uses small PTO datasets from distinct operand-string pairs, retaining leading zeros. Each set includes all single-digit pairs. Remaining pairs follow independent uniform sampling of each operand length from $1,\ldots,q$ and of each digit from $0,\ldots,R-1$, with duplicate pairs rejected. The ordinary-DFST datasets contain 25, 250, 100, and 10\,000 PTOs for \texttt{add2}, \texttt{add10}, \texttt{mult2}, and \texttt{mult10}, respectively. Addition uses $q=3$ and multiplication uses $q=5$. These experiments use the ordinary sampler without an additional teaching-set coverage core. The plotted DFST and Softmax DFST sweeps use data seed 42. Separately, we replicated strong length generalization across three data-generation seeds using the same deterministic initialization recipe.

Minibatch sampling chooses uniformly among occupied operand-length buckets, then uniformly among PTOs within the chosen bucket. Universal training first chooses a represented training task uniformly once per minibatch, then samples length buckets within that task. Consequently, training does not sample uniformly over all stored PTOs.

Ordinary DFSTs start from $A[x]=I_d$, $B[x]=0$, $C[x]=0$, and a trainable $h_0=\mathbf{1}_d$. In the ordinary-DFST schedule sweeps, hidden dimensions match the experts: 11, 19, 32, and 108 for the four arithmetic tasks, and 15 for universal simulation. Inputs use frozen one-hot embeddings. The resulting arithmetic controllers contain 891, 8227, 5280, and 162\,108 trainable parameters, respectively. Training uses \texttt{float32} arithmetic and logarithmic-depth prefix scans.

All DFST and Softmax DFST models begin PTO training from scratch without a pretrained checkpoint. Softmax DFSTs initialize the transition logits to $I_d$, output logits to zero, and initial-state logits to $\mathbf{1}_d$, then apply softmax normalization. Consequently, the effective initial state and output columns start uniformly, while transition columns favor the identity successor. Ordinary and softmax universal controllers contain 4335 and 12\,720 trainable parameters at dimensions 15 and 30, respectively.

\begin{table}[t]
\centering
\caption{\textbf{Generalization before Extraction.} Selected ordinary DFST runs; $N$ counts training PTOs and $q$ bounds training operand length. Caps indicate successful finite probes, without asserting a failure beyond the cap.}
\label{tab:current-results}
\small
\begin{tabular}{lrrrrrr}
\toprule
Task & $d$ & $N$ & $q$ & Updates & Warm-up & ULG / RLG \\
\midrule
\texttt{add2} & 11 & 25 & 3 & 500k & 50k & $\geq5000 / \geq5000$ \\
\texttt{add10} & 19 & 250 & 3 & 4M & 50k & $\geq5000 / \geq5000$ \\
\texttt{mult2} & 32 & 100 & 5 & 500k & 25k & $\geq1000 / \geq1000$ \\
\texttt{mult10} & 108 & 10\,000 & 5 & 4M & 0 & $\geq1000 / \geq1000$ \\
\bottomrule
\end{tabular}
\end{table}

\subsection{Optimizer and Learning-Rate Schedule} \label{app:opt-sched}

Ordinary DFST training uses Adam \citep{kingma2014adam} with $(\beta_1,\beta_2)=(0.9,0.999)$, $\epsilon=10^{-8}$, zero weight decay, batch size 32, and global gradient-norm clipping at 1. Let $T$ denote the update budget, $\omega$ the warm-up duration, $c$ the constant-rate duration, $\gamma$ the peak rate, and $\gamma_f$ the floor, with $\omega+c<T$. The schedule follows
\begin{equation}
\gamma(t)=
\begin{cases}
\gamma_f^{1-t/\omega}\gamma^{t/\omega}, & 0\leq t<\omega,\\[2pt]
\gamma, & \omega\leq t<\omega+c,\\[2pt]
\gamma^{1-(t-\omega-c)/(T-\omega-c)}\gamma_f^{(t-\omega-c)/(T-\omega-c)}, & \omega+c\leq t\leq T.
\end{cases}
\label{eq:ewec}
\end{equation}
A zero-duration phase contributes no updates. The DFST sweeps use $c=0$; GRU dynamics include the plateaus in \Cref{tab:dynamics-config}. Updates use $t=0,\ldots,T-1$; the endpoint at $T$ equals the floor. Within the DFST sweeps, binary arithmetic uses $\gamma=10^{-2}$; decimal arithmetic and universal simulation use $\gamma=10^{-3}$. The DFST sweeps use $\gamma_f=10^{-7}$; the separate multiplication dynamics run uses $10^{-9}$.

\begin{table}[t]
\centering
\caption{Ordinary-DFST Schedule Sweeps. Every Cartesian-product combination contributes one cell in the arithmetic plots. Universal simulation varies the training-set size and update budget.}
\label{tab:dfst-arith}
\small
\begin{tabular}{lccc}
\toprule
Task & Hidden dimension & Total updates $T$ & Warm-up $\omega$ \\
\midrule
\texttt{add2} & 11 & $\{250\mathrm{k},500\mathrm{k},1\mathrm{M}\}$ & $\{25\mathrm{k},50\mathrm{k},100\mathrm{k}\}$ \\
\texttt{mult2} & 32 & $\{250\mathrm{k},500\mathrm{k},1\mathrm{M}\}$ & $\{25\mathrm{k},50\mathrm{k},100\mathrm{k}\}$ \\
\texttt{add10} & 19 & $\{1\mathrm{M},2\mathrm{M},4\mathrm{M}\}$ & $\{0,50\mathrm{k},100\mathrm{k}\}$ \\
\texttt{mult10} & 108 & $\{1\mathrm{M},2\mathrm{M},4\mathrm{M}\}$ & $\{0,50\mathrm{k},100\mathrm{k}\}$ \\
Universal & 15 & $\{250\mathrm{k},500\mathrm{k}\}$ & $0$ \\
\bottomrule
\end{tabular}
\end{table}

\subsection{Loss and Training Loop} \label{app:loss}

For a minibatch, let $M_t$ mask valid timesteps and let $\mathbf{1}_{s_t}$ and $\mathbf{1}_{m_t}$ encode the expert write and motion actions. The masked two-head MSE reads
\begin{equation}
\L=\frac{\sum_t M_t\bigl(\|\hat s_t-\mathbf{1}_{s_t}\|_2^2+\|\hat m_t-\mathbf{1}_{m_t}\|_2^2\bigr)}{2\sum_t M_t}.
\label{eq:loss}
\end{equation}
The sum extends over the batch and time axes. Padding contributes neither recurrent updates nor loss. Training feeds expert observations into the controller and supervises the next write and motion. The ordinary-DFST runs use no curriculum, inverse-frequency weighting, or extraction during optimization. Logged losses average the action objective over updates in each logging window. The training-dynamics figure averages loss over 1000 updates and evaluates saved checkpoints every 100\,000 updates.

\subsection{Softmax DFST Training}\label{app:softmax-training}

The probability parametrization in \Cref{app:extraction} replaces each transition and output column by a softmax distribution and normalizes the initial state. The initial training sweep uses 100\,000 Adam updates at constant learning rate $0.1$, MSE, batch size 32, gradient clipping at 1, and zero regularization weight. For \texttt{add2}, dimensions $\{11,22,33\}$ combine with dataset sizes $\{25,50,75\}$; for \texttt{mult2}, dimensions $\{32,64,96\}$ combine with sizes $\{100,200,300\}$. Universal simulation uses dimensions $\{15,30,45\}$ and sizes $\{75,150,225\}$. The later annealing stage follows the separate procedure in \Cref{app:extraction}. Neither the softmax sweep nor annealing shares the ordinary-DFST learning-rate schedule by default.

\subsection{Additional State-Tracking Controllers} \label{app:rnn}

All controller families receive the currently observed symbol and emit separate write and motion scores. RNN and GRU heads act on the post-update hidden vector; DFST heads act on the previous vector together with the current observation. An Elman RNN uses
\begin{equation*}
h_t=\tanh(W_xx_t+W_hh_{t-1}+b),\qquad
\hat s_t=W_sh_t+b_s,\quad \hat m_t=W_mh_t+b_m.
\end{equation*}
GRU controllers replace the recurrence by gated reset, update, and candidate-state operations while retaining the same action heads. All recurrent baselines use one layer and frozen one-hot input embeddings. The RNN addition run initializes recurrent weights orthogonally, input weights with Xavier initialization, and recurrent biases to zero, with a fixed zero initial state. The RNN multiplication run and both GRU runs use zero input weights, zero recurrent biases, and a trainable all-ones initial state. For RNN multiplication, the recurrent matrix starts at the identity. For GRUs, only the candidate-state recurrent block starts at the identity; reset- and update-gate recurrent blocks start at zero. All RNN and GRU action heads use the linear layers' default random initialization.

\begin{table}[t]
\centering\small
\setlength{\tabcolsep}{3.5pt}
\begin{tabular}{lrrrrrrr}
\toprule
Task / controller & $d$ & $\gamma$ & $\gamma_f$ & $\omega$ & $c$ & Clip & Seed\\
\midrule
\texttt{add2} / DFST & 11 & $10^{-2}$ & $10^{-7}$ & 50k & 0 & 1 & 42\\
\texttt{add2} / RNN & 32 & $10^{-2}$ & $10^{-7}$ & 0 & 0 & 0.1 & 123\\
\texttt{add2} / GRU & 32 & $3\cdot10^{-3}$ & $10^{-7}$ & 100k & 800k & 0.1 & 17\\
\texttt{mult2} / DFST & 32 & $10^{-2}$ & $10^{-9}$ & 25k & 0 & 1 & 42\\
\texttt{mult2} / RNN & 64 & $3\cdot10^{-3}$ & $10^{-7}$ & 0 & 0 & 0.1 & 1\\
\texttt{mult2} / GRU & 64 & $3\cdot10^{-3}$ & $10^{-7}$ & 100k & 200k & 0.1 & 11\\
\bottomrule
\end{tabular}
\caption{\textbf{Controller-Comparison Training Configurations.} Every run uses $T=10^6$ Adam updates, MSE, batch size 32, zero weight decay, and no curriculum or inverse-frequency weighting. Warm-up and plateau durations count updates; cooldown lasts $T-\omega-c$. Clip denotes the global gradient-norm threshold. The seed controls dataset generation, model initialization, and training sampling; evaluation uses seed 42 for every run.}
\label{tab:dynamics-config}
\end{table}

Addition controllers contain 891, 1513, and 3977 trainable parameters for DFST, RNN, and GRU, respectively; multiplication controllers contain 5280, 5129, and 14089. Each addition dataset contains 25 PTOs with operands of at most three digits; each multiplication dataset contains 100 PTOs with operands of at most five digits. The different seeds produce different datasets within the same sampling recipe. Consequently, the training histories demonstrate successful learning across controller families without isolating architecture from initialization, data, and optimization choices.

\subsection{Length-Generalization Evaluation} \label{app:bp-eval}

Every probe streams an expert PTO and requires the predicted write \emph{and} motion to match at every real timestep, including the single final stopping action. ULG tests ten uniformly sampled pairs with both operands of exactly $L$ digits. RLG additionally tests all $R^2$ repdigit pairs $(a^L,b^L)$ with $a,b\in\{0,\ldots,R-1\}$: four pairs for binary arithmetic and 100 for decimal arithmetic. A deterministic generator indexed by the evaluation seed, task, and candidate length fixes random probes independently of the search history.

The search first tests the cap: 5000 digits for addition, 1000 for multiplication, and 500 input symbols for universal simulation. A successful cap probe terminates the search. Otherwise, binary search probes smaller lengths; RLG cannot exceed the measured ULG. Binary search assumes monotone pass/fail behavior with length and does not exhaustively test every shorter length. Consequently, reported values describe the implemented search, and the notation $\geq L_{\max}$ records a successful capped test, not an exact maximal horizon or a mathematical guarantee for all shorter inputs.

Evaluation uses batches of at most ten expert streams, chunks of 8192 transitions, and continuation state across chunks. Ordinary DFST evaluations, including the training-dynamics runs, use parallel matrix scans; the initial softmax evaluations use sequential recurrence. RNN and GRU evaluations retain the native recurrent implementation with continuation states across chunks, even when the shared evaluation configuration records the parallel scan option. These choices matter under finite precision: parallel and sequential products can differ through rounding, and ordinary-DFST sequential stabilization rescales hidden states while the parallel path uses unscaled products. All statements about agreement concern the recorded execution mode. Extraction and the conditional transfer arguments require the corresponding execution semantics rather than an unqualified interchange of numerical implementations.

\myparagraph{Teacher Forcing and Closed-Loop Agreement.}
For a deterministic recurrent execution rule and environment, exact next-action agreement along the complete expert PTO implies identical closed-loop execution on that input: induction gives the same environment, next observation, and controller history after every matching action. Floating-point parallel scans require additional care because the arithmetic association can differ from a stepwise rollout. We report the actual teacher-forced exact-match protocol and treat the autonomous extracted-FST equivalence proofs separately.

\myparagraph{Compute Reporting.}
The released configurations specify CUDA execution on single NVIDIA A100-SXM4-
40GB and Tesla V100-SXM2-32GB GPUs.

\subsection{Optimization Limitations and Dataset Coverage}\label{app:optimization-limits}

The ordinary-DFST sweep reaches the RLG cap in five of nine \texttt{add2}, three of nine \texttt{mult2}, five of nine \texttt{add10}, and two of nine \texttt{mult10} configurations. Additional updates need not improve RLG, and low training loss alone does not certify algorithmic correctness. The results demonstrate successful configurations across tasks and parametrizations, without asserting success for every schedule or seed.

Our Softmax DFST training attempts on decimal addition and multiplication failed to fit even very small training datasets. The softmax distributions quickly saturate, gradients vanish, and learning stalls, consistent with convergence to a poor local minimum. These failures concern fitting the training examples, before length generalization or extraction. Developing a robust training procedure for larger Softmax DFST controllers remains an open direction for future experiments.

Dataset coverage also affects generalization. Without guaranteed transition coverage, finite random sampling can omit a reachable expert transition with nonzero probability, leaving the corresponding behavior unrepresented in the training PTOs. An important research direction concerns identifying minimal teaching sets that cover all reachable expert transitions and provide sufficient information for exact learning and extreme length generalization. Transition coverage alone need not distinguish the expert from every alternative controller, so this objective also requires understanding which execution contexts resolve behavioral ambiguities. Our \href{https://github.com/fanagor/DEL/blob/main/docs/teaching_sets.md}{repository} implements preliminary teaching sets that exercise all reachable transitions of the task experts, but we have not yet evaluated these sets for neural-controller training.

%% file: Appendix/additional_experiments.tex
\section{Additional Generalization Results} \label{app:rnn-sweeps}

\subsection{Uniform and Robust Generalization}

\begin{figure}[t]
\centering
\begin{subfigure}{0.49\linewidth}
\includegraphics[width=\linewidth]{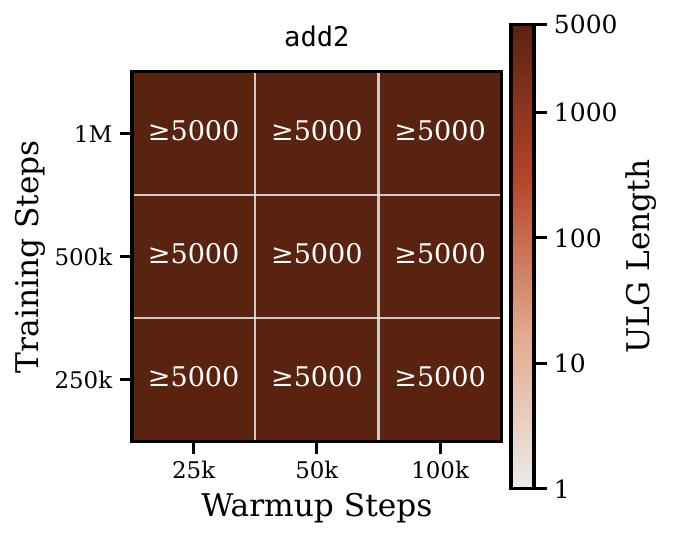}
\caption{Binary Addition.}
\end{subfigure}\hfill
\begin{subfigure}{0.49\linewidth}
\includegraphics[width=\linewidth]{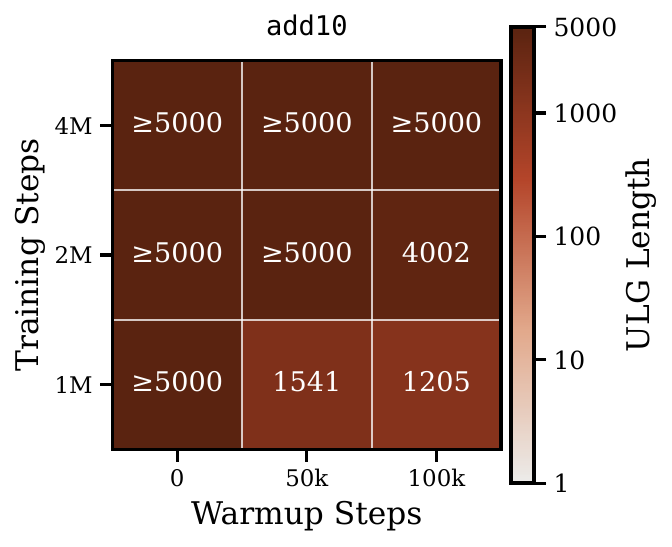}
\caption{Decimal Addition.}
\end{subfigure}\par\medskip
\begin{subfigure}{0.49\linewidth}
\includegraphics[width=\linewidth]{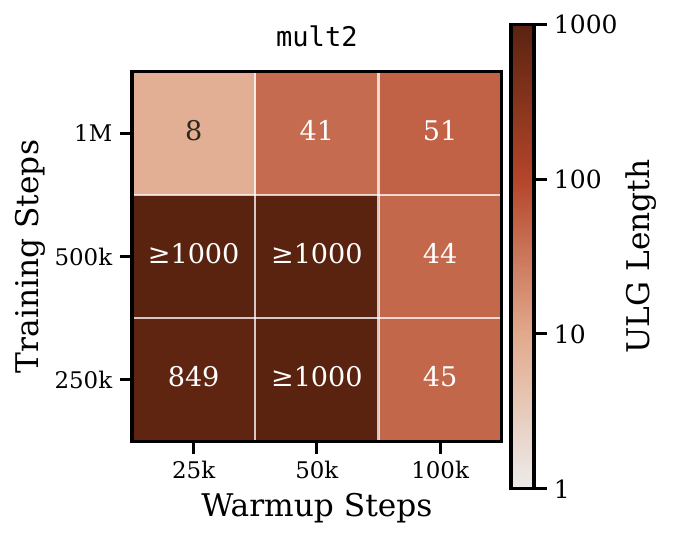}
\caption{Binary Multiplication.}
\end{subfigure}\hfill
\begin{subfigure}{0.49\linewidth}
\includegraphics[width=\linewidth]{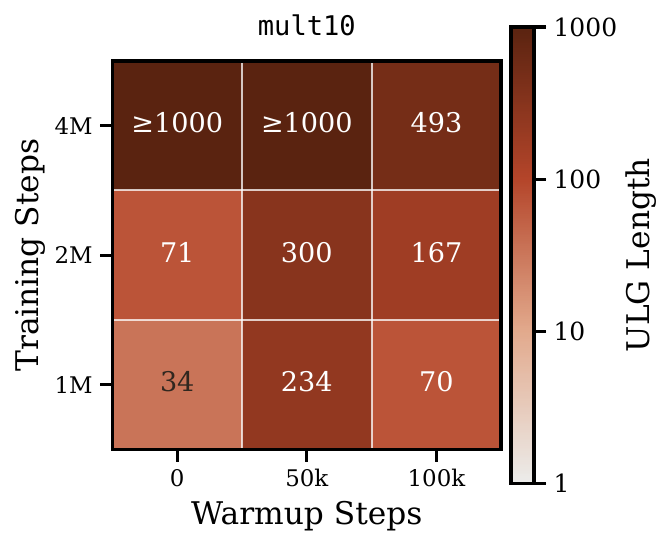}
\caption{Decimal Multiplication.}
\end{subfigure}
\caption{\textbf{DFST Uniform Length Generalization.} Each panel varies the update budget and warm-up duration. ULG requires exact action agreement along ten sampled expert trajectories; cap values denote successful finite probes. \Cref{fig:dfst-rlg-full} evaluates the same runs on structured inputs.}
\label{fig:dfst-ulg-full}
\end{figure}

\begin{figure}[t]
\centering
\begin{subfigure}{0.49\linewidth}
\includegraphics[width=\linewidth]{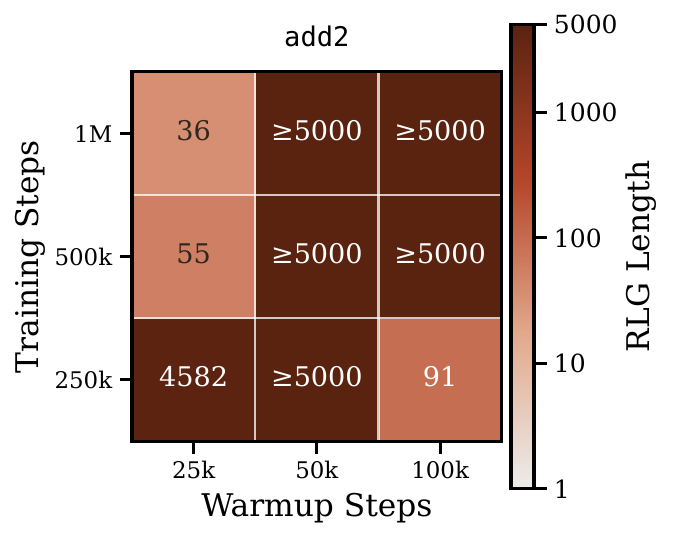}
\caption{Binary Addition.}
\end{subfigure}\hfill
\begin{subfigure}{0.49\linewidth}
\includegraphics[width=\linewidth]{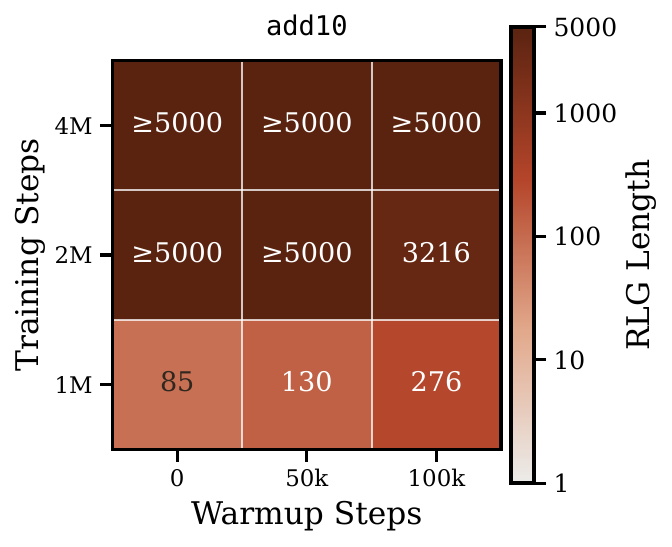}
\caption{Decimal Addition.}
\end{subfigure}\par\medskip
\begin{subfigure}{0.49\linewidth}
\includegraphics[width=\linewidth]{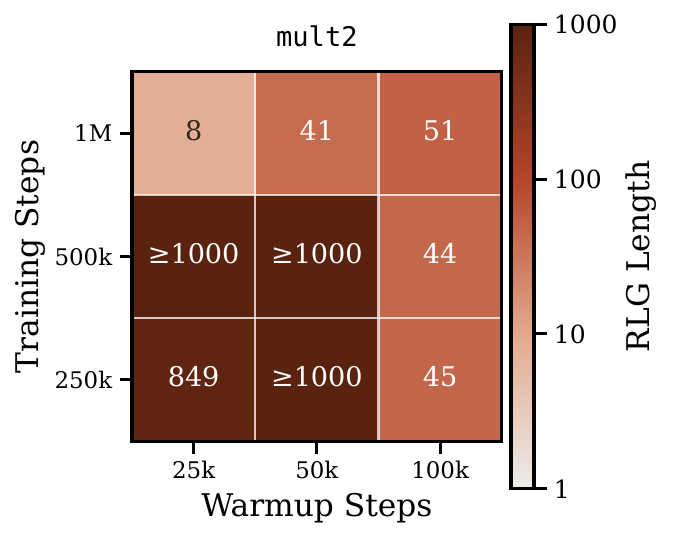}
\caption{Binary Multiplication.}
\end{subfigure}\hfill
\begin{subfigure}{0.49\linewidth}
\includegraphics[width=\linewidth]{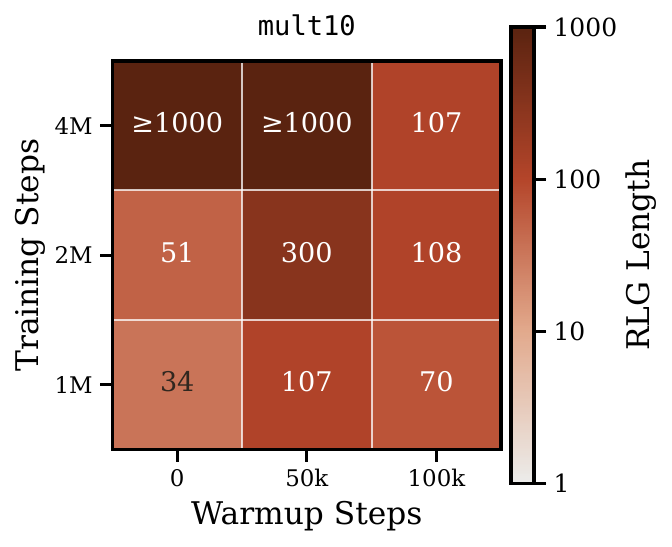}
\caption{Decimal Multiplication.}
\end{subfigure}
\caption{\textbf{DFST Robust Length Generalization.} The training schedules match \Cref{fig:dfst-ulg-full}. RLG supplements random-input testing with every repdigit cross-pair, four in binary and 100 in decimal. Cap values denote successful finite probes.}
\label{fig:dfst-rlg-full}
\end{figure}

\Cref{fig:dfst-ulg-full,fig:dfst-rlg-full} present the complete ULG and RLG sweeps for ordinary DFSTs. Uniform success can coexist with failures on structured repdigit inputs: for example, every \texttt{add2} run reaches ULG 5000, while only five of nine runs reach RLG 5000. The two measurements therefore reveal distinct failure modes.

\subsection{Softmax Parametrization}

\begin{figure}[t]
\centering
\begin{subfigure}{\linewidth}
\includegraphics[width=\linewidth]{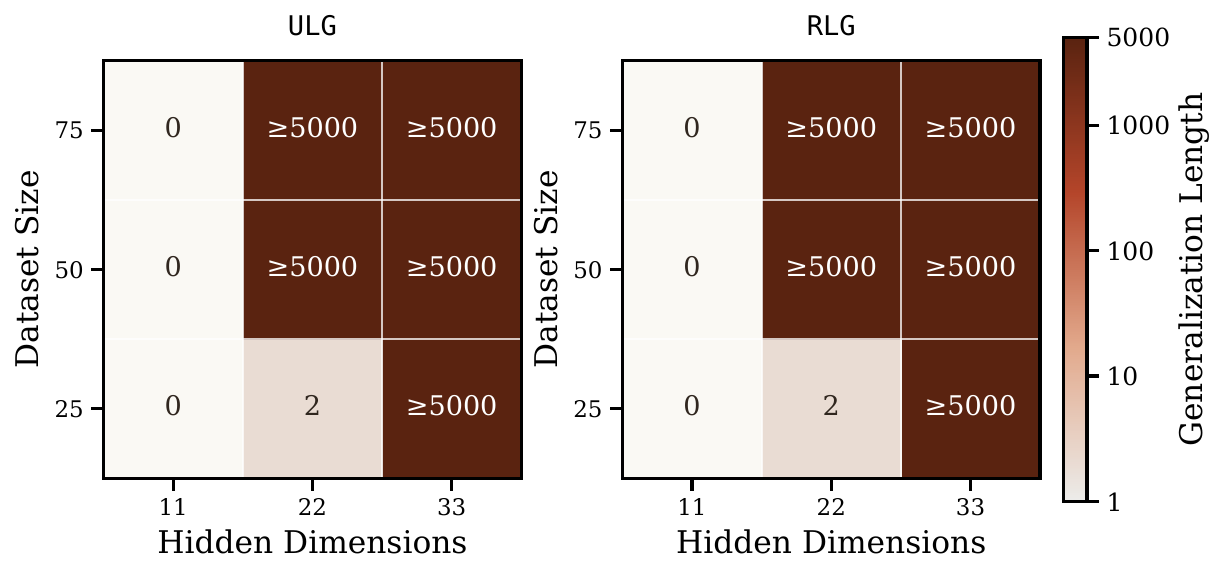}
\caption{Binary Addition.}
\end{subfigure}\par\medskip
\begin{subfigure}{\linewidth}
\includegraphics[width=\linewidth]{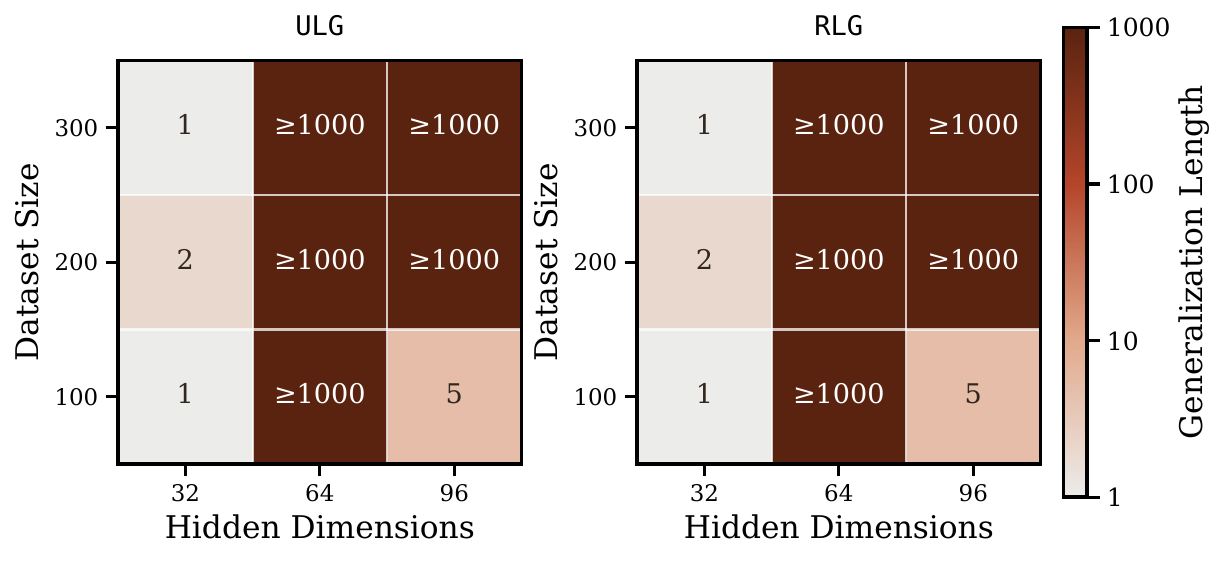}
\caption{Binary Multiplication.}
\end{subfigure}
\caption{\textbf{Initial Softmax DFST Generalization.} Each panel compares ULG and RLG across hidden dimensions and training-set sizes after 100k updates. The displayed results precede the annealing stage.}
\label{fig:softmax-arithmetic}
\end{figure}

\Cref{fig:softmax-arithmetic} reports the initial differentiable-training sweep, before annealing or argmax extraction. Five of nine binary-addition configurations reach both caps. Five of nine binary-multiplication configurations also reach both caps, providing starting points for the interpretable extraction procedure in \Cref{app:extraction}. Every recorded cell gives identical ULG and RLG, although ordinary DFST sweeps exhibit gaps between the two metrics. Increased state dimension and additional data can help optimization, but the sweep does not support monotone improvement with either quantity. The cap-reaching checkpoints use unregularized training; sharpening follows as a separate continuation stage.

\subsection{Training Dynamics across Controller Families}\label{app:training-dynamics}

The \href{https://github.com/fanagor/DEL/tree/main/analyses/training_dynamics}{released training histories} include loss logs, run configurations, and checkpoint evaluations for all six runs below.

\begin{figure}[t]
\centering
\begin{subfigure}{\linewidth}
\includegraphics[width=\linewidth]{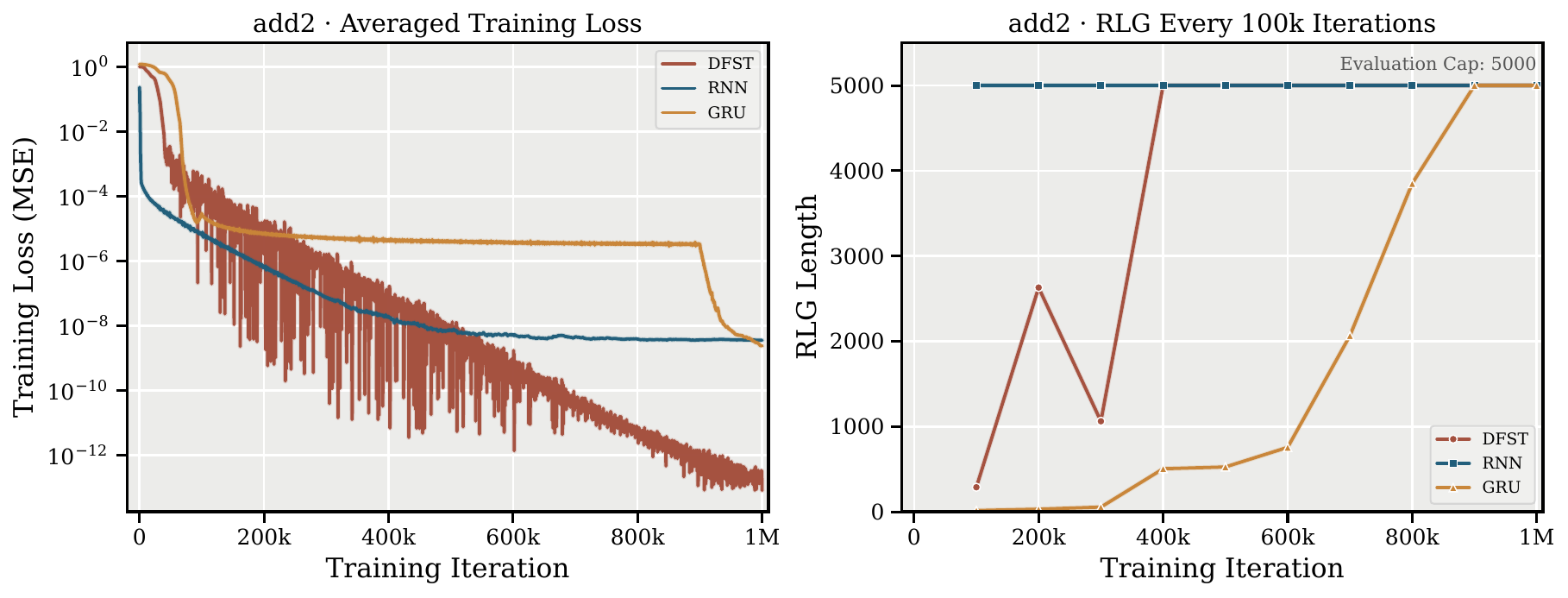}
\caption{Binary Addition: 5000-Digit Cap.}
\end{subfigure}\par\medskip
\begin{subfigure}{\linewidth}
\includegraphics[width=\linewidth]{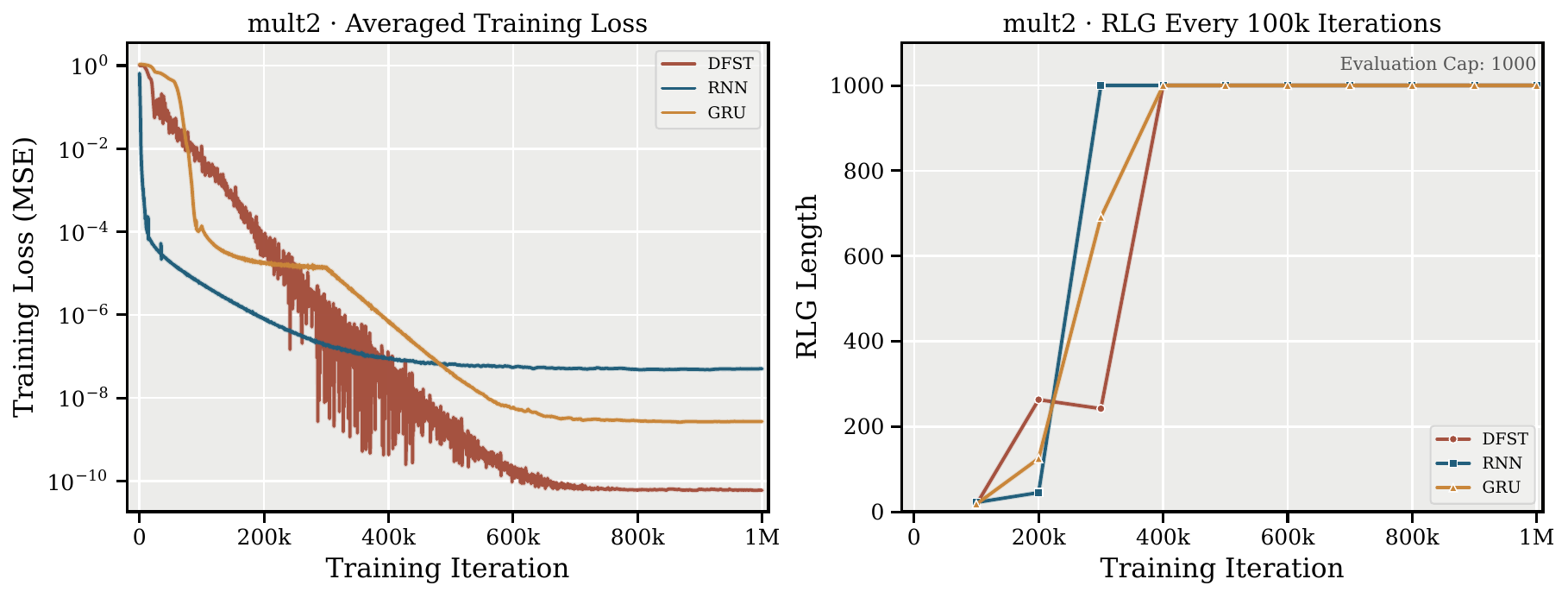}
\caption{Binary Multiplication: 1000-Digit Cap.}
\end{subfigure}
\caption{\textbf{Training Dynamics for Three State-Tracking Controllers.} Left panels show training MSE averaged over 1000 updates; right panels show RLG measured every 100k updates using all four binary repdigit pairs. Every final checkpoint reaches the task cap. Curves describe selected runs with different dimensions, seeds, initializations, and schedules (\Cref{tab:dynamics-config}), rather than an isolated comparison of architectures.}
\label{fig:dynamics-overlay}
\end{figure}

\Cref{fig:dynamics-overlay} compares million-update DFST, RNN, and GRU runs on both binary tasks. Each addition run uses 25 training PTOs with operands of at most three digits; each multiplication run uses 100 PTOs with operands of at most five digits. All six final checkpoints reach ULG and RLG caps of 5000 for addition and 1000 for multiplication. RLG tests all four binary repdigit cross-pairs.

\begin{table}[t]
\centering\small
\begin{tabular}{llrrr}
\toprule
Task & Controller & First ULG cap & First RLG cap & Final ULG / RLG\\
\midrule
\texttt{add2} & DFST & 200k & 400k & $\geq5000 / \geq5000$\\
 & RNN & 100k & 100k & $\geq5000 / \geq5000$\\
 & GRU & 500k & 900k & $\geq5000 / \geq5000$\\
\texttt{mult2} & DFST & 400k & 400k & $\geq1000 / \geq1000$\\
 & RNN & 300k & 300k & $\geq1000 / \geq1000$\\
 & GRU & 400k & 400k & $\geq1000 / \geq1000$\\
\bottomrule
\end{tabular}
\caption{\textbf{First Recorded Cap-Reaching Checkpoints and Final Results.} Checkpoints occur every 100k updates; first recorded success does not identify the exact crossing time. Final results refer to one million updates. Caps bound the tests rather than the algorithms.}
\label{tab:dynamics-checkpoints}
\end{table}

\begin{figure}[t]
\centering
\begin{subfigure}{\linewidth}
\includegraphics[width=\linewidth]{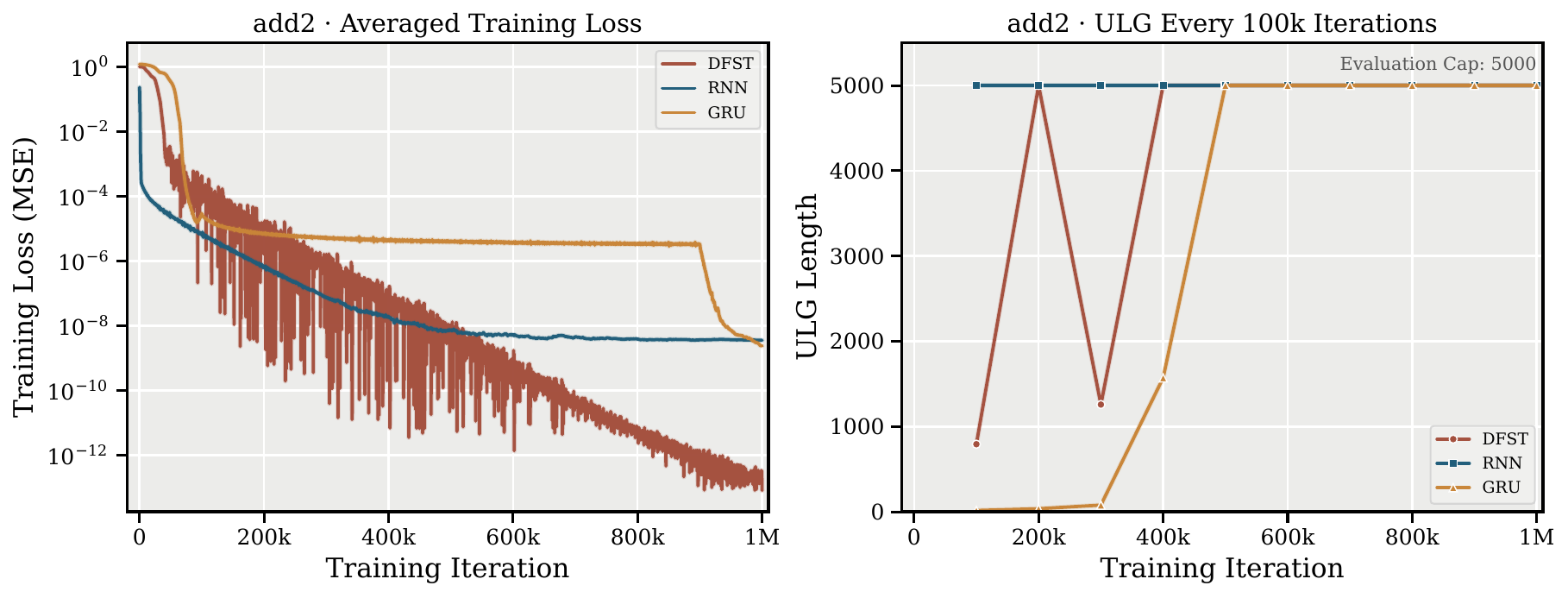}
\caption{Binary Addition: 5000-Digit Cap.}
\end{subfigure}\par\medskip
\begin{subfigure}{\linewidth}
\includegraphics[width=\linewidth]{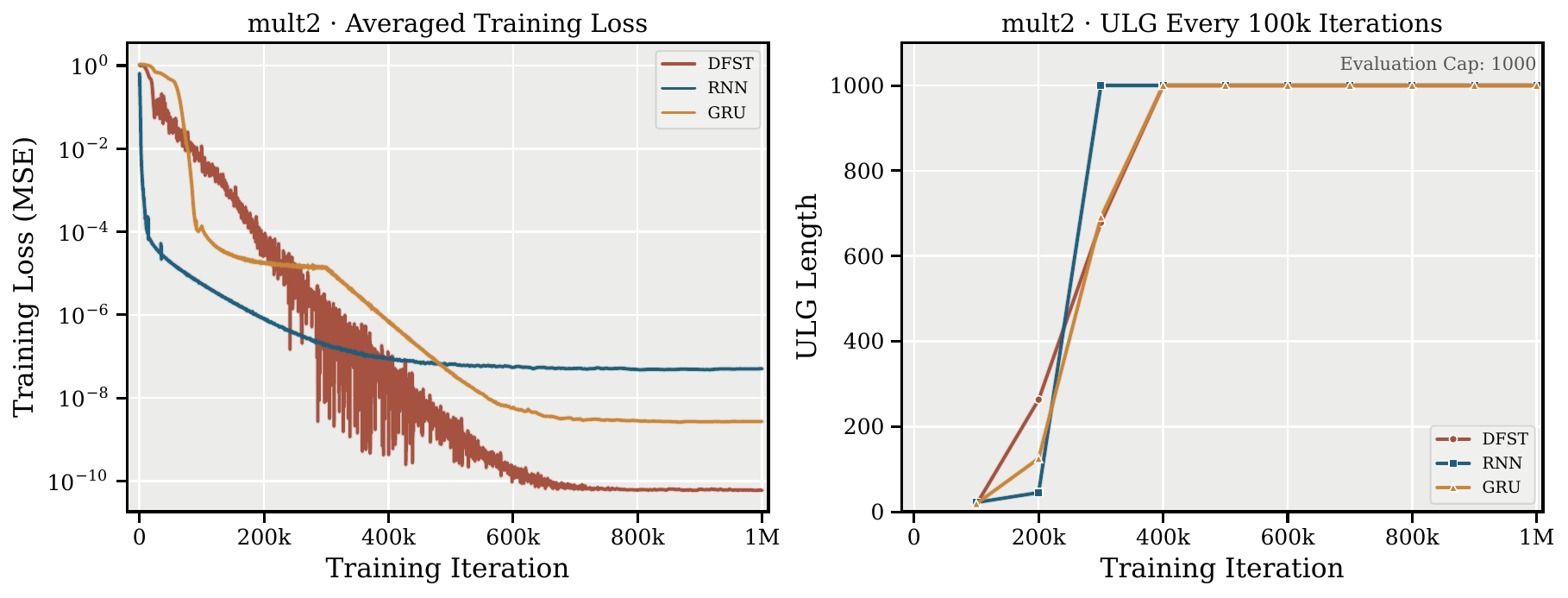}
\caption{Binary Multiplication: 1000-Digit Cap.}
\end{subfigure}
\caption{\textbf{Uniform Generalization during Training.} The loss curves and checkpoints match \Cref{fig:dynamics-overlay}; the right panels show ULG. Comparing ULG and RLG distinguishes random-input success from robustness to repdigit inputs.}
\label{fig:dynamics-ulg}
\end{figure}

The addition curves separate random-input and structured-input generalization. The RNN already passes both caps at the first evaluated checkpoint. At 200k updates, DFST reaches ULG 5000 but RLG 2630; both measurements fall at 300k, to 1258 and 1062, before reaching both caps at 400k. GRU reaches ULG 5000 at 500k updates while RLG reaches only 526; RLG subsequently reaches 5000 at 900k. These trajectories show that additional optimization can improve behavior on structured inputs after random-input tests saturate, and that intermediate generalization need not improve monotonically. \Cref{fig:dynamics-ulg} supplies the corresponding ULG curves.

All multiplication runs reach both caps by 400k updates. After first reaching an RLG cap, every later recorded checkpoint retains the cap. Across both tasks, the shared outcome supports state tracking with PTO supervision across multilinear, ungated recurrent, and gated recurrent controllers. Different hidden dimensions, initializations, seeds, and learning-rate schedules prevent a causal ranking of controller architectures; the histories establish successful learning for the selected configurations without quantifying success probabilities over random seeds.

The DFST multiplication trajectory uses a learning-rate floor of $10^{-9}$, distinct from the $10^{-7}$ floor in the schedule sweep. Both GRU runs include a constant-rate phase between exponential warm-up and cooldown. The complete settings appear in \Cref{tab:dynamics-config}.

%% file: Appendix/Universal.tex
\section{Universal-Agent Experiments} \label{app:universal}

The universal expert uses 15 states, a nine-symbol alphabet, and eight motion tokens. A single learned controller receives the encoded program through the environment; no controller weights change between simulated tasks.

\subsection{Teleportation Runes} \label{app:runes}

The controller observes only the symbol under the grid head and emits a symbol to write and a motion token. The environment applies the write before the motion. The alphabet contains $\{\l,0,1,\mathtt L,\mathtt R,\mathtt S,X,Y,Z\}$; motions comprise $\{L,R,U,D,S,X,Y,Z\}$, with $S$ stopping execution. The three symbols $X,Y,Z$ provide relocatable bookmarks. Writing a rune stores the symbol underneath the rune at the current cell. Rewriting the same rune restores the hidden symbol at the previous location and relocates the bookmark. A motion token $X$, $Y$, or $Z$ jumps directly to the corresponding bookmark; writing a rune alone does not trigger a jump. Jumping to an unplaced rune leaves the head in place; valid simulations place each bookmark before use. Ordinary writes overwrite the current cell and remove any rune occupying that cell. The simulation relation must therefore track both visible grid contents and hidden symbols beneath runes.

During simulation, $Z$ marks the simulated tape head, $X$ identifies the program-table block for the current simulated state, and $Y$ bookmarks the instruction currently undergoing decoding. These operations shorten transport between the program and tape while preserving a finite controller and an unbounded environment.

\subsection{Program Encoding} \label{app:encoding}

For a binary Turing machine with $s$ states, including the halting state, choose ternary width $\kappa=\max\{1,\lceil\log_3 s\rceil\}$. Encode state indices using $\mathtt{L},\mathtt{R},\mathtt{S}$ for ternary digits 0, 1, 2. Order program-table rows lexicographically by state index and read symbol, with blank preceding 0 and 1. Each row contains
\begin{equation*}
[\,x,\ \l,\ \operatorname{code}(q),\ 0,\ \operatorname{code}(q'),\ 0,\ w,\ m\,],
\end{equation*}
where $(q',w,m)$ denotes the Turing-machine transition on $(q,x)$. The two delimiter zeros separate state fields. State codes use most-significant trits first. Thus the table contains $3s$ rows and $2\kappa+6$ columns, with three consecutive rows per state. Unoccupied grid cells contain blanks. For example, with $s=3$, a transition $(q,x)=(0,1)\mapsto(2,0,R)$ contributes the row $[1,\l,\mathtt L,0,\mathtt S,0,0,\mathtt R]$. Initially, $X$ covers the blank in column 1 of the first program row. The input word occupies columns starting at 1 of the row below the program table, and the universal head starts at the first input symbol.

The controller starts in state 0. Each simulated step first records the tape-head position with $Z$ and jumps to $X$. A vertical search within the current-state block selects the row matching the read symbol. The controller then crosses the current-state code and decodes the next-state code one trit at a time: $Y$ bookmarks the instruction field, while $X$ advances through matching state-code columns using vertical searches. After locating the successor-state block, the controller returns $X$ to column 1, reads the write symbol and motion through $Y$, and jumps to $Z$ to execute the tape action. A simulated $S$ action performs the specified write and stops the universal controller. Rune relocation restores covered program symbols, allowing repeated interpretation of the same table without changing controller weights. The metric-extraction equivalence result in \Cref{app:exact} covers correctly encoded programs and nonempty binary input words. The annealed controller also admits the empty word; the two theorem domains remain distinct.

The simulation requires $O(|Q_M|\,|M[z]|)$ actions, where $Q_M$ denotes the state set of $M$, including the halting state, and $|M[z]|$ counts the actions of $M$ on $z$, including the stopping action. Each teleportation counts as one action.

\subsection{Training and Evaluation Programs} \label{app:simulated-tasks}

Training samples 75, 150, or 225 PTOs with input lengths at most five from five programs: \texttt{parity}, \texttt{shift-left}, \texttt{shift-right}, \texttt{scaling-A}(5), and \texttt{scaling-B}(5). Static sampling includes at least one example from each training program and fills the remaining slots by random task and input sampling. The implemented sampler permits empty words, although the metric-extraction equivalence theorem restricts the unbounded claim to nonempty words.

Evaluation includes all five training families and five held-out programs: bit inversion (\texttt{flip}), prefix parity (\texttt{cum-sum}), parity-dependent bit deletion (\texttt{bit-delete}), reversal, and copying. The last two programs require repeated scans over growing tape regions. The scaling tests set the program parameter to the candidate input length, so evaluation at 500 increases both input length and encoded machine size. \texttt{scaling-A}(s) executes $s$ nonhalting steps followed by one stopping action; \texttt{scaling-B}(s) follows a parity-dependent walk through a finite chain of states. These families test new program sizes as well as longer inputs.

\Cref{tab:binary-tm-behavior} summarizes the implemented programs. Alternatively, readers can inspect animated executions through the public \href{https://github.com/fanagor/DEL/blob/main/notebooks/controller_walkthrough.ipynb}{controller-animation notebook} or the \href{https://github.com/fanagor/DEL#interactive-tape-and-grid-viewer}{animation command}, for example \texttt{del animate tp:copy '1011'}.

\begin{table}[htbp]
\centering
\small
\setlength{\tabcolsep}{4pt}
\renewcommand{\arraystretch}{1.15}
\caption{\textbf{Implemented Binary Turing Machines.} Examples start from $1011$ in cells $0,\ldots,3$, with the head at cell 0. Notation $v@j$ places word $v$ starting at cell $j$; $\l$ denotes a blank, and all unlisted cells remain blank. Scaling examples use parameter $s=5$.}
\label{tab:binary-tm-behavior}
\begin{tabular}{@{}p{.17\linewidth}p{.57\linewidth}p{.21\linewidth}@{}}
\toprule
Program & Tape transformation & Final tape example \\
\midrule
\texttt{parity} & Erases the input and writes the parity of the number of ones immediately after the original word. & $1@4$ \\
\texttt{shift-left} & Moves the entire word one cell left, preserving bit order. & $1011@{-1}$ \\
\texttt{shift-right} & Moves the entire word one cell right, preserving bit order. & $1011@1$ \\
\texttt{scaling-A}$(s)$ & Performs $s$ state-indexed bit-flipping moves, then a symbol-preserving stopping action; see the rule below. & $00111@{-1}$ \\
\texttt{scaling-B}$(s)$ & Performs $s$ parity-dependent rewrites, changing direction after encountering blanks; the final rewrite stops. & $11011@0$ \\
\midrule
\texttt{flip} & Replaces each bit $b$ with $1-b$ in place. & $0100@0$ \\
\texttt{cum-sum} & Replaces each bit with the parity of the input prefix ending at that position. & $1101@0$ \\
\texttt{bit-delete} & Retains zeros for even input length and ones for odd input length; replaces other bits with blanks without compacting the tape. & $\l0\l\l@0$ \\
\texttt{reverse} & Reverses the word and relocates the result to cells $-(n-1),\ldots,0$ for nonempty input of length $n$. & $1101@{-3}$ \\
\texttt{copy} & Retains the input and appends a second copy after one blank cell. & $1011\l1011@0$ \\
\bottomrule
\end{tabular}
\end{table}

For \texttt{scaling-A}$(s)$, step $i=0,\ldots,s-1$ maps $0$ to $1$ and moves left, or maps $1$ to $0$ and moves right. On a blank, step $i$ writes $(i+1)\bmod2$ and moves left for even $i$, right for odd $i$. For \texttt{scaling-B}$(s)$, initialize parity $p=0$ and direction right. Reading a bit $b$ writes $p\mathbin\oplus b$ and updates $p$ to that value; reading a blank writes $p$ and reverses the direction for the next step. Each nonfinal step moves in the direction held before the read; step $s$ writes and stops.

\begin{figure}[t]
\centering
\begin{subfigure}{0.49\linewidth}
\includegraphics[width=\linewidth]{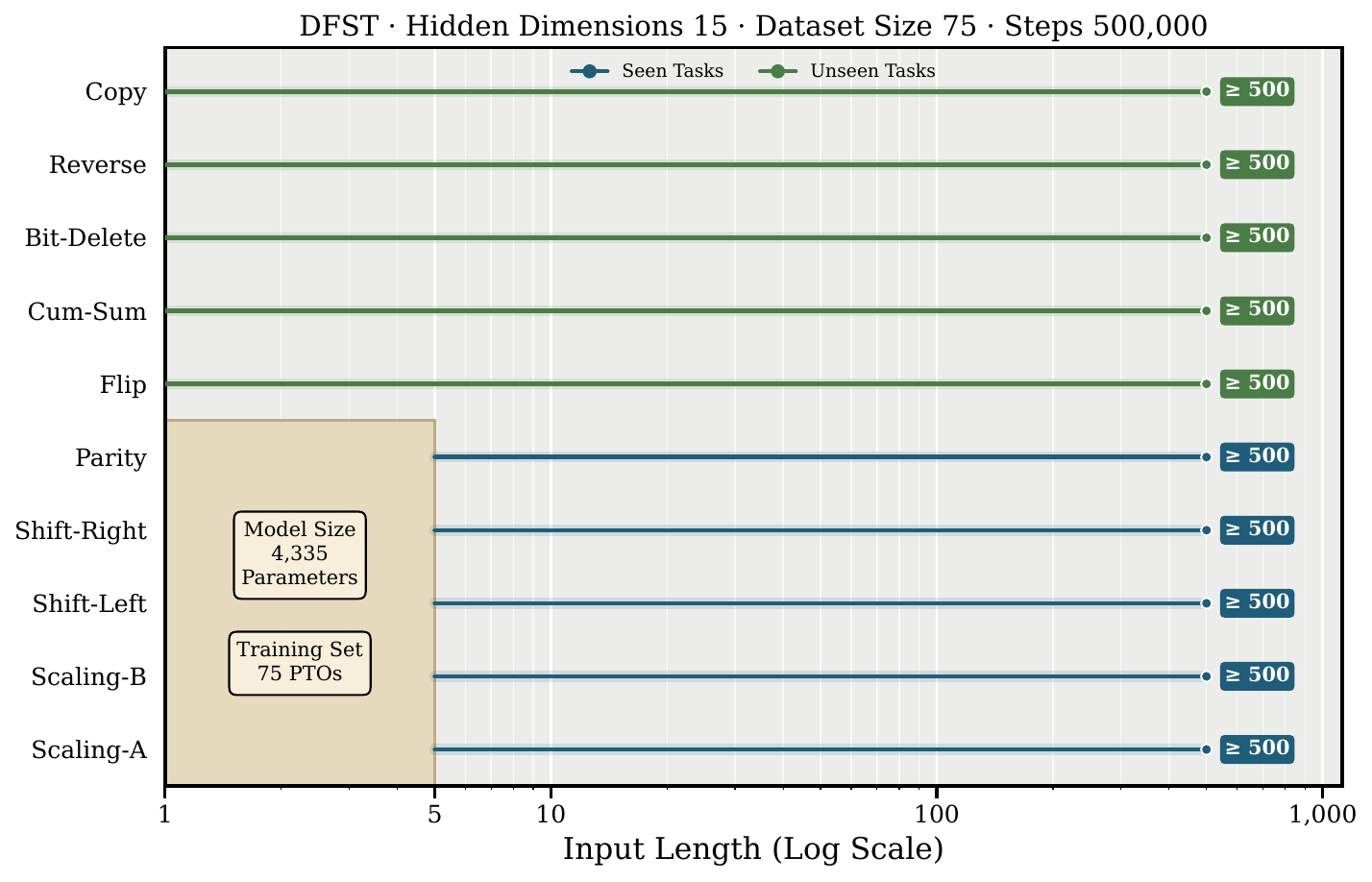}
\caption{DFST: $d=15$, $N=75$, 500k Updates.}
\end{subfigure}\hfill
\begin{subfigure}{0.49\linewidth}
\includegraphics[width=\linewidth]{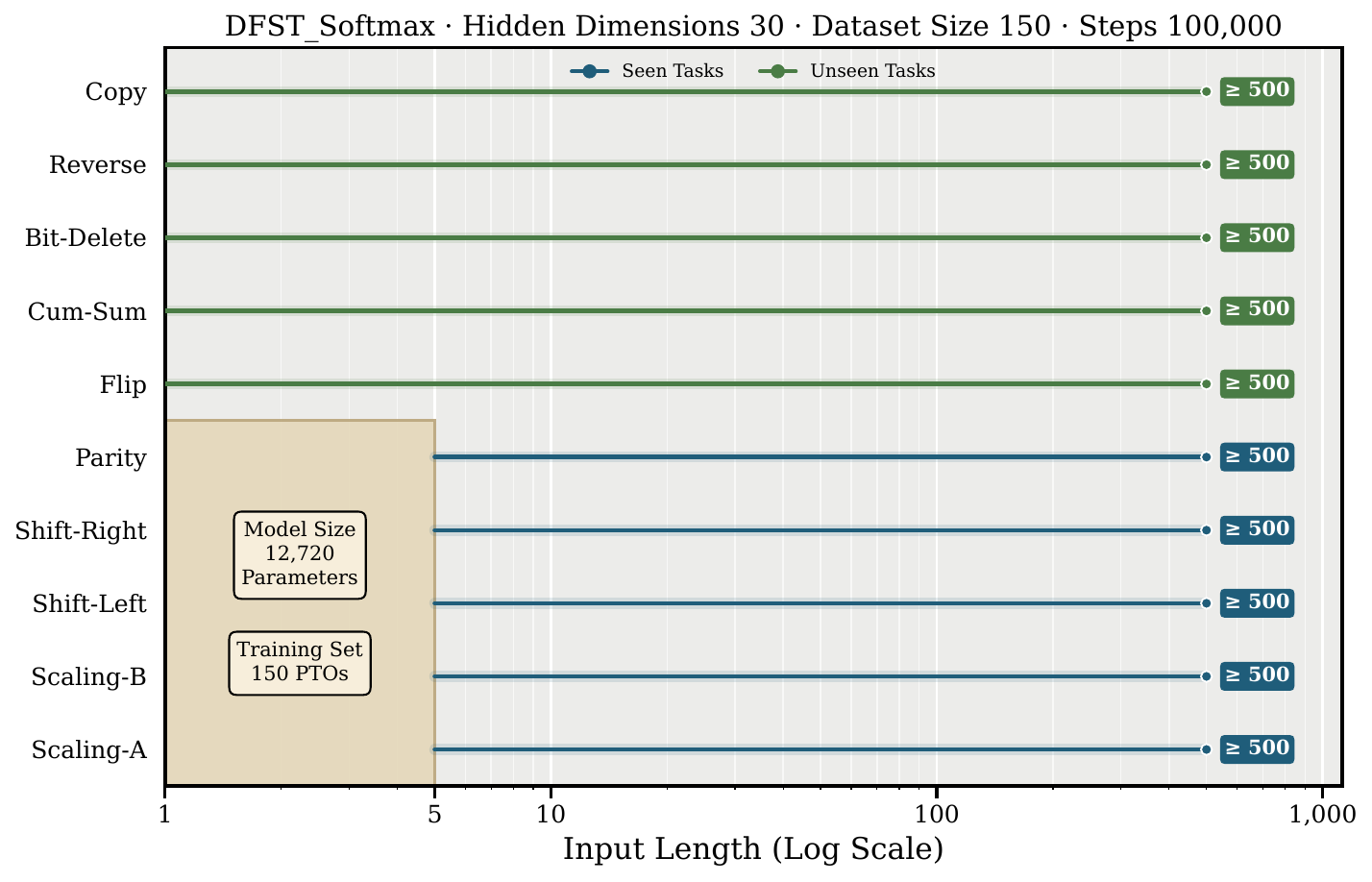}
\caption{Softmax DFST: $d=30$, $N=150$, 100k Updates.}
\end{subfigure}
\caption{\textbf{Generalization across Encoded Programs before Extraction.} Selected differentiably trained controllers reach the 500-symbol cap on all ten evaluation tasks. Training inputs contain at most five symbols. The scaling tests also increase the program parameter from 5 to 500.}
\label{fig:universal-current}
\end{figure}

\subsection{Generalization across Training Configurations}

\begin{figure}[t]
\centering
\begin{subfigure}{\linewidth}
\includegraphics[width=\linewidth]{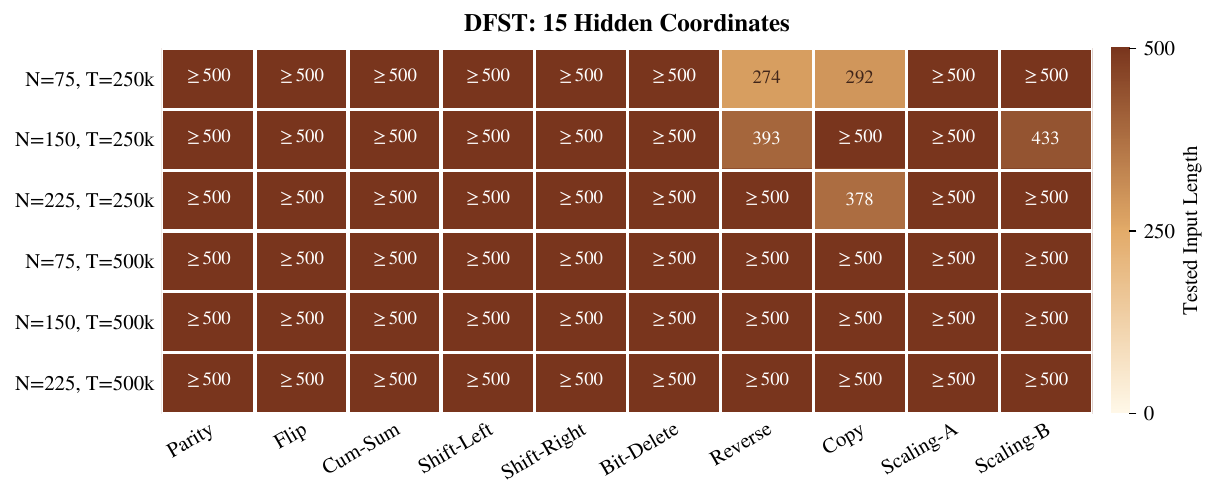}
\caption{Ordinary DFST: Dataset Size and Update Budget.}
\end{subfigure}\par\medskip
\begin{subfigure}{\linewidth}
\includegraphics[width=\linewidth]{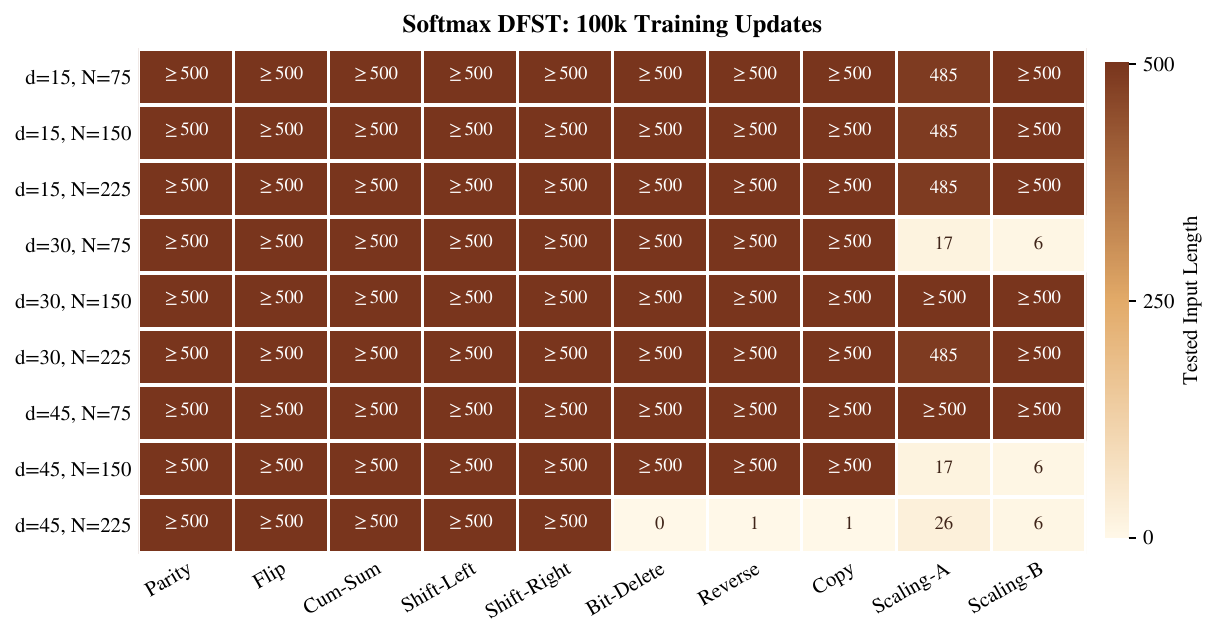}
\caption{Softmax DFST: Hidden Dimension and Dataset Size.}
\end{subfigure}
\caption{\textbf{Complete Universal-Simulation Sweeps before Extraction.} Each cell reports the tested input length under exact PTO agreement. The notation $\geq500$ marks the evaluation cap. Training inputs contain at most five symbols; $N$ counts PTOs, $d$ denotes hidden dimension, and $T$ counts training updates. Scaling tests increase the encoded machine size together with input length.}
\label{fig:universal-sweeps}
\end{figure}

\Cref{fig:universal-sweeps} collects every evaluated universal configuration across all ten tasks. All three ordinary-DFST runs with 500k updates reach the 500-symbol cap on every task. At 250k updates, the runs reach the cap on eight, eight, and nine tasks for $N=75,150,225$, respectively. The remaining failures involve reversal, copying, or the second scaling family.

For Softmax DFSTs, two of nine configurations reach the cap on all ten tasks after 100k updates: $(d,N)=(30,150)$ and $(45,75)$. Several other configurations succeed on all eight fixed programs but fail on at least one scaling family. Thus generalization across fixed programs does not guarantee extrapolation in encoded machine size, and increasing dimension or dataset size does not uniformly improve the outcome.

\subsection{PTO Lengths at the Evaluation Caps} \label{app:pto-lengths}

Operand length understates the execution horizon. Direct replay of the expert transition maps gives 50\,070\,010 PTO steps for adding two 5000-digit copies of the maximal digit and 15\,012\,991 steps for multiplying two 1000-digit copies, in both radix 2 and radix 10. These counts include exactly one stopping action. The maximal-digit examples belong to the evaluated repdigit family. The counts describe expert trajectories used by the cap probes; the counts do not constitute additional neural-model evaluations.

For universal simulation, replaying the ten deterministic evaluation inputs at length 500 for each task gives the maxima below. The longest sampled trajectory contains 15\,316\,822 steps, for copying. Thus a successful 500-symbol probe can require more than fifteen million consecutive action matches.

\begin{table}[h!]
\centering
\caption{Maximum Expert PTO Length among the 10 Seeded 500-Symbol Evaluation Inputs per Universal Task. Each count includes the final stopping action once.}
\label{tab:pto-horizons}
\small
\begin{tabular}{lr@{\qquad}lr}
\toprule
Task & PTO steps & Task & PTO steps \\
\midrule
\texttt{parity} & 9295 & \texttt{flip} & 9295 \\
\texttt{cum-sum} & 9298 & \texttt{bit-delete} & 24\,838 \\
\texttt{shift-left} & 41\,806 & \texttt{shift-right} & 54\,086 \\
\texttt{reverse} & 14\,911\,144 & \texttt{copy} & 15\,316\,822 \\
\texttt{scaling-A}(500) & 25\,784 & \texttt{scaling-B}(500) & 427\,803 \\
\bottomrule
\end{tabular}
\end{table}

%% file: Appendix/Extraction.tex
\section{Extracting Interpretable Finite-State Controllers}
\label{app:extraction}

A trained recurrent controller already induces a finite-state machine under fixed finite-precision execution. Extraction seeks a compact description of the learned policy, suitable for interpretation and mathematical verification. We use two complementary constructions: metric extraction groups hidden states with similar geometry; annealing encourages a Softmax DFST to approach deterministic transitions. Neither construction uses expert hidden-state labels. The equivalence proofs in \Cref{app:exact} concern the resulting discrete controllers. Transferring those proofs to the neural controllers requires the additional conditions stated below.

Executable notebooks accompany both \href{https://github.com/fanagor/DEL/blob/main/notebooks/metric_FST_extraction.ipynb}{metric extraction} and \href{https://github.com/fanagor/DEL/blob/main/notebooks/anneal_dfst_softmax_to_fst.ipynb}{annealing}, with saved cases and instructions for analyzing another trained checkpoint.

\subsection{Metric Extraction from DFST Trajectories}
\label{app:metric-extraction}

\myparagraph{Autonomous Trajectory Collection.}
Fix a trained controller with hidden-state update $f_x$ and decoded action $g_x$, where $x\in\Sigma$ and $g_x(h)\in\Sigma\times\M$. For a DFST, $f_x(h)=A[x]h$ and
\begin{equation*}
 g_x(h)=\bigl(\argmax B[x]h,\;\argmax C[x]h\bigr).
\end{equation*}
The readouts use the hidden state \emph{before} the transition. From each sampled valid initial environment $z$, run the neural controller autonomously: the predicted write and movement determine the next observation. Record
\begin{equation*}
 (h_{t-1},x_t,s_t,m_t,h_t),\qquad 1\leq t\leq L(z).
\end{equation*}
Thus extraction observes the controller's actual policy rather than a teacher-forced continuation after an erroneous action. A stopping action contributes the final PTO triple exactly once. Nonfinite states and runs exceeding the collection budget trigger failure rather than a fictitious halt. The metric case studies use the unrescaled sequential DFST recurrence during collection; any transfer claim refers to the same arithmetic and execution convention.

\myparagraph{Geometric Grouping and Deterministic Closure.}
For a finite trajectory collection, connect two nonterminal hidden-state occurrences whenever the Euclidean distance does not exceed $\eps$. Take the connected components of this proximity graph. Connectedness permits chains of nearby points; a component need not have diameter at most $\eps$.
For each component $q$ and observation $x$, inspect all recorded transitions. Conflicting write--movement pairs invalidate the threshold. Otherwise, merge the successor components whenever the same pair $(q,x)$ reaches several successors, and repeat the consistency check and successor merging. Keep a distinguished halt state separate from nonterminal components; a required halt/nonhalt merger also invalidates the threshold. Every successful merger reduces the finite component count, so closure either terminates with a deterministic graph or detects a conflict. In the accepted graph $G(S,\eps)$, an unobserved state--symbol pair remains undefined.

Threshold selection uses the complete extraction sample before measuring graph stability. Starting from a proposed $\eps$, halve the threshold after a conflict and restart from the original observations. The implementation permits 65 positive-threshold attempts followed by a zero-threshold attempt; an unresolved conflict aborts extraction. A successful threshold supplies consistency with the collected trajectories, without certifying behavior beyond those trajectories.

\myparagraph{Exact Quotient Reduction.}
Redirect stopping transitions to a single halt node, preserving the written symbol. Conventional halt self-loops only describe behavior after execution has stopped. Apply Hopcroft partition refinement to the resulting partial Mealy transducer, distinguishing final status, output labels, and definedness. Two states can share a quotient class only when every symbol has matching definedness and, for each defined transition, matching outputs and successor classes. The quotient therefore preserves all defined traces and all missing transitions. This reduction differs from completing unknown transitions or merging states merely because the observed samples fail to distinguish the states. The case studies additionally check the quotient map transition by transition and replay every extraction trajectory.

\begin{table}[t]
\centering
\small
\setlength{\tabcolsep}{5pt}
\begin{tabular}{lrrrrr}
\toprule
Task & Inputs & $\eps$ & Raw states & Raw edges & Quotient states \\
\midrule
\texttt{add2} & 100 & $0.5$ & 23 & 44 & 22 \\
\texttt{mult2} & 100 & $0.125$ & 70 & 133 & 61 \\
\texttt{add10} & 250 & $0.5$ & 52 & 223 & 52 \\
\texttt{mult10} & 2,000 & $0.25$ & 195 & 1,282 & 182 \\
\texttt{universal} & 100 & $0.5$ & 49 & 78 & 44 \\
\bottomrule
\end{tabular}
\caption{\textbf{Metric Extraction Case Studies.} Raw counts describe the final deterministic graph before quotient reduction and before conventional halt loops. State counts include the halt node. Arithmetic extraction samples use operands of at most 10 digits for addition and five digits for binary multiplication. Decimal multiplication combines 355 random inputs with operands of at most two digits and 1,645 protocol-covering teaching inputs with operands of at most five digits; these inputs support extraction and do not constitute additional gradient-training examples. Universal extraction uses length-five inputs across all ten evaluation task families.}
\label{tab:metric-extraction}
\end{table}

\begin{figure}[p]
\centering
\begin{subfigure}{0.49\linewidth}
\includegraphics[width=\linewidth]{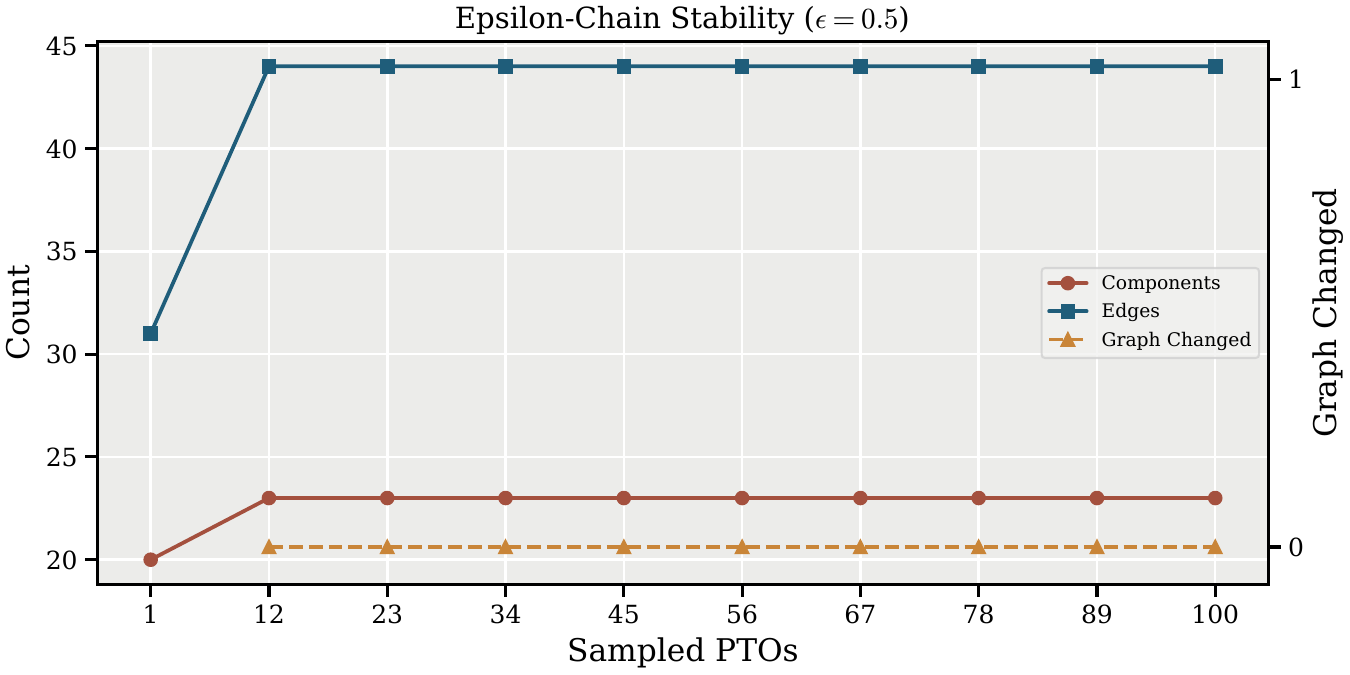}
\caption{Binary Addition}
\end{subfigure}\hfill
\begin{subfigure}{0.49\linewidth}
\includegraphics[width=\linewidth]{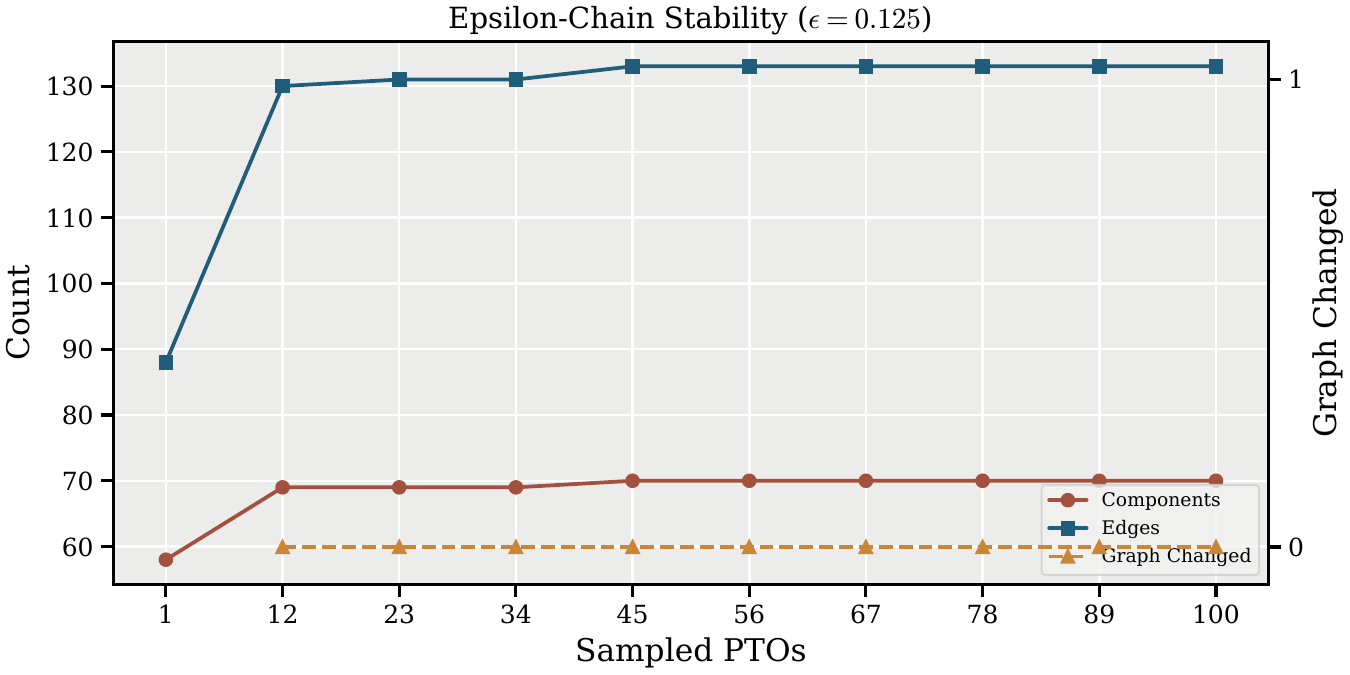}
\caption{Binary Multiplication}
\end{subfigure}
\medskip
\begin{subfigure}{0.49\linewidth}
\includegraphics[width=\linewidth]{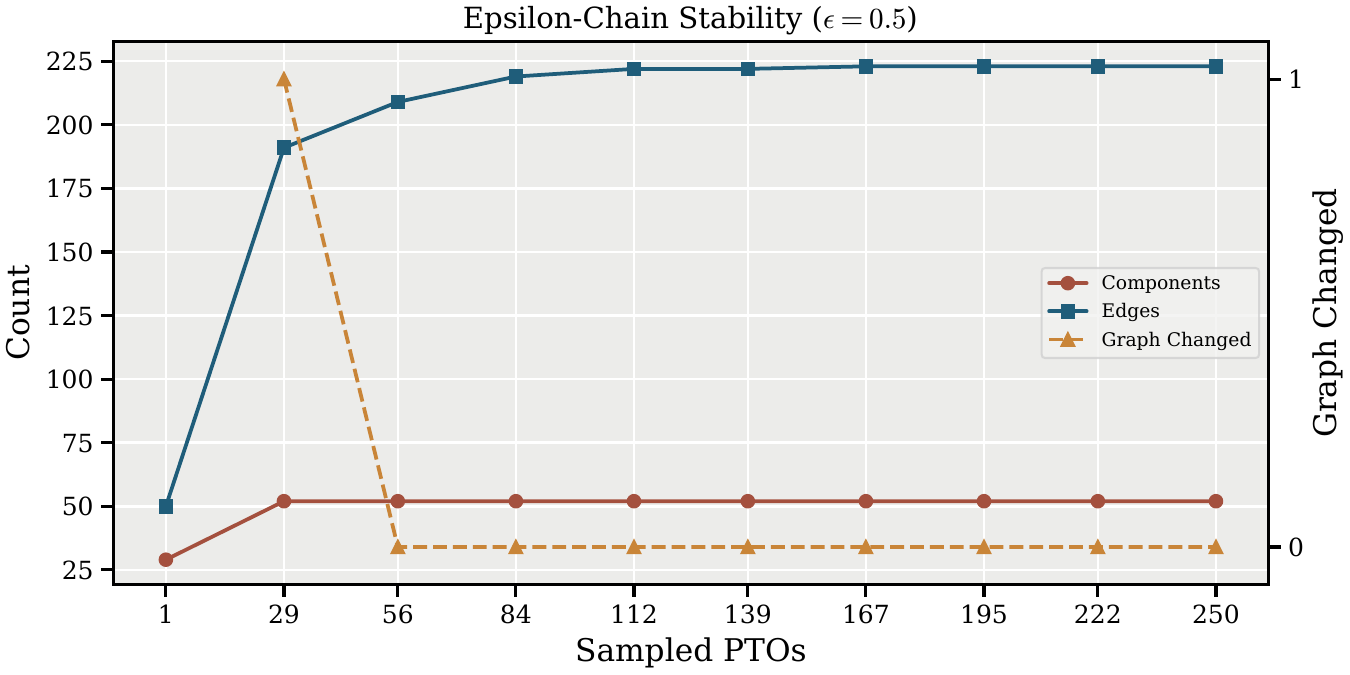}
\caption{Decimal Addition}
\end{subfigure}\hfill
\begin{subfigure}{0.49\linewidth}
\includegraphics[width=\linewidth]{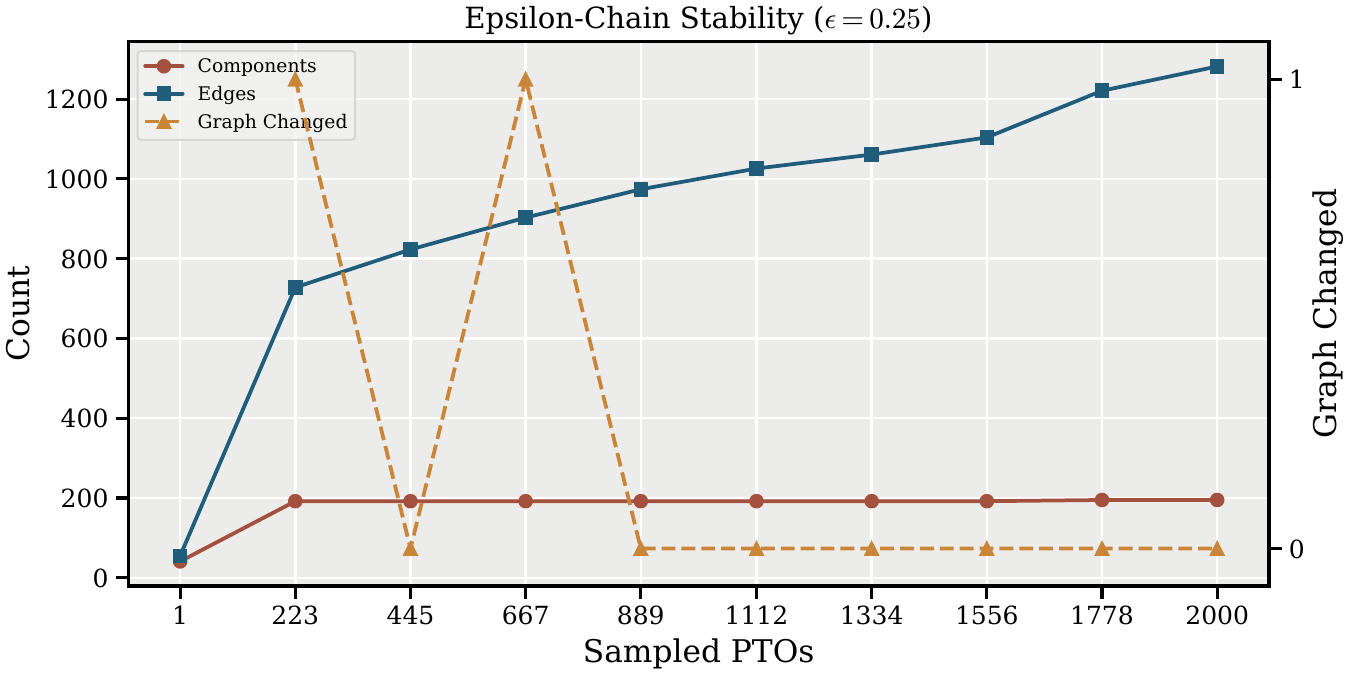}
\caption{Decimal Multiplication}
\end{subfigure}
\medskip
\begin{subfigure}{0.49\linewidth}
\includegraphics[width=\linewidth]{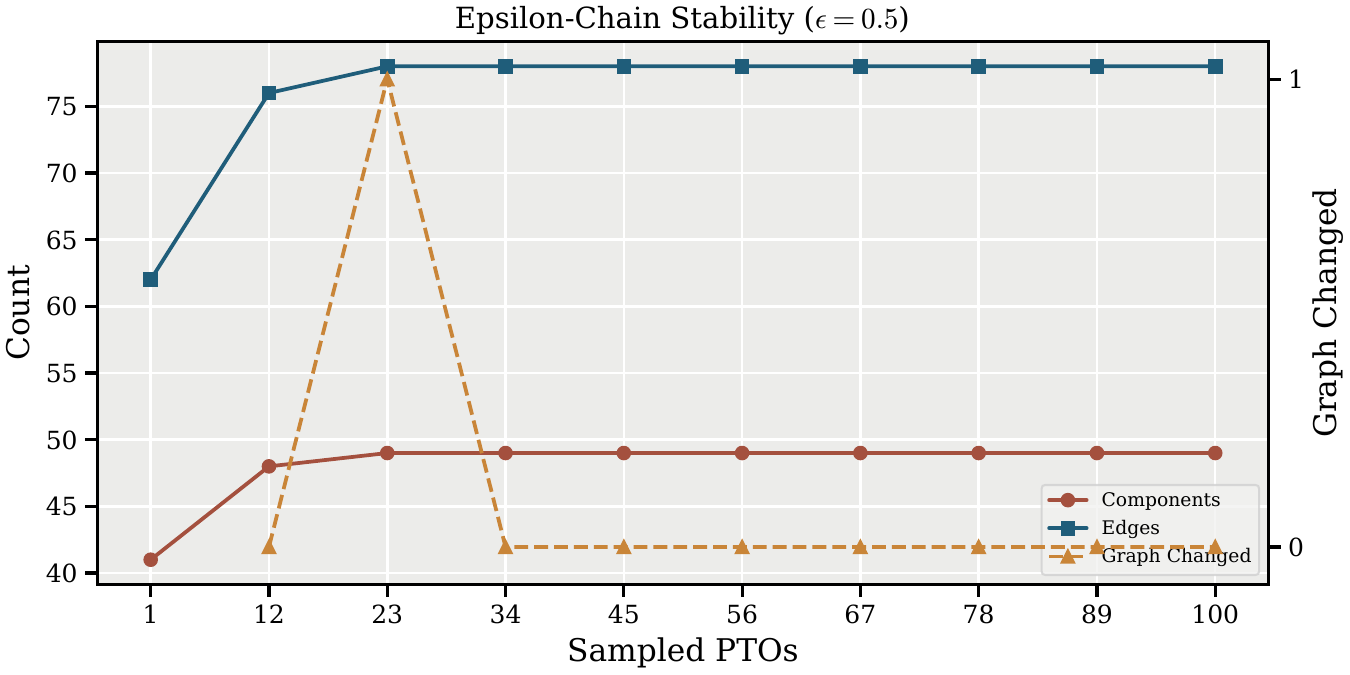}
\caption{Universal Controller}
\end{subfigure}
\caption{\textbf{Empirical Graph Stability under Metric Extraction.} The plots track components, observed edges, and changes as additional autonomous trajectories enter the extraction sample. The graph comparison also checks full rooted labeled structure independently of state numbering. Late plateaus support the geometric hypothesis on the sampled inputs; \Cref{ass:stability} quantifies over all valid executions.}
\label{fig:metric-stability}
\label{fig:graph-stability-full}
\end{figure}

\myparagraph{What Graph Stability Requires.}
At the accepted threshold, add trajectories one at a time and compare the rooted, input/output-labeled graphs up to state renaming. The comparison preserves the initial state, halt status, missing edges, and every transition label. Equal numbers of nodes and edges alone do not establish graph identity. In the five case studies, the final unchanged suffix begins at sample 4, 42, 143, 1,870, and 23 for \texttt{add2}, \texttt{mult2}, \texttt{add10}, \texttt{mult10}, and \texttt{universal}, respectively. \Cref{fig:metric-stability} displays the corresponding finite observations.

\Cref{ass:stability} requires substantially more: every finite enlargement of the sample by valid executions must preserve the same rooted labeled graph. For a nonhalting universal computation, the requirement extends to every finite execution prefix. A finite plateau provides evidence for this assumption, never a proof. Under the assumption, every valid neural trajectory follows a path already represented by the stable graph. Trace-preserving quotient reduction then transfers the neural policy to the compact controller, as formalized in \Cref{thm:faithfulness}. The assumption must refer to the deployed recurrence, rounding convention, and state normalization; changing those conventions requires a corresponding argument.

\subsection{Annealing and Argmax Extraction from Softmax DFSTs}
\label{app:softmax-extraction}

\myparagraph{A Differentiable Probabilistic Parametrization.}
The model \texttt{DFST\_Softmax} replaces unconstrained DFST tensors by column-stochastic tensors. For trainable logits $\Theta^A,\Theta^B,\Theta^C$ and $\theta^h$, define
\begin{align}
 A[x]_{ij}&=\frac{\exp(\Theta^A[x]_{ij})}{\sum_{k=1}^{d}\exp(\Theta^A[x]_{kj})}, &
 h_{0,i}&=\frac{\exp(\theta^h_i)}{\sum_{k=1}^{d}\exp(\theta^h_k)},\label{eq:softmax-dfst-A}\\
 B[x]_{sj}&=\frac{\exp(\Theta^B[x]_{sj})}{\sum_{u\in\Sigma}\exp(\Theta^B[x]_{uj})}, &
 C[x]_{mj}&=\frac{\exp(\Theta^C[x]_{mj})}{\sum_{v\in\M}\exp(\Theta^C[x]_{vj})}.
\end{align}
The case studies use a fixed one-hot input embedding. In exact arithmetic, $h_t=A[x_t]h_{t-1}$ remains in the probability simplex, and $B[x_t]h_{t-1}$ and $C[x_t]h_{t-1}$ give distributions over writes and movements. Execution selects the largest coordinate of each readout. Softmax changes the parameterization, while the recurrence still admits the associative matrix scan used for DFST training. A deterministic FST corresponds to simplex vertices: one-hot initial state and one-hot columns of all three tensors.

\myparagraph{Differentiable Concentration toward Deterministic Transitions.}
After ordinary PTO training, continue optimizing the same supervised objective with a concentration penalty. With $K=|\Sigma|$, define
\begin{equation}
 \mathcal R(\theta)=
 \sum_{P\in\{A,B,C\}}\frac{1}{Kd}\sum_{x\in\Sigma}\sum_{j=1}^{d}
 \left(1-\sum_i P[x]_{ij}^{\,2}\right)
 +1-\sum_i h_{0,i}^{\,2}.
 \label{eq:annealing-penalty}
\end{equation}
Each summand vanishes at a simplex vertex and penalizes distributed probability mass. At update $k=1,\ldots,N$, optimize
\begin{equation}
 \mathcal L_{\mathrm{PTO}}(\theta)+\lambda_k\mathcal R(\theta),\qquad
 \lambda_k=\lambda_{\mathrm{start}}
 \left(\frac{\lambda_{\mathrm{end}}}{\lambda_{\mathrm{start}}}\right)^{(k-1)/(N-1)}.
 \label{eq:annealing-schedule}
\end{equation}
All three case studies start from a 100,000-update checkpoint and use a fresh Adam optimizer at constant learning rate $10^{-3}$ for $N=5,000$ additional updates, with $\lambda_{\mathrm{start}}=10^{-5}$ and $\lambda_{\mathrm{end}}=10^5$. The continuation retains the training data, supervised loss, clipping, and sampling settings. Thus annealing strengthens a differentiable penalty, rather than imposing a hard state assignment during optimization. The phrase \emph{exponential annealing} refers to the regularization schedule, not to a softmax-temperature schedule.

\myparagraph{Reading Out the Discrete Controller.}
For each hidden coordinate $j$ and symbol $x$, set
\begin{equation}
 \begin{aligned}
 \delta(j,x)&=\argmax_i A[x]_{ij}, &
 w(j,x)&=\argmax_s B[x]_{sj},\\
 m(j,x)&=\argmax_v C[x]_{vj}, &
 q_0&=\argmax_i h_{0,i}.
 \end{aligned}
 \label{eq:argmax-extraction}
\end{equation}
Fix deterministic tie breaking. Retain states reachable from $q_0$, introduce the halt convention, and apply the same exact quotient reduction as above. The extracted transition reads the pre-transition coordinate $j$, matching the neural recurrence. Unlike metric extraction, argmax extraction defines an action for every symbol at each retained state. The final minimization yields 17 states for binary addition, 52 for binary multiplication, and 28 for universal simulation, including halt states. Argmax supplies a deterministic decoder of differentiably learned probabilities; the decoder itself does not supply gradients or an equivalence proof.

\begin{figure}[p]
\centering
\begin{subfigure}{\linewidth}
\includegraphics[width=\linewidth]{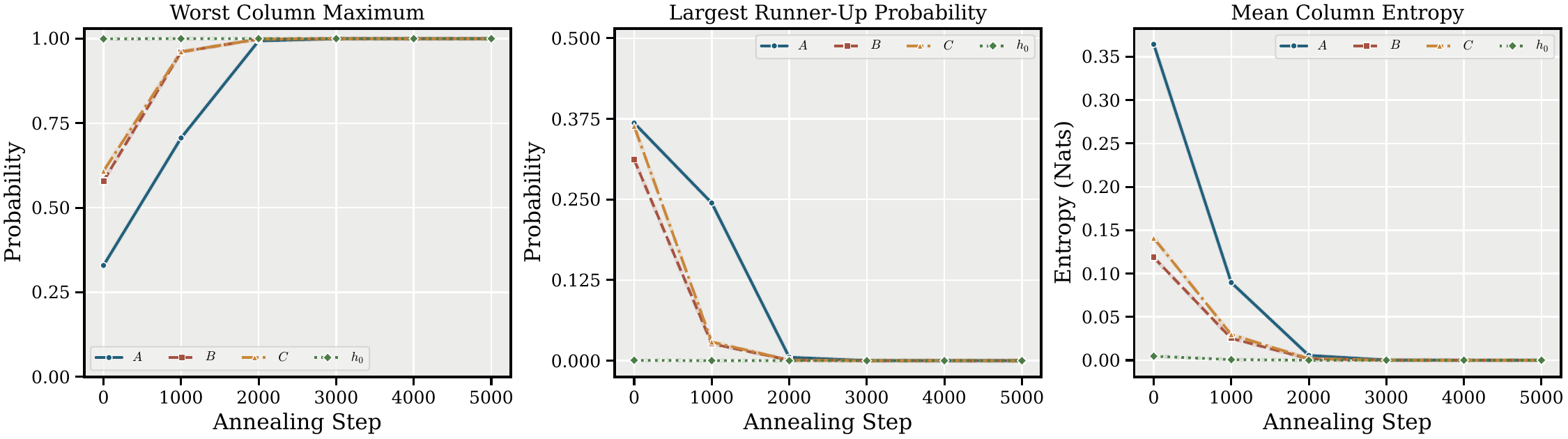}
\caption{Binary Addition}
\end{subfigure}\medskip

\begin{subfigure}{\linewidth}
\includegraphics[width=\linewidth]{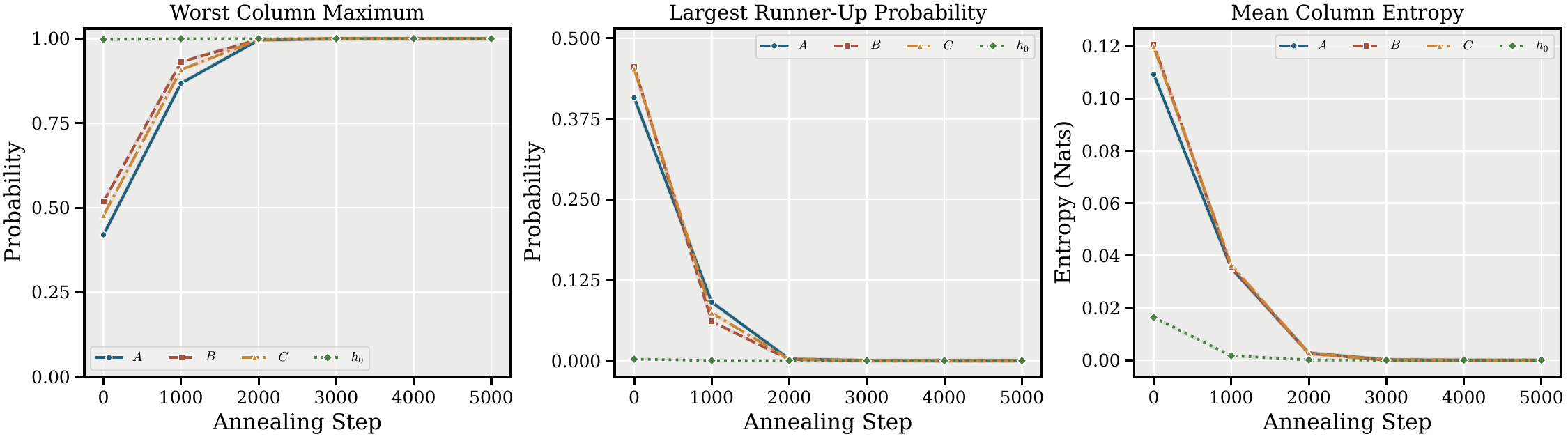}
\caption{Binary Multiplication}
\end{subfigure}\medskip

\begin{subfigure}{\linewidth}
\includegraphics[width=\linewidth]{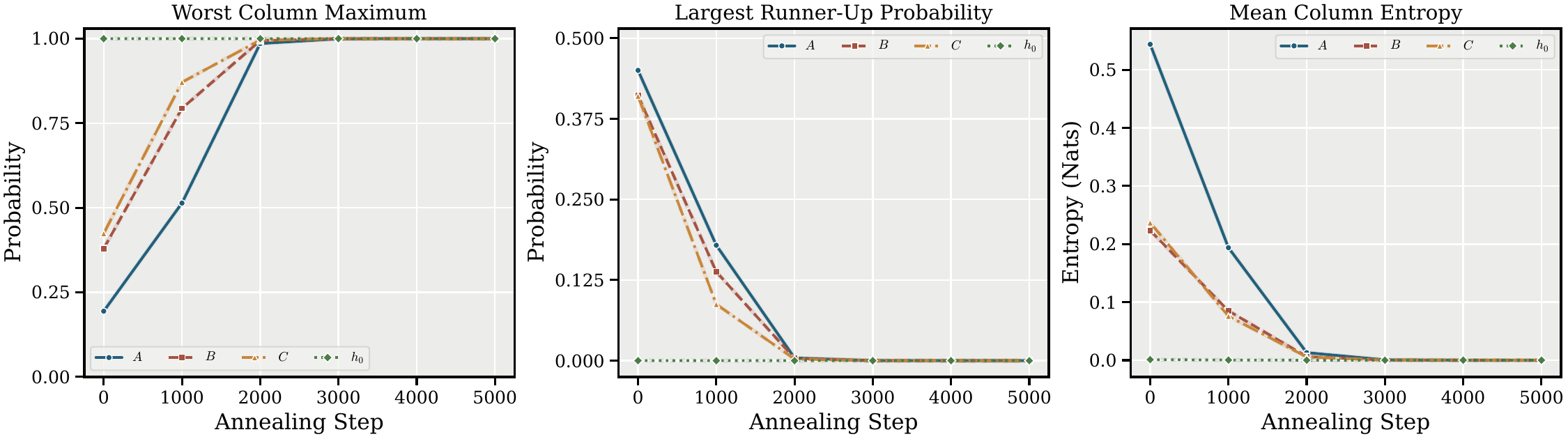}
\caption{Universal Controller}
\end{subfigure}
\caption{\textbf{Concentration during Differentiable Annealing.} Exponentially increasing the concentration penalty sharpens the transition and readout columns. At the final update, the smallest column maximum across $A,B,C,h_0$ exceeds $1-1.2\cdot10^{-7}$ for both arithmetic controllers and $1-2.4\cdot10^{-7}$ for the universal controller. These finite-precision measurements demonstrate near-deterministic parameters, but do not establish equivalence with the extracted FST for unbounded execution lengths.}
\label{fig:annealing-sharpness}
\end{figure}

\myparagraph{Uniform Leakage and Readout Margins.}
Fix a deterministic recurrent implementation, including rounding, and the effective readout tensors $B,C$ with entries in $[0,1]$. Let $\widehat T$ denote the unminimized columnwise-argmax controller before adjoining the conventional halt node. Each state $q$ corresponds to the one-hot vector $e_q$; motion $S$ terminates execution. Fix tie breaking and the valid input domain $\mathcal Z$.

For each $z\in\mathcal Z$, run $\widehat T$ and record the pre-action states $q_t(z)$ and observations $x_{t+1}(z)$. Drive the implemented neural recurrence from the trained initial state with this same observation stream, obtaining $h_t(z)$. The index $t$ includes every pre-action state through the stopping action, or every finite time for a nonhalting execution. Define the hidden-state leakage by
\begin{equation}
 \ell_t(z)=\|h_t(z)-e_{q_t(z)}\|_1
 =|h_{t,q_t(z)}(z)-1|+\sum_{j\ne q_t(z)}|h_{t,j}(z)|.
 \label{eq:softmax-leakage}
\end{equation}
This definition controls both off-target mass and drift of the intended coordinate, without assuming that rounded states remain exactly normalized.

For $P\in\{B,C\}$, write $a_P(q,x)=\argmax_i P[x]_{iq}$ for the prescribed write or motion. Over state--symbol pairs occurring on valid executions of $\widehat T$, define the minimum readout margin
\begin{equation}
 \mu=\min_{(q,x),\,P\in\{B,C\}}\left(P[x]_{a_P(q,x),q}-\max_{b\ne a_P(q,x)}P[x]_{bq}\right).
 \label{eq:softmax-readout-margin}
\end{equation}
The minimum ranges over finitely many columns. A certified superset of the reachable pairs supplies a conservative lower bound. Let $\widetilde y^P_t(z)$ denote the implemented readout, while $P[x_{t+1}(z)]h_t(z)$ denotes the exact real matrix-vector product of the stored tensor and hidden-state values.

\begin{assumption}[Uniformly Bounded Leakage]\label{ass:softmax-leakage}
There exist constants $\varepsilon,\rho\geq0$, independent of input and execution length, with $\varepsilon+2\rho<\mu$, such that every $z\in\mathcal Z$ and every pre-action time $t$ satisfy
\begin{equation}
 \ell_t(z)\leq\varepsilon,\qquad
 \|\widetilde y^P_t(z)-P[x_{t+1}(z)]h_t(z)\|_\infty\leq\rho
 \quad\text{for }P\in\{B,C\}.
 \label{eq:softmax-leakage-bound}
\end{equation}
The constant $\rho$ bounds per-component readout rounding error; exact readout arithmetic permits $\rho=0$.
\end{assumption}

\begin{proposition}[Faithfulness under Bounded Leakage]\label{prop:softmax-leakage}
Under \Cref{ass:softmax-leakage}, the neural controller and $\widehat T$ produce identical PTOs on all valid inputs, including simultaneous stopping and every finite prefix of a nonhalting execution. Any trace-preserving quotient inherits this equivalence.
\end{proposition}
\begin{proof}
Fix $z,t$, put $q=q_t(z)$, $x=x_{t+1}(z)$, and $h=h_t(z)$, and consider either readout head $P$. Let $a=a_P(q,x)$ and let $b\ne a$ denote any competing action. Since the entries of $P[x]$ lie in $[0,1]$, the row difference satisfies $\|P[x]_{a:}-P[x]_{b:}\|_\infty\leq1$. Therefore
\begin{align*}
 (P[x]h)_a-(P[x]h)_b
 &=P[x]_{aq}-P[x]_{bq}
   +(P[x]_{a:}-P[x]_{b:})(h-e_q)\\
 &\geq\mu-\|h-e_q\|_1\geq\mu-\varepsilon.
\end{align*}
Readout rounding changes each score by at most $\rho$, so the implemented gap satisfies
\begin{equation*}
 \widetilde y^P_{t,a}(z)-\widetilde y^P_{t,b}(z)
 \geq\mu-\varepsilon-2\rho>0.
\end{equation*}
Thus both implemented heads choose the prescribed actions on the supplied observation stream.

Now compare autonomous executions. Initially, both controllers receive the same grid, head position, and rune records. If all earlier actions agree, both environments supply $x_{t+1}(z)$, and the deterministic neural implementation reaches the same $h_t(z)$ as on the supplied stream. The margin argument gives the same write and motion, preserving the environment correspondence. A stopping action terminates both executions; otherwise induction continues. This proves agreement through every finite prefix and simultaneous stopping. Trace-preserving quotient reduction retains every such PTO.
\end{proof}

Sharp readout columns allow a larger tolerance for hidden-state leakage. For example, if every prescribed entry exceeds $1-\beta$ and every competing entry remains below $\beta$, then $\mu>1-2\beta$. Uniform leakage bounds concern the entire recurrent execution, whereas column sharpness concerns individual parameters. Exact one-hot tensors and initialization satisfy \Cref{ass:softmax-leakage} with $\varepsilon=\rho=0$ whenever the implemented arithmetic preserves vertex operations.

\myparagraph{Evidence and Proof Play Different Roles.}
The saved finite checks for metric binary addition, binary multiplication, decimal addition, and universal simulation, and for all three annealed controllers, each report 100 exact PTO matches on 100 inputs of length 100. Arithmetic checks use two length-100 operands; universal checks sample length-100 binary words and task instances. These checks compare the \emph{extracted} controller against the expert. The neural length-generalization experiments in \Cref{sec:exp} provide separate evidence before extraction. Neither evidence source replaces the inductive proofs in \Cref{app:exact}, and neither proves \Cref{ass:stability,ass:softmax-leakage}. The eight proofs establish exact algorithmic learning for the extracted controllers; the stated graph-stability or bounded-leakage assumptions transfer the corresponding conclusion to the trained neural controllers.

%% file: Appendix/Proving_Exact_Learning.tex
\section{Proving Exact Learning}\label{app:exact}

\myparagraph{LLM Assistance.}
We used GPT-6 Astra to generate the controller-equivalence proofs and verified the arguments ourselves. Executable certificate checkers validate the finite transition obligations; the unbounded equivalence claims additionally rely on the mathematical invariants and induction arguments presented below. We take responsibility for the correctness of the final statements and proofs.

We separate three claims: equality of sampled PTOs, exact equivalence of an extracted controller, and transfer of that equivalence to the trained neural controller. The proofs below establish the second claim without a length bound. Transfer requires the graph-stability or bounded-leakage assumptions discussed in \Cref{sec:exact-learning}; finite experiments alone establish neither assumption.

The \href{https://github.com/fanagor/DEL/tree/main/analyses}{public proof bundles} supply machine-readable transition maps and executable checks of the finite certificate obligations. The arguments below establish the unbounded claims independently of the accompanying code.

\subsection{Why Finite Testing Cannot Certify General Equivalence}\label{app:finite-precision}

With fixed weights and deterministic arithmetic, $D$ recurrent scalars stored in $b$ bits induce at most $2^{bD}$ memory encodings. Restricting each scalar to a finite IEEE binary32 value gives $(2^{32}-2^{24})^D$ encodings, counting signed zeros separately. Here $D=d$ for DFSTs, Softmax DFSTs, RNNs, and GRUs. This enormous FST remains difficult to inspect; ordinary FST equivalence on arbitrary read strings does not resolve equivalence under an unbounded environment.

\begin{proof}[Proof of \Cref{lem:undecidable}]
Suppose an algorithm computes the claimed horizon. Given a Turing machine $N$ and input $w$, construct two grid agents that simulate exactly $|z|$ steps of $N(w)$ on input $z$, using identical visible actions throughout the simulation. If the simulation detects halting, the agents emit different final actions; otherwise both agents stop with the same PTO. Every such execution terminates. The agents agree on every input precisely when $N(w)$ never halts. A computed horizon would reduce this universal equivalence claim to finitely many terminating comparisons, deciding nonhalting and therefore halting. This contradiction proves the claim. The obstruction concerns controllers interacting with an unbounded environment; ordinary finite-state transducer equivalence on unrestricted read strings remains decidable.
\end{proof}

\subsection{Transferring an Extracted Certificate to a Trained Controller}\label{app:faithfulness-proof}

Let $G(S,\eps)$ denote the deterministic, trace-consistent graph extracted from valid neural executions $S$ at threshold $\eps$, including finite prefixes of nonhalting executions.

\begin{assumption}[Graph Stability] \label{ass:stability}
For some $\eps>0$ and finite sample $S$, every finite extension $S'$ by valid neural execution traces or prefixes yields a successful extraction and a graph $G(S',\eps)$ isomorphic to $G(S,\eps)$, preserving the root, read symbols, output labels, and stopping actions.
\end{assumption}

\begin{proof}[Proof of \Cref{thm:faithfulness}]
For metric extraction, fix a valid input and any finite execution prefix of the trained controller. Extend the extraction data by that prefix. Trace consistency embeds every observed transition into the resulting graph. Graph stability identifies the extended rooted, input/output-labeled graph with the original graph, including the initial node and stopping convention. Thus the original graph reproduces every action of the prefix. Induction synchronizes the controller and graph: equal grids and head positions provide equal read symbols; equal output labels produce equal writes and motions and preserve the next configuration. Rune records, when present, evolve identically under the same actions. Since the argument permits every prefix length, the complete finite PTOs agree, or every finite prefix agrees for a nonterminating execution. Any subsequent merging that preserves all defined traces preserves this conclusion. Composing this equality with an extracted-controller certificate proves equivalence to the expert. Equality of node or edge counts alone would not supply the labeled-graph identity used in this argument. For Softmax DFSTs, \Cref{prop:softmax-leakage} establishes equality of every finite PTO prefix under \Cref{ass:softmax-leakage}. Trace-preserving reduction and composition with the expert-equivalence certificate give the same conclusion.
\end{proof}

\subsection{A Common Simulation Principle}\label{app:simulation-principle}
\begingroup
\providecommand{\blank}{\lambda}
\providecommand{\digits}{\{0,1\}}
\providecommand{\EXPERT}{\mathcal E}
\providecommand{\CTRL}{\mathcal C}
\providecommand{\PTO}{\operatorname{PTO}}
\providecommand{\triple}[3]{(#1,#2,\mathsf{#3})}
\providecommand{\ph}[1]{\mathsf{#1}}
\providecommand{\code}[1]{\texttt{#1}}

Write $\EXPERT$ for an expert and $\CTRL$ for an extracted controller. A transition $\delta(q,x)=(q',w,M)$ reads $x$, writes $w$, performs motion $M$, and enters state $q'$. Motion $S$ writes and stops without moving. PTOs contain read/write/motion triples and count the final stopping action once; halt loops contribute no additional observations. Arithmetic inputs occupy row $0$ from column $0$, with all other cells blank and the head initially at $(0,0)$. Arithmetic coordinates increase rightward and downward. Operand words may have unequal positive lengths and leading zeros.

A \emph{phase} records the expert control state together with finite geometric or arithmetic context. A phase need not correspond to a distinct machine state. The following lemma explains why a finite certificate can establish equality of arbitrarily long executions.

\begin{lemma}[Phase Simulation]\label{lem:phase-simulation}
Suppose a protocol covers every valid expert execution. Assign expert state $g_i$ and a nonempty controller-state set $Q_i$ to each phase $i$. Assume the initial states belong to the initial phase relation. For each permitted edge $i\xrightarrow{x}j$ and every $q\in Q_i$, suppose
\begin{equation}\label{eq:phase-simulation}
\delta_{\EXPERT}(g_i,x)=(g_j,w,M),\qquad
\delta_{\CTRL}(q,x)=(q',w,M),\qquad q'\in Q_j.
\end{equation}
Assume stopping occurs exactly on edges to terminal phases. Then the two controllers produce identical PTOs on the valid input domain, with identical environment configurations at every step.
\end{lemma}
\begin{proof}
Relate configurations with the same grid, head, and, where applicable, rune records, and control states $g_i$ and $q\in Q_i$. Initially this relation holds. Protocol coverage provides an allowed read symbol; equal configurations provide the same symbol to both controllers. Equation~\eqref{eq:phase-simulation} gives the same write and motion and relates the successor states. The deterministic environment update preserves the relation. Induction proves equality of every finite prefix. The stopping condition gives simultaneous termination, or equal infinite PTOs. No undefined transition can occur along a valid execution.
\end{proof}

The tables below give the necessary transition maps on valid executions. An entry $x:(j,w,M)$ specifies a successor \emph{phase} $j$, with expert successor $g_j$ and controller successor in $Q_j$; ordinary transition tables instead list actual successor states. Multivalued $Q_i$ capture different internal realizations of the same phase. Each listed branch applies to every state in the corresponding set. Undefined or unlisted branches require no completion when the coverage proof excludes those branches. Local identities can therefore be checked by finite substitution, while the geometric invariants supply the unbounded claim.

\begin{center}
\small\setlength{\tabcolsep}{5pt}
\begin{tabular}{llrrrr}
\toprule
Task & Extraction & States & Phases & Related pairs & Local checks\\
\midrule
\texttt{add2} & Metric &22&27&28&50\\
 & Annealing &17&27&30&52\\
\texttt{mult2} & Metric &61&67&67&130\\
 & Annealing &52&94&109&201\\
\texttt{add10} & Metric &52&59&60&266\\
\texttt{mult10} & Metric &182&207&208&1,509\\
\texttt{universal} & Metric &44&92&96&160\\
 & Annealing &28&91&93&162\\
\bottomrule
\end{tabular}
\end{center}
State counts include the designated halt state. All eight finite certificates satisfy the local obligations. The following arguments establish coverage for every input in each stated domain.

\subsection{Binary Addition: Two Certificates, One Arithmetic Invariant}\label{app:proof-add2}
Let $D=\{0,1\}$ and consider $a+b$ with $a\in D^m$, $b\in D^n$, $m,n\ge1$. Write $A=\operatorname{val}_2(a)$ and $B=\operatorname{val}_2(b)$. The expert starts in state $0$ and halts in state $10$. The metric and annealed controllers start in state $0$ and halt in states $21$ and $16$, respectively.
\begin{theorem}[Exact Binary Addition]\label{thm:add2-equiv}
Both extracted controllers produce the expert PTO on every valid binary-addition input, with identical grids and head positions at every step.
\end{theorem}
\subsubsection{Shared Expert Behavior and Observation Protocol}
\input{Appendix/Proof_Tables/add2_shared}

\subsubsection{Metric-Extraction Certificate}
\input{Appendix/Proof_Tables/m_add2}

\subsubsection{Annealing Certificate}
\input{Appendix/Proof_Tables/a_add2}

\begin{proof}[Proof of \Cref{thm:add2-equiv}]
The common round invariant proves protocol coverage and termination. The two state relations separately satisfy \eqref{eq:phase-simulation}: the metric certificate contains $50$ local identities, and the annealing certificate contains $52$. Both relations contain the initial pair and terminate only after the common final $S$ action. Apply \Cref{lem:phase-simulation} to each relation. The final grid contains the padded representation of $A+B$ specified by \Cref{add2:lem:termination}.
\end{proof}

\subsection{Binary Multiplication: Shared Products and Refined Protocols}\label{app:proof-mult2}
Consider $a\,x\,b$ with $a\in\{0,1\}^m$, $b\in\{0,1\}^n$, $m,n\ge1$, where $x$ denotes the multiplication separator. Write $A=\operatorname{val}_2(a)$ and $B=\operatorname{val}_2(b)$. The expert has initial state $0$ and halt state $31$. Both extracted controllers start in state $0$; metric extraction yields halt state $60$, and annealing yields halt state $51$.
\begin{theorem}[Exact Binary Multiplication]\label{thm:mult2-equiv}
Both extracted controllers produce the expert PTO on every valid binary-multiplication input. Both executions terminate with the $(m+n)$-bit representation of $AB$, including leading-zero padding, on row $n-1$ in columns $-n,\ldots,m-1$. The final head stops at $(-n-1,n-1)$.
\end{theorem}
\subsubsection{Shared Expert Behavior}
\input{Appendix/Proof_Tables/mult2_shared}

\subsubsection{Metric-Extraction Certificate}
\input{Appendix/Proof_Tables/m_mult2}

\subsubsection{Annealing Certificate}
The annealed controller reuses internal states more extensively. The refined protocol remembers unchanged factor bits and distinguishes one-digit scans from longer scans; these annotations belong to the proof, without adding memory to either controller.
\input{Appendix/Proof_Tables/a_mult2}

\begin{proof}[Proof of \Cref{thm:mult2-equiv}]
The copy, partial-product, and accumulation invariants establish the claimed product and termination for the expert. The metric protocol covers every valid execution through $67$ phases; the refined annealing protocol covers every valid execution through $94$ phases. The respective transition maps satisfy all $130$ and $201$ local identities, including the initial and terminal conditions. Applying \Cref{lem:phase-simulation} separately to the two protocols proves both equivalence claims.
\end{proof}

\subsection{Decimal Addition}\label{app:proof-add10}
\def\digits{\{0,\ldots,9\}}
Let $D=\digits$ and consider $a+b$, with $a\in D^m$, $b\in D^n$ and $m,n\ge1$. Write $A=\operatorname{val}_{10}(a)$ and $B=\operatorname{val}_{10}(b)$. The expert and metric-extracted controller start in state $0$ and halt in states $18$ and $51$, respectively.
\begin{theorem}[Exact Decimal Addition]\label{thm:add10-equiv}
The metric-extracted controller produces the expert PTO on every valid decimal-addition input, with identical grids and head positions at every step.
\end{theorem}
\input{Appendix/Proof_Tables/add10}

\begin{proof}[Proof of \Cref{thm:add10-equiv}]
The accumulator invariant and the ten digit-transport families cover every possible decimal digit pair and carry. Every valid execution follows the $59$-phase protocol and terminates. The state relation satisfies all $266$ instances of \eqref{eq:phase-simulation}, including the initial pair and final stopping action. Thus \Cref{lem:phase-simulation} proves equality of the complete PTOs and final configurations.
\end{proof}

\subsection{Decimal Multiplication}\label{app:proof-mult10}
Let $D=\{0,\ldots,9\}$ and consider $a\,x\,b$, with $a\in D^m$, $b\in D^n$, $m,n\ge1$. Write $A=\operatorname{val}_{10}(a)$ and $B=\operatorname{val}_{10}(b)$. The expert and metric-extracted controller start in state $0$ and halt in states $107$ and $181$, respectively.
\begin{theorem}[Exact Decimal Multiplication]\label{thm:mult10-equiv}
The metric-extracted controller produces the expert PTO on every valid decimal-multiplication input. Both executions terminate with the $(m+n)$-digit decimal representation of $AB$, including leading-zero padding, on row $n-1$ in columns $-n,\ldots,m-1$. The final head stops at $(-n-1,n-1)$.
\end{theorem}
\input{Appendix/Proof_Tables/mult10}

\begin{proof}[Proof of \Cref{thm:mult10-equiv}]
The staircase, partial-product, accumulator, and boundary invariants establish coverage of the $207$-phase protocol for arbitrary operand lengths. The displayed state relation contains $208$ phase--state pairs and satisfies $1{,}509$ local identities. The protocol excludes every undefined controller transition and terminates after the common final action. Apply \Cref{lem:phase-simulation}; the accumulator invariant supplies the final padded product and head position.
\end{proof}

\subsection{Universal Computation: Two Certificates for One Interpreter}\label{app:proof-universal}
A simulated Turing machine has states $\{0,\ldots,N-1\}$, $N\ge1$, tape alphabet $\Gamma=\{\blank,0,1\}$, and a total transition function
\begin{equation*}
\delta:\{0,\ldots,N-1\}\times\Gamma\longrightarrow
\{0,\ldots,N-1\}\times\Gamma\times\{L,R,S\}.
\end{equation*}
The simulated machine starts in state $0$; motion $S$ stops after writing. The encoding below admits arbitrary program sizes. Both interpreter controllers start in state $0$; the expert, metric controller, and annealed controller have halt states $14$, $43$, and $27$, respectively.
\begin{theorem}[Exact Universal Interpretation]\label{thm:universal-equiv}
For every correctly encoded machine and every nonempty binary input, both extracted controllers produce the expert PTO. The annealed controller additionally covers empty binary input. At every step, the visible grids, heads, and rune records agree, including symbols hidden beneath runes. Both controllers and the expert halt simultaneously, or all corresponding finite execution prefixes agree forever.
\end{theorem}
\subsubsection{Shared Encoding and Expert Invariants}
\input{Appendix/Proof_Tables/universal_shared}

\subsubsection{Metric-Extraction Certificate: Nonempty Inputs}
\input{Appendix/Proof_Tables/m_universal}

\subsubsection{Annealing Certificate: Arbitrary Binary Inputs}
\input{Appendix/Proof_Tables/a_universal}

\begin{proof}[Proof of \Cref{thm:universal-equiv}]
The endpoint and prefix invariant keeps every program search inside the correct code-prefix interval. The no-missing-middle-trit lemma justifies the refined search branches. Both protocols therefore cover every finite expert execution prefix on the respective domains. The metric relation satisfies $160$ local identities, and the annealing relation satisfies $162$. Equal writes and motions preserve visible cells and hidden rune symbols simultaneously. The prefix form of \Cref{lem:phase-simulation} gives equality of complete finite PTOs and of every finite prefix for nonhalting computations. Faithful expert simulation identifies the common execution with the encoded Turing machine.
\end{proof}
\paragraph{Why the Metric Theorem Excludes Empty Input.}
The metric controller lacks an initial blank transition. Consider a one-state machine with $\delta(0,s)=(0,s,S)$ for every $s\in\Gamma$ and empty tape input. The expert starts with $(\blank,Z,X)$, whereas the metric controller cannot take the first step. Hence extending the metric theorem to empty input would give a false claim. Later blank readings remain covered. The annealed controller handles the initial blank and needs no such restriction.

\endgroup

%% file: Appendix/Proof_Tables/add2_shared.tex
\label{add2:sec:common2}
\paragraph{Accumulator and Marker Invariant}\label{add2:sec:part1}
Write $B=\sum_{j=0}^{n-1}b_j2^j$, with $b_0$ the rightmost digit.
The expert scans to the separator at column $m$, steps left and down,
writes a working $+$ below column $m-1$, and enters state $8$ above the marker.

\begin{lemma}[Round Invariant]\label{add2:lem:invariant}
After processing $k$ right-operand digits, $0\le k\le n$, the expert enters
state $8$ at $(p_k,0)$, where $p_k=m-1-k$. At this point:
\begin{itemize}
\item Row $1$ contains exactly one nonblank symbol: a $+$ at column $p_k$.
\item Row $0$ contains a contiguous accumulator on $[\ell_k,m-1]$, with value
$C_k=A+\sum_{j<k}b_j2^j$ and left edge $\ell_k\le p_k+1$.
Column $m-1$ carries the units digit.
\item The separator remains at column $m$. The unprocessed prefix of $b$
occupies columns $m+1,\ldots,m+n-k$. All remaining cells contain blanks.
\end{itemize}
\end{lemma}
\begin{proof}
Initialization gives $k=0$, $\ell_0=0$, $C_0=A$, and $p_0=m-1$.
State $8$ preserves the digit at $p_k$, or writes $0$ on a blank.
A blank at $p_k$ forces $\ell_k=p_k+1$, so zero extension preserves contiguity.

For $k<n$, the right scan reaches the blank after the unprocessed prefix,
steps left to digit $b_k$ at $r_k=m+n-k$, erases $b_k$, and descends.
The expert transports $b_k$ left along row $1$ to the marker. The distance satisfies
\begin{equation}
r_k-p_k=n+1\ge2. \label{eq:proof:add2.1}
\end{equation}
Consequently, the first two transport steps read blanks. At the marker, the
expert ascends and adds $b_k$ to the accumulator digit $e$ at column $p_k$.
If $e+b_k<2$, the expert writes $e+b_k$ and descends. Otherwise, the expert
writes $e+b_k-2$, moves left, and propagates one carry. Each digit $1$
changes to $0$; the first smaller digit increases by one, or a blank changes
to $1$. The update adds exactly $b_k2^k$ and preserves contiguity.

The expert returns along row $1$, erases the marker, steps left, writes the
next marker at $p_{k+1}=p_k-1$, and ascends into state $8$.
The earlier zero extension gives $\ell_{k+1}\le p_k=p_{k+1}+1$.
The accumulator now has value $C_{k+1}$, establishing the next round.
\end{proof}

\begin{lemma}[Termination and Final Configuration]\label{add2:lem:termination}
The expert terminates with only the padded base-$2$ representation of $A+B$
on row $0$, ending at column $m-1$. The final head stops at $(m-n-1,0)$.
Writing $\ell_2(s)$ for the binary digit count, with $\ell_2(0)=1$,
the output width equals $L=\max\{m,n+1,\ell_2(A+B)\}$.
\end{lemma}
\begin{proof}
Each round consumes one of the $n$ right-operand digits. Every scan crosses
a finite interval, and every carry reaches a digit below $1$ or a blank.
After round $n$, state $8$ ensures a digit at $p_n$. The next right scan
returns to the separator. State $3$ erases the separator and descends.
State $9$ walks left, erases the final marker, ascends, preserves the
accumulator digit, and stops. The invariant gives $C_n=A+B$, and cleanup
removes all work symbols.
The original $m$ columns persist, and marker extension reaches $p_n$,
ensuring at least $n+1$ digits. Every additional carry extension contributes
a digit required by the current numerical value. Partial sums never decrease,
so the final width equals $L$.
\end{proof}

\label{add2:sec:common3}
\label{add2:sec:protocol}
The following phases refine the expert states by geometric context.
The protocol covers every valid execution and may also admit additional
observation sequences. Write $d\in\digits$. An edge $x:j$ reads $x$ and
advances to phase $j$; $d:j$ permits either digit. Expert state $g_i$
supplies the write and motion from \Cref{m_add2:sec:tables}.
A dash denotes a terminal phase with no outgoing transition.

\begin{center}
\small\renewcommand{\arraystretch}{1.12}
\begin{tabular}{r l c l}
\toprule
$i$ & Phase meaning & $g_i$ & Allowed reading $:$ next phase\\
\midrule
0 & First input digit & 0 & $d:1$ \\
1 & Seek the separator & 0 & $d:1\quad +:2$ \\
2 & Initial descent & 1 & $d:3$ \\
3 & Place the marker & 1 & $\blank:4$ \\
4 & Ensure an accumulator digit & 8 & $d:6\quad \blank:5$ \\
5 & Digit after zero extension & 2 & $d:6$ \\
6 & Scan toward separator & 2 & $d:6\quad +:7$ \\
7 & First cell after separator & 2 & $d:8\quad \blank:22$ \\
8 & Scan remaining right operand & 2 & $d:8\quad \blank:9$ \\
9 & Read rightmost factor digit & 3 & $0:10\quad 1:14$ \\
10 & Transport zero: first blank & 4 & $\blank:11$ \\
11 & Transport zero: second blank & 4 & $\blank:12$ \\
12 & Transport zero: walk left & 4 & $\blank:12\quad +:13$ \\
13 & Add zero above marker & 4 & $d:21$ \\
14 & Transport one: first blank & 5 & $\blank:15$ \\
15 & Transport one: second blank & 5 & $\blank:16$ \\
16 & Transport one: walk left & 5 & $\blank:16\quad +:17$ \\
17 & Add one above marker & 5 & $0:21\quad 1:18$ \\
18 & Propagate a carry & 6 & $1:18\quad 0:19\quad \blank:19$ \\
19 & First return cell after carry & 7 & $\blank:20$ \\
20 & Return to the marker & 7 & $\blank:20\quad +:3$ \\
21 & At marker, without carry & 7 & $+:3$ \\
22 & Erase the input separator & 3 & $+:23$ \\
23 & First cleanup cell & 9 & $\blank:24$ \\
24 & Find and erase final marker & 9 & $\blank:24\quad +:25$ \\
25 & Preserve result and stop & 9 & $d:26$ \\
26 & Halted & 10 & $\text{--}$ \\
\bottomrule
\end{tabular}
\end{center}

\begin{lemma}[Protocol Coverage]\label{add2:lem:protocol}
Every execution of $\EXPERT$ from a valid input follows this protocol, starting
at phase $0$ and ending at phase $26$.
\end{lemma}
\begin{proof}
The nonempty first operand establishes phases $0$--$3$.
\Cref{add2:lem:invariant} places a digit immediately after a zero extension
in phase $4$, establishing phase $5$. The right scan meets the separator before any blank.
To the separator's right, a first blank means the next left step reads the separator
(phase $22$); after any digit, that step reads a digit (phase $9$).
Equation~\eqref{eq:proof:add2.1} guarantees two initial blank readings while transporting either
bit. The cell above the marker contains a digit. A carry ends strictly left
of the marker, so the first return cell contains a blank. A noncarry addition returns directly to
the marker. Erasing and shifting the marker reaches a blank. Finally,
cleanup starts below the separator on a blank, reaches the marker and then
a digit. These facts justify every branch of the table. Induction over the
steps and rounds proves coverage; \Cref{add2:lem:termination} supplies the final phase.
\end{proof}

%% file: Appendix/Proof_Tables/m_add2.tex
\label{m_add2:sec:common4}
\paragraph{Phase Relation and Local Checks}\label{m_add2:sec:part3}
For phase $i$, write $Q_i$ for the controller-state set in the following table.
The relation pairs an expert configuration in phase $i$ with a controller
configuration having the \emph{same grid and head} and a state in $Q_i$.

\begin{center}
\small\renewcommand{\arraystretch}{1.12}
\begin{tabular}{r l c l}
\toprule
Phase $i$ & Meaning & Expert state $g_i$ & Controller states $Q_i$\\
\midrule
0 & First input digit & 0 & $\{0\}$ \\
1 & Seek the separator & 0 & $\{1\}$ \\
2 & Initial descent & 1 & $\{2\}$ \\
3 & Place the marker & 1 & $\{3\}$ \\
4 & Ensure an accumulator digit & 8 & $\{4\}$ \\
5 & Digit after zero extension & 2 & $\{15\}$ \\
6 & Scan toward separator & 2 & $\{5\}$ \\
7 & First cell after separator & 2 & $\{6\}$ \\
8 & Scan remaining right operand & 2 & $\{6\}$ \\
9 & Read rightmost factor digit & 3 & $\{7\}$ \\
10 & Transport zero: first blank & 4 & $\{12\}$ \\
11 & Transport zero: second blank & 4 & $\{13\}$ \\
12 & Transport zero: walk left & 4 & $\{13\}$ \\
13 & Add zero above marker & 4 & $\{14\}$ \\
14 & Transport one: first blank & 5 & $\{8\}$ \\
15 & Transport one: second blank & 5 & $\{9\}$ \\
16 & Transport one: walk left & 5 & $\{9\}$ \\
17 & Add one above marker & 5 & $\{10\}$ \\
18 & Propagate a carry & 6 & $\{18\}$ \\
19 & First return cell after carry & 7 & $\{19,20\}$ \\
20 & Return to the marker & 7 & $\{20\}$ \\
21 & At marker, without carry & 7 & $\{11\}$ \\
22 & Erase the input separator & 3 & $\{7\}$ \\
23 & First cleanup cell & 9 & $\{16\}$ \\
24 & Find and erase final marker & 9 & $\{16\}$ \\
25 & Preserve result and stop & 9 & $\{17\}$ \\
26 & Halted & 10 & $\{21\}$ \\
\bottomrule
\end{tabular}
\end{center}

\textbf{Local Transition Identity.} For every listed $q\in Q_i$ and every allowed
edge $i\xrightarrow{x}j$ from the protocol, the transition maps satisfy
\begin{equation}
\delta_\EXPERT(g_i,x)=(g_j,w,M),\qquad
\delta_\CTRL(q,x)=(q',w,M),\qquad q'\in Q_j. \label{eq:proof:m-add2.2}
\end{equation}
Every required controller entry exists. A transition reaches the halt
state exactly when $j=26$, with motion $S$.
The certificate contains \textbf{28 phase--state pairs} (including the terminal pair)
and \textbf{50 local checks}. The checks cover 43 distinct nonterminal
controller entries. \Cref{m_add2:sec:tables} provides the tables for direct
inspection; direct substitution checks every instance of~\eqref{eq:proof:m-add2.2}.

\paragraph{Transition Tables}\label{m_add2:sec:tables}
Each entry lists $(\text{next state},\text{written symbol},\text{motion})$.
The symbol $\blank$ means blank; a dash means an undefined transition.
State numbers identify the extracted controllers throughout each certificate.

\textbf{Binary-Addition Expert.}\quad Initial state $0$; halt $10$.
\begin{center}
\footnotesize\renewcommand{\arraystretch}{1.12}
\setlength{\tabcolsep}{3pt}
\begin{tabular}{rcccc}
\toprule
State & $\blank$ & $+$ & $0$ & $1$\\
\midrule
0 & $\triple{10}{\blank}{S}$ & $\triple{1}{+}{L}$ & $\triple{0}{0}{R}$ & $\triple{0}{1}{R}$ \\
1 & $\triple{8}{+}{U}$ & $\triple{10}{+}{S}$ & $\triple{1}{0}{D}$ & $\triple{1}{1}{D}$ \\
2 & $\triple{3}{\blank}{L}$ & $\triple{2}{+}{R}$ & $\triple{2}{0}{R}$ & $\triple{2}{1}{R}$ \\
3 & $\triple{10}{\blank}{S}$ & $\triple{9}{\blank}{D}$ & $\triple{4}{\blank}{D}$ & $\triple{5}{\blank}{D}$ \\
4 & $\triple{4}{\blank}{L}$ & $\triple{4}{+}{U}$ & $\triple{7}{0}{D}$ & $\triple{7}{1}{D}$ \\
5 & $\triple{5}{\blank}{L}$ & $\triple{5}{+}{U}$ & $\triple{7}{1}{D}$ & $\triple{6}{0}{L}$ \\
6 & $\triple{7}{1}{D}$ & $\triple{10}{+}{S}$ & $\triple{7}{1}{D}$ & $\triple{6}{0}{L}$ \\
7 & $\triple{7}{\blank}{R}$ & $\triple{1}{\blank}{L}$ & $\triple{10}{0}{S}$ & $\triple{10}{1}{S}$ \\
8 & $\triple{2}{0}{R}$ & $\triple{10}{+}{S}$ & $\triple{2}{0}{R}$ & $\triple{2}{1}{R}$ \\
9 & $\triple{9}{\blank}{L}$ & $\triple{9}{\blank}{U}$ & $\triple{10}{0}{S}$ & $\triple{10}{1}{S}$ \\
10 & $\triple{10}{\blank}{S}$ & $\triple{10}{+}{S}$ & $\triple{10}{0}{S}$ & $\triple{10}{1}{S}$ \\
\bottomrule
\end{tabular}
\end{center}

\textbf{Metric-Extracted Controller.}\quad Initial state $0$; halt $21$.
\begin{center}
\footnotesize\renewcommand{\arraystretch}{1.12}
\setlength{\tabcolsep}{3pt}
\begin{tabular}{rcccc}
\toprule
State & $\blank$ & $+$ & $0$ & $1$\\
\midrule
0 & \text{--} & \text{--} & $\triple{1}{0}{R}$ & $\triple{1}{1}{R}$ \\
1 & \text{--} & $\triple{2}{+}{L}$ & $\triple{1}{0}{R}$ & $\triple{1}{1}{R}$ \\
2 & \text{--} & \text{--} & $\triple{3}{0}{D}$ & $\triple{3}{1}{D}$ \\
3 & $\triple{4}{+}{U}$ & \text{--} & \text{--} & \text{--} \\
4 & $\triple{15}{0}{R}$ & \text{--} & $\triple{5}{0}{R}$ & $\triple{5}{1}{R}$ \\
5 & \text{--} & $\triple{6}{+}{R}$ & $\triple{5}{0}{R}$ & $\triple{5}{1}{R}$ \\
6 & $\triple{7}{\blank}{L}$ & \text{--} & $\triple{6}{0}{R}$ & $\triple{6}{1}{R}$ \\
7 & \text{--} & $\triple{16}{\blank}{D}$ & $\triple{12}{\blank}{D}$ & $\triple{8}{\blank}{D}$ \\
8 & $\triple{9}{\blank}{L}$ & \text{--} & \text{--} & \text{--} \\
9 & $\triple{9}{\blank}{L}$ & $\triple{10}{+}{U}$ & \text{--} & \text{--} \\
10 & \text{--} & \text{--} & $\triple{11}{1}{D}$ & $\triple{18}{0}{L}$ \\
11 & \text{--} & $\triple{3}{\blank}{L}$ & \text{--} & \text{--} \\
12 & $\triple{13}{\blank}{L}$ & \text{--} & \text{--} & \text{--} \\
13 & $\triple{13}{\blank}{L}$ & $\triple{14}{+}{U}$ & \text{--} & \text{--} \\
14 & \text{--} & \text{--} & $\triple{11}{0}{D}$ & $\triple{11}{1}{D}$ \\
15 & \text{--} & \text{--} & $\triple{5}{0}{R}$ & $\triple{5}{1}{R}$ \\
16 & $\triple{16}{\blank}{L}$ & $\triple{17}{\blank}{U}$ & \text{--} & \text{--} \\
17 & \text{--} & \text{--} & $\triple{21}{0}{S}$ & $\triple{21}{1}{S}$ \\
18 & $\triple{20}{1}{D}$ & \text{--} & $\triple{19}{1}{D}$ & $\triple{18}{0}{L}$ \\
19 & $\triple{20}{\blank}{R}$ & \text{--} & \text{--} & \text{--} \\
20 & $\triple{20}{\blank}{R}$ & $\triple{3}{\blank}{L}$ & \text{--} & \text{--} \\
21 & $\triple{21}{\blank}{S}$ & $\triple{21}{+}{S}$ & $\triple{21}{0}{S}$ & $\triple{21}{1}{S}$ \\
\bottomrule
\end{tabular}
\end{center}

%% file: Appendix/Proof_Tables/a_add2.tex
\label{a_add2:sec:common4}
\paragraph{Phase Relation and Local Checks}\label{a_add2:sec:part3}
For phase $i$, write $Q_i$ for the controller-state set in the following table.
The relation pairs an expert configuration in phase $i$ with a controller
configuration having the \emph{same grid and head} and a state in $Q_i$.

\begin{center}
\small\renewcommand{\arraystretch}{1.12}
\begin{tabular}{r l c l}
\toprule
Phase $i$ & Meaning & Expert state $g_i$ & Controller states $Q_i$\\
\midrule
0 & First input digit & 0 & $\{0\}$ \\
1 & Seek the separator & 0 & $\{0\}$ \\
2 & Initial descent & 1 & $\{7\}$ \\
3 & Place the marker & 1 & $\{3,9\}$ \\
4 & Ensure an accumulator digit & 8 & $\{6\}$ \\
5 & Digit after zero extension & 2 & $\{6\}$ \\
6 & Scan toward separator & 2 & $\{4\}$ \\
7 & First cell after separator & 2 & $\{8\}$ \\
8 & Scan remaining right operand & 2 & $\{4\}$ \\
9 & Read rightmost factor digit & 3 & $\{13\}$ \\
10 & Transport zero: first blank & 4 & $\{2\}$ \\
11 & Transport zero: second blank & 4 & $\{2\}$ \\
12 & Transport zero: walk left & 4 & $\{2\}$ \\
13 & Add zero above marker & 4 & $\{7\}$ \\
14 & Transport one: first blank & 5 & $\{4\}$ \\
15 & Transport one: second blank & 5 & $\{13\}$ \\
16 & Transport one: walk left & 5 & $\{15\}$ \\
17 & Add one above marker & 5 & $\{1\}$ \\
18 & Propagate a carry & 6 & $\{1\}$ \\
19 & First return cell after carry & 7 & $\{5\}$ \\
20 & Return to the marker & 7 & $\{5\}$ \\
21 & At marker, without carry & 7 & $\{3,5,9\}$ \\
22 & Erase the input separator & 3 & $\{12\}$ \\
23 & First cleanup cell & 9 & $\{11\}$ \\
24 & Find and erase final marker & 9 & $\{14\}$ \\
25 & Preserve result and stop & 9 & $\{10\}$ \\
26 & Halted & 10 & $\{16\}$ \\
\bottomrule
\end{tabular}
\end{center}

\textbf{Local Transition Identity.} For every listed $q\in Q_i$ and every allowed
edge $i\xrightarrow{x}j$ from the protocol, the transition maps satisfy
\begin{equation}
\delta_\EXPERT(g_i,x)=(g_j,w,M),\qquad
\delta_\CTRL(q,x)=(q',w,M),\qquad q'\in Q_j. \label{eq:proof:a-add2.2}
\end{equation}
Every required controller entry exists. A transition reaches the halt
state exactly when $j=26$, with motion $S$.
The certificate contains \textbf{30 phase--state pairs} (including the terminal pair)
and \textbf{52 local checks}. The checks cover 37 distinct nonterminal
controller entries. \Cref{a_add2:sec:tables} provides the tables for direct
inspection; direct substitution checks every instance of~\eqref{eq:proof:a-add2.2}.

\paragraph{Transition Tables}\label{a_add2:sec:tables}
Each entry lists $(\text{next state},\text{written symbol},\text{motion})$.
The symbol $\blank$ means blank; a dash means an undefined transition.
State numbers identify the extracted controllers throughout each certificate.

\textbf{Annealed Controller.}\quad Initial state $0$; halt $16$.
\begin{center}
\footnotesize\renewcommand{\arraystretch}{1.12}
\setlength{\tabcolsep}{3pt}
\begin{tabular}{rcccc}
\toprule
State & $\blank$ & $+$ & $0$ & $1$\\
\midrule
0 & $\triple{13}{\blank}{L}$ & $\triple{7}{+}{L}$ & $\triple{0}{0}{R}$ & $\triple{0}{1}{R}$ \\
1 & $\triple{5}{1}{D}$ & $\triple{3}{\blank}{L}$ & $\triple{5}{1}{D}$ & $\triple{1}{0}{L}$ \\
2 & $\triple{2}{\blank}{L}$ & $\triple{7}{+}{U}$ & $\triple{5}{\blank}{D}$ & $\triple{4}{\blank}{D}$ \\
3 & $\triple{6}{+}{U}$ & $\triple{3}{\blank}{L}$ & $\triple{5}{1}{D}$ & $\triple{1}{0}{L}$ \\
4 & $\triple{13}{\blank}{L}$ & $\triple{8}{+}{R}$ & $\triple{4}{0}{R}$ & $\triple{4}{1}{R}$ \\
5 & $\triple{5}{\blank}{R}$ & $\triple{9}{\blank}{L}$ & $\triple{4}{0}{R}$ & $\triple{4}{1}{R}$ \\
6 & $\triple{6}{0}{R}$ & $\triple{7}{+}{L}$ & $\triple{4}{0}{R}$ & $\triple{4}{1}{R}$ \\
7 & $\triple{6}{+}{U}$ & $\triple{3}{\blank}{L}$ & $\triple{9}{0}{D}$ & $\triple{3}{1}{D}$ \\
8 & $\triple{12}{\blank}{L}$ & $\triple{8}{+}{R}$ & $\triple{4}{0}{R}$ & $\triple{4}{1}{R}$ \\
9 & $\triple{6}{+}{U}$ & $\triple{9}{\blank}{L}$ & $\triple{4}{0}{R}$ & $\triple{8}{1}{R}$ \\
10 & $\triple{14}{+}{U}$ & $\triple{10}{\blank}{D}$ & $\triple{16}{0}{S}$ & $\triple{16}{1}{S}$ \\
11 & $\triple{14}{\blank}{L}$ & $\triple{10}{\blank}{D}$ & $\triple{16}{0}{S}$ & $\triple{16}{1}{S}$ \\
12 & $\triple{15}{\blank}{L}$ & $\triple{11}{\blank}{D}$ & $\triple{4}{\blank}{D}$ & $\triple{4}{\blank}{D}$ \\
13 & $\triple{15}{\blank}{L}$ & $\triple{11}{\blank}{D}$ & $\triple{2}{\blank}{D}$ & $\triple{4}{\blank}{D}$ \\
14 & $\triple{14}{\blank}{L}$ & $\triple{10}{\blank}{U}$ & $\triple{4}{0}{R}$ & $\triple{4}{1}{R}$ \\
15 & $\triple{15}{\blank}{L}$ & $\triple{1}{+}{U}$ & $\triple{4}{0}{R}$ & $\triple{4}{1}{R}$ \\
16 & $\triple{16}{\blank}{S}$ & $\triple{16}{+}{S}$ & $\triple{16}{0}{S}$ & $\triple{16}{1}{S}$ \\
\bottomrule
\end{tabular}
\end{center}

%% file: Appendix/Proof_Tables/mult2_shared.tex
\label{mult2:sec:common2}
\paragraph{Copying and Partial Products}\label{mult2:sec:part1}
Write $b=b_{n-1}\cdots b_0$, so $B=\sum_{i=0}^{n-1}b_i2^i$.
Assign global weight $2^{m-1-z}$ to column $z$.
Column $m-1$ carries the units digit of the final product.

\begin{lemma}[Copy Staircase]\label{mult2:lem:copy}
Before product formation, the expert constructs exactly $n$ occupied rows.
For $0\le i<n$, row $i$ contains
\begin{equation}
a\,x\,b_{n-1}\cdots b_i
\quad\text{starting in column }-i. \label{eq:proof:mult2.1}
\end{equation}
The last row contains one right-operand bit. All remaining cells contain blanks.
\end{lemma}
\begin{proof}
Row $0$ contains the input. On row $i$, the expert scans to the first blank,
steps left to $b_i$, then steps left again. A digit at this position implies
at least two remaining right-operand bits. The expert copies every symbol
except $b_i$ one row down and one column left, proceeding from right to left.
Every destination contains a blank before writing; the source word remains unchanged.

The first descent reaches a blank on the new row. Each later descent reaches
the copied right neighbor: a digit, except directly below the last digit of
$a$, where the copied separator $x$ occupies the lower cell. The cell below
the source separator contains a digit. After copying the first digit of $a$,
the expert reads the blank immediately to the left and descends to the first
digit of the new row.

The new row satisfies \eqref{eq:proof:mult2.1} for $i+1$, with one fewer right-operand bit.
When one right-operand bit remains, the second left step reads $x$ and starts
product formation. Thus $n-1$ copies produce exactly $n$ occupied rows.
\end{proof}

\begin{lemma}[Aligned Partial Products]\label{mult2:lem:products}
After product formation, row $i$ contains the $(m+1)$-bit representation of
$A b_i$ in columns $-i-1,\ldots,m-i-1$, with a leading $0$.
In global column weights, row $i$ represents $A b_i2^i$.
All remaining cells contain blanks.
\end{lemma}
\begin{proof}
The expert forms products from row $n-1$ upwards. On row $i$, the expert
erases the rightmost factor bit $b_i$, the remaining factor prefix, and $x$,
then walks left across $a$. For $b_i=0$, the expert writes zeros; for $b_i=1$,
the expert preserves the digits of $a$. Both branches write one additional
$0$ on the left blank. The resulting word has the claimed support and value,
with a leading zero because $A b_i<2^m$.

The expert scans right to the blank at column $m-i$ and moves up.
For $i>0$, the upper cell contains the last digit of $a$ on row $i-1$;
the next cell contains $x$. The expert scans the remaining factor prefix,
selects the last bit, and repeats. Each upper prefix contains at least two
bits, so another digit follows the selected bit during leftward erasure.
For $i=0$, the upper cell contains a blank, ending product formation and
starting summation on row $0$.
\end{proof}

Both passes terminate: each copy shortens the remaining factor prefix, and
product formation visits each of the $n$ rows once. Every horizontal scan
crosses a finite word.

\paragraph{A Fixed-Width Addition Invariant}\label{mult2:sec:part2}
\begin{lemma}[Accumulated Product]\label{mult2:lem:merge}
After merging through row $k$, $0\le k<n$, all rows above $k$ contain only blanks.
Row $k$ contains a contiguous $(m+k+1)$-bit accumulator on columns
$-k-1,\ldots,m-1$, with value
\begin{equation*}
S_k=A\sum_{i=0}^{k}b_i2^i.
\end{equation*}
Rows below $k$ retain the partial products from \Cref{mult2:lem:products}.
At the inspection of the next row, the expert occupies state $20$ with head
$(-k-2,k+1)$.
\end{lemma}
\begin{proof}
For $k=0$, the initial left scan crosses the product on row $0$, reads the
leading $0$, reaches the blank at column $-2$, and descends.

If row $k+1$ exists, the inspected cell contains the leading $0$ at column
$-k-2$. The expert scans right across the lower product to the first blank,
at column $m-k-1$. Moving up reaches an accumulator digit because
\begin{equation*}
-k-1\ \le\ m-k-1\ \le\ m-1.
\end{equation*}
The expert scans the remaining upper digits, reaches the right blank, and
steps left to column $m-1$. The lower cell contains a blank because the
original lower product ends at $m-k-2$. The first column therefore adds
the upper bit to zero without a carry.

The expert processes the remaining columns from right to left. For upper
bit $u$, lower bit $v$ (with a blank counting as zero), and incoming carry
$c$, the update satisfies
\begin{equation}
u+v+c=w+2c',\qquad w\in\{0,1\},\quad c'\in\{0,1\}. \label{eq:proof:mult2.2}
\end{equation}
The expert erases the upper bit, writes $w$ below, and stores $c'$ in the
control state. The processed lower suffix contains the correct sum suffix,
the unprocessed cells retain the previous values, and the carry records
the contribution to the next column. Equation~\eqref{eq:proof:mult2.2} preserves this invariant.

The original lower word extends one column farther left than the upper word,
with a leading $0$ at $-k-2$. After consuming the upper word, carry zero
preserves this bit, whereas carry one changes the bit to $1$ and clears.
The expert reaches the blank at $-k-3$ on row $k+1$ and descends.
Row $k$ now contains only blanks, and row $k+1$ represents
$S_k+A b_{k+1}2^{k+1}=S_{k+1}$ on the claimed support.
\end{proof}

\begin{lemma}[Carry Confinement]\label{mult2:lem:carry}
During an upper-digit update with incoming carry $1$, the lower cell
contains an original binary digit. If the carry outlives the upper word,
the next lower cell contains the original leading $0$.
\end{lemma}
\begin{proof}
A blank gap lies to the right of the original lower product. The first
column has carry zero, and $u+0+0\le1$, so no carry originates in the gap.
A carry first arises inside the lower digit block. Further leftward steps
remain inside that block until the leading zero. The upper word ends one
column before that zero, where $0+0+1=1$ clears the carry.
\end{proof}

Each merge consumes finitely many upper digits and at most one extra carry
column. After $n-1$ merges, state $20$ reads the blank below the final row
and moves up to state $30$, which emits the final $S$ action and enters halt
state $31$. \Cref{mult2:lem:merge} gives the $(m+n)$-bit product and final head
$(-n-1,n-1)$. All other cells contain blanks.

%% file: Appendix/Proof_Tables/m_mult2.tex
\label{m_mult2:sec:common3}
The observation protocol refines expert states by geometric context. Each phase records the permitted read symbols and successor phases. \Cref{m_mult2:sec:common4} gives the complete protocol together with the controller-state relation; the shared tables avoid repeating the same phase definitions.

\begin{lemma}[Protocol Coverage]\label{m_mult2:lem:coverage}
Every expert execution on a well-formed multiplication follows the
$67$ phases in \Cref{m_mult2:sec:common4}, from phase $0$ to phase $66$.
\end{lemma}
\begin{proof}
\Cref{mult2:lem:copy} supplies the copy-phase conditions; \Cref{mult2:lem:products}
supplies the product-phase conditions. \Cref{mult2:lem:merge} and
\Cref{mult2:lem:carry} supply the alignment, addition, carry and boundary conditions.
The explanations below the three tables account for the first and last cell readings in each phase. Induction over the individual transitions, the copied rows,
and the merged rows establishes coverage. The finite-pass arguments in
the expert invariants give termination in the final phase.
\end{proof}

\label{m_mult2:sec:common4}
\paragraph{The Copying Phases}\label{m_mult2:sec:part3}
The following three tables define a finite protocol and a simulation relation.
Phase $i$ pairs expert state $g_i$ with controller state $q\in Q_i$ on an identical
grid and head. Each set $Q_i$ contains the single listed state. An entry
\begin{equation*}
x:(j,w,M)
\end{equation*}
means: read $x$, write $w$, move $M$, and advance to phase $j$.
Here $d\in\{0,1\}$ denotes either digit, consistently within each entry.
The index $j$ denotes a phase. Successors comprise expert state $g_j$
and controller state $q'\in Q_j$. The listed transitions specify the required local identities. The protocol starts at phase $0$.

\begingroup

\footnotesize\renewcommand{\arraystretch}{1.12}
\setlength{\tabcolsep}{2pt}
\begin{longtable}{r l c l l}
\toprule$i$ & Phase & $g_i$ & $Q_i$ & Read $:$ (next phase, write, move)\\\midrule
\endhead
0 & First left-operand digit & 0 & $\{0\}$ & $d:\triple{1}{d}{R}$ \\
1 & Scan left operand & 0 & $\{1\}$ & $d:\triple{1}{d}{R}\quad x:\triple{2}{x}{R}$ \\
2 & First right-operand digit & 0 & $\{2\}$ & $d:\triple{3}{d}{R}$ \\
3 & Scan right operand & 0 & $\{3\}$ & $d:\triple{3}{d}{R}\quad \blank:\triple{4}{\blank}{L}$ \\
4 & Skip last right-operand digit & 1 & $\{4\}$ & $d:\triple{5}{d}{L}$ \\
5 & Copy more, or start products & 2 & $\{5\}$ & $0:\triple{6}{0}{D}\quad 1:\triple{8}{1}{D}\quad x:\triple{22}{x}{R}$ \\
6 & First copied 0: descend & 3 & $\{52\}$ & $\blank:\triple{7}{\blank}{L}$ \\
7 & First copied 0: write & 6 & $\{53\}$ & $\blank:\triple{10}{0}{U}$ \\
8 & First copied 1: descend & 4 & $\{6\}$ & $\blank:\triple{9}{\blank}{L}$ \\
9 & First copied 1: write & 7 & $\{7\}$ & $\blank:\triple{10}{1}{U}$ \\
10 & Continue after first copy & 9 & $\{8\}$ & $0:\triple{11}{0}{D}\quad 1:\triple{13}{1}{D}\quad x:\triple{16}{x}{D}$ \\
11 & Copy 0 above a lower digit & 3 & $\{16\}$ & $d:\triple{12}{d}{L}$ \\
12 & Write copied 0 & 6 & $\{17\}$ & $\blank:\triple{15}{0}{U}$ \\
13 & Copy 1 above a lower digit & 4 & $\{9\}$ & $d:\triple{14}{d}{L}$ \\
14 & Write copied 1 & 7 & $\{10\}$ & $\blank:\triple{15}{1}{U}$ \\
15 & Continue the copy & 9 & $\{11\}$ & $0:\triple{11}{0}{D}\quad 1:\triple{13}{1}{D}\quad x:\triple{16}{x}{D}\quad \blank:\triple{21}{\blank}{D}$ \\
16 & Descend below operator & 5 & $\{12\}$ & $d:\triple{17}{d}{L}$ \\
17 & Write shifted operator & 8 & $\{13\}$ & $\blank:\triple{18}{x}{U}$ \\
18 & Copy last left-operand digit & 9 & $\{14\}$ & $0:\triple{19}{0}{D}\quad 1:\triple{20}{1}{D}$ \\
19 & Copied 0 above lower operator & 3 & $\{55\}$ & $x:\triple{12}{x}{L}$ \\
20 & Copied 1 above lower operator & 4 & $\{15\}$ & $x:\triple{14}{x}{L}$ \\
21 & Begin scan on copied row & 0 & $\{0\}$ & $d:\triple{1}{d}{R}$ \\
\bottomrule
\end{longtable}
\endgroup

\textbf{Why the Listed Read Symbols Cover Every Copy.}
The nonempty operands ensure that the initial scan reads a digit, the
separator, and at least one factor digit before reaching a blank. The second
left step sees either a digit to copy or $x$ when only one factor bit remains.
\Cref{mult2:lem:copy} identifies the three possible cells below a copied digit:
blank for the first copy, $x$ immediately after copying the separator, and
a digit otherwise. Every destination contains a blank. Copying the first digit of $a$
exposes the left blank; restarting on the new row therefore reads a digit.
These facts establish the table's geometric restrictions.

The protocol permits arbitrary operand lengths and may admit additional
observation sequences. Every valid execution follows permitted branches;
checking all such branches therefore establishes the required local agreement.

\paragraph{The One-Bit Product Phases}\label{m_mult2:sec:part4}
Phase $i$ pairs expert state $g_i$ with controller state $q\in Q_i$.
An edge $x:(j,w,M)$ reads $x$, writes $w$, performs motion $M$, and advances
to phase $j$.
The product phases preserve the exact leading-zero convention of the expert.

\begingroup

\footnotesize\renewcommand{\arraystretch}{1.12}
\setlength{\tabcolsep}{2pt}
\begin{longtable}{r l c l l}
\toprule$i$ & Phase & $g_i$ & $Q_i$ & Read $:$ (next phase, write, move)\\\midrule
\endhead
22 & Read the last remaining factor bit & 10 & $\{18\}$ & $0:\triple{23}{\blank}{L}\quad 1:\triple{26}{\blank}{L}$ \\
23 & Erase operator for factor 0 & 11 & $\{56\}$ & $x:\triple{24}{\blank}{L}$ \\
24 & First digit of product by 0 & 13 & $\{32\}$ & $d:\triple{25}{0}{L}$ \\
25 & Continue product by 0 & 13 & $\{33\}$ & $d:\triple{25}{0}{L}\quad \blank:\triple{29}{0}{R}$ \\
26 & Erase operator for factor 1 & 12 & $\{19\}$ & $x:\triple{27}{\blank}{L}$ \\
27 & First digit of product by 1 & 14 & $\{20\}$ & $d:\triple{28}{d}{L}$ \\
28 & Continue product by 1 & 14 & $\{21\}$ & $d:\triple{28}{d}{L}\quad \blank:\triple{29}{0}{R}$ \\
29 & Scan completed partial product & 15 & $\{22\}$ & $d:\triple{29}{d}{R}\quad \blank:\triple{30}{\blank}{U}$ \\
30 & Inspect the row above & 16 & $\{23\}$ & $d:\triple{31}{d}{R}\quad \blank:\triple{39}{\blank}{D}$ \\
31 & Cross upper operator & 17 & $\{24\}$ & $x:\triple{32}{x}{R}$ \\
32 & First bit in upper factor prefix & 17 & $\{25\}$ & $d:\triple{33}{d}{R}$ \\
33 & Scan upper factor prefix & 17 & $\{26\}$ & $d:\triple{33}{d}{R}\quad \blank:\triple{34}{\blank}{L}$ \\
34 & Read upper row factor bit & 10 & $\{27\}$ & $0:\triple{35}{\blank}{L}\quad 1:\triple{37}{\blank}{L}$ \\
35 & First remaining bit to erase (0) & 11 & $\{30\}$ & $d:\triple{36}{\blank}{L}$ \\
36 & Erase factor prefix (0) & 11 & $\{31\}$ & $d:\triple{36}{\blank}{L}\quad x:\triple{24}{\blank}{L}$ \\
37 & First remaining bit to erase (1) & 12 & $\{28\}$ & $d:\triple{38}{\blank}{L}$ \\
38 & Erase factor prefix (1) & 12 & $\{29\}$ & $d:\triple{38}{\blank}{L}\quad x:\triple{27}{\blank}{L}$ \\
\bottomrule
\end{longtable}
\endgroup

\textbf{First and Subsequent Product Rows.}
The bottom row contains one factor bit. Erasing that bit moves the head
directly onto $x$. Every higher row has at least two right-operand bits:
after erasing the selected last bit, the expert erases another digit
before reaching the separator. These facts justify the separate ``single'' and
``first remaining bit'' phases in the table.

\textbf{Upper-Row Alignment.}
A finished product occupies $[-i-1,m-i-1]$ on row $i$. The right scan
ends at column $m-i$. An existing upper row has the last left-operand digit
at that column, followed immediately by $x$. After that separator comes
a nonempty remaining right operand. If no row above exists, the expert reads a blank and starts summation. \Cref{mult2:lem:products} proves these
conditions for every operand length, including operands containing only zeros.

\textbf{Product-Scan Coverage.}
Multiplication by $0$ turns each left-operand digit into $0$; multiplication
by $1$ preserves each digit. Both passes have at least one digit to process and
finish on a blank by writing a padding $0$. The subsequent right scan sees
only digits followed by a blank. A completed product row contains no separators.

\paragraph{Summation, Carry and the Final Stop}\label{m_mult2:sec:part5}
These phases refine the expert's addition states by the position in a
merge and the carry condition. This refinement supplies every required transition in the partial controller table on all well-formed multiplications.

\begingroup

\footnotesize\renewcommand{\arraystretch}{1.12}
\setlength{\tabcolsep}{2pt}
\begin{longtable}{r l c l l}
\toprule$i$ & Phase & $g_i$ & $Q_i$ & Read $:$ (next phase, write, move)\\\midrule
\endhead
39 & Start summing the product rows & 18 & $\{34\}$ & $\blank:\triple{40}{\blank}{L}$ \\
40 & First digit of initial left scan & 19 & $\{35\}$ & $0:\triple{41}{0}{L}\quad 1:\triple{42}{1}{L}$ \\
41 & Left scan after reading zero & 19 & $\{36\}$ & $0:\triple{41}{0}{L}\quad 1:\triple{42}{1}{L}\quad \blank:\triple{43}{\blank}{D}$ \\
42 & Left scan after reading one & 19 & $\{35\}$ & $0:\triple{41}{0}{L}\quad 1:\triple{42}{1}{L}$ \\
43 & Inspect fresh lower leading cell & 20 & $\{37\}$ & $0:\triple{44}{0}{R}\quad \blank:\triple{65}{\blank}{U}$ \\
44 & First digit after lower leading zero & 21 & $\{22\}$ & $d:\triple{45}{d}{R}$ \\
45 & Scan the lower partial product & 21 & $\{22\}$ & $d:\triple{45}{d}{R}\quad \blank:\triple{46}{\blank}{U}$ \\
46 & First upper alignment digit & 28 & $\{23\}$ & $d:\triple{47}{d}{R}$ \\
47 & Next upper alignment cell & 28 & $\{24\}$ & $d:\triple{48}{d}{R}\quad \blank:\triple{49}{\blank}{L}$ \\
48 & Finish aligning the upper row & 28 & $\{45\}$ & $d:\triple{48}{d}{R}\quad \blank:\triple{49}{\blank}{L}$ \\
49 & Read upper units digit & 22 & $\{38\}$ & $0:\triple{50}{\blank}{D}\quad 1:\triple{51}{\blank}{D}$ \\
50 & Write units digit 0 below & 24 & $\{39\}$ & $\blank:\triple{52}{0}{U}$ \\
51 & Write units digit 1 below & 25 & $\{54\}$ & $\blank:\triple{52}{1}{U}$ \\
52 & Return above, carry zero & 23 & $\{40\}$ & $\blank:\triple{53}{\blank}{L}$ \\
53 & Read upper bit, carry zero & 22 & $\{41\}$ & $0:\triple{54}{\blank}{D}\quad 1:\triple{55}{\blank}{D}\quad \blank:\triple{60}{\blank}{D}$ \\
54 & Add value zero below & 24 & $\{42\}$ & $d:\triple{52}{d}{U}\quad \blank:\triple{52}{0}{U}$ \\
55 & Add value one below & 25 & $\{46\}$ & $\blank:\triple{52}{1}{U}\quad 0:\triple{52}{1}{U}\quad 1:\triple{56}{0}{U}$ \\
56 & Return above, carry one & 27 & $\{47\}$ & $\blank:\triple{57}{\blank}{L}$ \\
57 & Read upper bit, carry one & 26 & $\{48\}$ & $0:\triple{58}{\blank}{D}\quad 1:\triple{59}{\blank}{D}\quad \blank:\triple{62}{\blank}{D}$ \\
58 & Add one inside lower digit block & 25 & $\{50\}$ & $0:\triple{52}{1}{U}\quad 1:\triple{56}{0}{U}$ \\
59 & Add two inside lower digit block & 29 & $\{49\}$ & $d:\triple{56}{d}{U}$ \\
60 & Preserve unneeded leading zero & 19 & $\{43\}$ & $0:\triple{61}{0}{L}$ \\
61 & Blank to the left of merged row & 19 & $\{44\}$ & $\blank:\triple{43}{\blank}{D}$ \\
62 & Apply final carry to leading zero & 25 & $\{57\}$ & $0:\triple{63}{1}{U}$ \\
63 & Return after final carry & 23 & $\{58\}$ & $\blank:\triple{64}{\blank}{L}$ \\
64 & Step down left of merged row & 22 & $\{59\}$ & $\blank:\triple{61}{\blank}{D}$ \\
65 & Emit final stop & 30 & $\{51\}$ & $\blank:\triple{66}{\blank}{S}$ \\
66 & Halted & 31 & $\{60\}$ & $\text{--}$ \\
\bottomrule
\end{longtable}
\endgroup

\textbf{Boundary Restrictions.}
The initial left scan can encounter a blank only after reading the product's
leading $0$. The next inspection reads either a blank or the fresh row's leading $0$.
Alignment always starts on an upper digit, and the first lower units cell contains a blank. \Cref{mult2:lem:carry} excludes lower blanks in the carry-one digit
phases and forces $0$ for the final carry extension. With no final carry, the expert preserves the untouched leading $0$; after either branch, the next cell to the left of the result contains a blank. The final stop also reads a blank.

These facts, together with contiguity of every accumulator, justify all
read restrictions in this table. The restrictions exclude every missing transition
that could otherwise obstruct a valid multiplication.

\textbf{Local Transition Identity.}
For each of the $130$ listed read/action cases and the unique $q\in Q_i$, direct comparison with the
two transition maps gives
\begin{equation}
\delta_\EXPERT(g_i,x)=(g_j,w,M),\qquad
\delta_\CTRL(q,x)=(q',w,M),\qquad q'\in Q_j. \label{eq:proof:m-mult2.3}
\end{equation}
Every required entry exists. A transition reaches a halt state exactly
when $j=66$, with motion $S$. The certificate covers all $119$ defined
nonterminal controller entries. The remaining four entries contain halt loops.

%% file: Appendix/Proof_Tables/a_mult2.tex
\label{a_mult2:sec:common3}
The observation protocol refines expert states by geometric context. Each phase records the permitted read symbols and successor phases. \Cref{a_mult2:sec:common4} gives the complete protocol together with the controller-state relation; the shared tables avoid repeating the same phase definitions.

\begin{lemma}[Protocol Coverage]\label{a_mult2:lem:coverage}
Every expert execution on a well-formed multiplication follows the
$94$ phases in \Cref{a_mult2:sec:common4}, from phase $0$ to phase $93$.
\end{lemma}
\begin{proof}
The copy and product invariants (\Cref{mult2:lem:copy}--\Cref{mult2:lem:products})
give all copying and product-reading conditions. The scan annotations
remember unchanged cells and distinguish one digit from at least two;
the explanations in \Cref{a_mult2:sec:part3}--\Cref{a_mult2:sec:part5} prove these refinements.
\Cref{mult2:lem:merge}--\Cref{mult2:lem:carry} and the blank-gap argument in
\Cref{a_mult2:sec:part6} supply the addition and boundary conditions. Induction over
each transition, each copied row and each merged row establishes coverage.
The finite-pass arguments in the expert invariants give termination in phase $93$.
\end{proof}

\label{a_mult2:sec:common4}
\paragraph{Copying and Remembered Factor Bits}\label{a_mult2:sec:part3}
The next four tables define a finite protocol and simulation relation.
Phase $i$ pairs expert state $g_i$ with a controller state $q\in Q_i$,
with identical grids and heads. An entry $x:(j,w,M)$ means read $x$,
write $w$, move $M$, and advance to \emph{phase} $j$.
The symbol $d\in\{0,1\}$ means either digit, consistently within that entry.
For \emph{each} $q\in Q_i$, the actual successor lies in $Q_j$.

\begingroup

\footnotesize\renewcommand{\arraystretch}{1.12}
\setlength{\tabcolsep}{2pt}
\begin{longtable}{r l c l l}
\toprule$i$ & Phase & $g_i$ & $Q_i$ & Read $:$ (next phase, write, move)\\\midrule
\endhead
0 & First left-operand digit & 0 & $\{0\}$ & $d:\triple{1}{d}{R}$ \\
1 & Scan left operand & 0 & $\{0,24\}$ & $d:\triple{1}{d}{R}\quad x:\triple{2}{x}{R}$ \\
2 & First right-operand digit & 0 & $\{7\}$ & $0:\triple{3}{0}{R}\quad 1:\triple{4}{1}{R}$ \\
3 & Scan factor; last bit 0 & 0 & $\{48\}$ & $0:\triple{3}{0}{R}\quad 1:\triple{4}{1}{R}\quad \blank:\triple{5}{\blank}{L}$ \\
4 & Scan factor; last bit 1 & 0 & $\{48\}$ & $0:\triple{3}{0}{R}\quad 1:\triple{4}{1}{R}\quad \blank:\triple{6}{\blank}{L}$ \\
5 & Revisit last factor bit 0 & 1 & $\{14\}$ & $0:\triple{7}{0}{L}$ \\
6 & Revisit last factor bit 1 & 1 & $\{14\}$ & $1:\triple{8}{1}{L}$ \\
7 & Copy or use remembered bit 0 & 2 & $\{49\}$ & $0:\triple{9}{0}{D}\quad 1:\triple{11}{1}{D}\quad x:\triple{25}{x}{R}$ \\
8 & Copy or use remembered bit 1 & 2 & $\{19\}$ & $0:\triple{9}{0}{D}\quad 1:\triple{11}{1}{D}\quad x:\triple{26}{x}{R}$ \\
9 & First copied 0: descend & 3 & $\{16\}$ & $\blank:\triple{10}{\blank}{L}$ \\
10 & First copied 0: write & 6 & $\{13\}$ & $\blank:\triple{13}{0}{U}$ \\
11 & First copied 1: descend & 4 & $\{4\}$ & $\blank:\triple{12}{\blank}{L}$ \\
12 & First copied 1: write & 7 & $\{41\}$ & $\blank:\triple{13}{1}{U}$ \\
13 & Continue after first copy & 9 & $\{35\}$ & $0:\triple{14}{0}{D}\quad 1:\triple{16}{1}{D}\quad x:\triple{19}{x}{D}$ \\
14 & Copy 0 above a lower digit & 3 & $\{44,47\}$ & $d:\triple{15}{d}{L}$ \\
15 & Write copied 0 & 6 & $\{9,49\}$ & $\blank:\triple{18}{0}{U}$ \\
16 & Copy 1 above a lower digit & 4 & $\{26\}$ & $d:\triple{17}{d}{L}$ \\
17 & Write copied 1 & 7 & $\{2,3\}$ & $\blank:\triple{18}{1}{U}$ \\
18 & Continue the copy & 9 & $\{34\}$ & $0:\triple{14}{0}{D}\quad 1:\triple{16}{1}{D}\quad x:\triple{19}{x}{D}\quad \blank:\triple{24}{\blank}{D}$ \\
19 & Descend below operator & 5 & $\{45\}$ & $d:\triple{20}{d}{L}$ \\
20 & Write shifted operator & 8 & $\{5\}$ & $\blank:\triple{21}{x}{U}$ \\
21 & Copy last left-operand digit & 9 & $\{34\}$ & $0:\triple{22}{0}{D}\quad 1:\triple{23}{1}{D}$ \\
22 & Copied 0 above lower operator & 3 & $\{44\}$ & $x:\triple{15}{x}{L}$ \\
23 & Copied 1 above lower operator & 4 & $\{26\}$ & $x:\triple{17}{x}{L}$ \\
24 & Begin scan on copied row & 0 & $\{36\}$ & $d:\triple{1}{d}{R}$ \\
\bottomrule
\end{longtable}
\endgroup

\textbf{Coverage.} Both operands contain at least one digit. The initial scan therefore
meets a digit, $x$, and a right-operand digit before a blank. Phases $3$--$8$
remember the last factor bit: the scan's last digit, the digit revisited
by the first left step, and (if only one factor bit remains) the digit
selected for multiplication all refer to the same unchanged cell.

\Cref{mult2:lem:copy} proves the remaining restrictions: the first descent
sees a blank; subsequent descents see the already copied right-neighbor
digit, except below the last left-operand digit, where the lower cell
contains $x$. Every destination contains a blank. Copying $x$ exposes a
source digit. Restarting on a copied row also reads a digit. These conditions
hold for arbitrary operand lengths and values.

\paragraph{Forming and Scanning Partial Products}\label{a_mult2:sec:part4}
The product scan starts just to the right of the new padding zero.
``One'' denotes a scan after exactly one product digit; the remaining
scan phases follow at least two product digits. Each scan also records the most recently read digit.

\begingroup

\footnotesize\renewcommand{\arraystretch}{1.12}
\setlength{\tabcolsep}{2pt}
\begin{longtable}{r l c l l}
\toprule$i$ & Phase & $g_i$ & $Q_i$ & Read $:$ (next phase, write, move)\\\midrule
\endhead
25 & Select remembered factor bit 0 & 10 & $\{10\}$ & $0:\triple{27}{\blank}{L}$ \\
26 & Select remembered factor bit 1 & 10 & $\{3\}$ & $1:\triple{28}{\blank}{L}$ \\
27 & Erase operator for factor 0 & 11 & $\{10\}$ & $x:\triple{29}{\blank}{L}$ \\
28 & Erase operator for factor 1 & 12 & $\{4\}$ & $x:\triple{31}{\blank}{L}$ \\
29 & First digit of product by 0 & 13 & $\{25\}$ & $d:\triple{30}{0}{L}$ \\
30 & Continue product by 0 & 13 & $\{25\}$ & $d:\triple{30}{0}{L}\quad \blank:\triple{33}{0}{R}$ \\
31 & First digit of product by 1 & 14 & $\{6\}$ & $d:\triple{32}{d}{L}$ \\
32 & Continue product by 1 & 14 & $\{6\}$ & $d:\triple{32}{d}{L}\quad \blank:\triple{33}{0}{R}$ \\
33 & First digit after product padding & 15 & $\{15,36\}$ & $0:\triple{34}{0}{R}\quad 1:\triple{35}{1}{R}$ \\
34 & After one product digit 0 & 15 & $\{0\}$ & $0:\triple{36}{0}{R}\quad 1:\triple{37}{1}{R}\quad \blank:\triple{38}{\blank}{U}$ \\
35 & After one product digit 1 & 15 & $\{24\}$ & $0:\triple{36}{0}{R}\quad 1:\triple{37}{1}{R}\quad \blank:\triple{39}{\blank}{U}$ \\
36 & Product scan; last digit 0 & 15 & $\{24\}$ & $0:\triple{36}{0}{R}\quad 1:\triple{37}{1}{R}\quad \blank:\triple{40}{\blank}{U}$ \\
37 & Product scan; last digit 1 & 15 & $\{24\}$ & $0:\triple{36}{0}{R}\quad 1:\triple{37}{1}{R}\quad \blank:\triple{41}{\blank}{U}$ \\
38 & Inspect above one-digit product 0 & 16 & $\{39\}$ & $d:\triple{42}{d}{R}\quad \blank:\triple{52}{\blank}{D}$ \\
39 & Inspect above one-digit product 1 & 16 & $\{17\}$ & $d:\triple{42}{d}{R}\quad \blank:\triple{53}{\blank}{D}$ \\
40 & Inspect above; last digit 0 & 16 & $\{17\}$ & $d:\triple{42}{d}{R}\quad \blank:\triple{54}{\blank}{D}$ \\
41 & Inspect above; last digit 1 & 16 & $\{17\}$ & $d:\triple{42}{d}{R}\quad \blank:\triple{55}{\blank}{D}$ \\
42 & Cross upper operator & 17 & $\{8,43\}$ & $x:\triple{43}{x}{R}$ \\
43 & First bit in upper factor prefix & 17 & $\{16\}$ & $0:\triple{44}{0}{R}\quad 1:\triple{45}{1}{R}$ \\
44 & Scan factor; last bit 0 & 17 & $\{16\}$ & $0:\triple{44}{0}{R}\quad 1:\triple{45}{1}{R}\quad \blank:\triple{46}{\blank}{L}$ \\
45 & Scan factor; last bit 1 & 17 & $\{16\}$ & $0:\triple{44}{0}{R}\quad 1:\triple{45}{1}{R}\quad \blank:\triple{47}{\blank}{L}$ \\
46 & Select remembered factor bit 0 & 10 & $\{13\}$ & $0:\triple{48}{\blank}{L}$ \\
47 & Select remembered factor bit 1 & 10 & $\{13\}$ & $1:\triple{50}{\blank}{L}$ \\
48 & First remaining bit to erase (0) & 11 & $\{27\}$ & $d:\triple{49}{\blank}{L}$ \\
49 & Erase factor prefix (0) & 11 & $\{10\}$ & $d:\triple{49}{\blank}{L}\quad x:\triple{29}{\blank}{L}$ \\
50 & First remaining bit to erase (1) & 12 & $\{22\}$ & $d:\triple{51}{\blank}{L}$ \\
51 & Erase factor prefix (1) & 12 & $\{1\}$ & $d:\triple{51}{\blank}{L}\quad x:\triple{31}{\blank}{L}$ \\
\bottomrule
\end{longtable}
\endgroup

\textbf{Coverage.} \Cref{mult2:lem:products} gives a nonempty block of $m$
product digits after the padding zero. This justifies the first-digit phase
and the split into exactly one or at least two scanned digits. At the right blank, the most recently read digit equals the units digit.

The bottom row contains one factor bit; erasing that bit exposes $x$.
Every higher row contains at least two factor bits, so another digit
precedes $x$ during erasure. An upward return from a completed row reaches
either a blank or the last left-operand digit followed by $x$.
The upper factor scan records the last bit, which remains unchanged until selection. All these facts follow from the copy staircase.

\paragraph{Starting the Sum and Aligning a Row}\label{a_mult2:sec:part5}
These phases retain the last product digit until the next visit and
distinguish a one-digit left operand from a longer operand. The remembered
bit and length class serve as proof annotations without adding machine registers.

\begingroup

\footnotesize\renewcommand{\arraystretch}{1.12}
\setlength{\tabcolsep}{2pt}
\begin{longtable}{r l c l l}
\toprule$i$ & Phase & $g_i$ & $Q_i$ & Read $:$ (next phase, write, move)\\\midrule
\endhead
52 & Begin one-digit sum; last bit 0 & 18 & $\{20\}$ & $\blank:\triple{56}{\blank}{L}$ \\
53 & Begin one-digit sum; last bit 1 & 18 & $\{42\}$ & $\blank:\triple{57}{\blank}{L}$ \\
54 & Begin sum; remember last digit 0 & 18 & $\{42\}$ & $\blank:\triple{58}{\blank}{L}$ \\
55 & Begin sum; remember last digit 1 & 18 & $\{42\}$ & $\blank:\triple{59}{\blank}{L}$ \\
56 & Revisit sole product digit 0 & 19 & $\{28\}$ & $0:\triple{60}{0}{L}$ \\
57 & Revisit sole product digit 1 & 19 & $\{11\}$ & $1:\triple{60}{1}{L}$ \\
58 & Revisit product units digit 0 & 19 & $\{11\}$ & $0:\triple{62}{0}{L}$ \\
59 & Revisit product units digit 1 & 19 & $\{11\}$ & $1:\triple{62}{1}{L}$ \\
60 & Read the sole leading zero & 19 & $\{2\}$ & $0:\triple{61}{0}{L}$ \\
61 & Left boundary after sole digit & 19 & $\{23\}$ & $\blank:\triple{66}{\blank}{D}$ \\
62 & Second digit of initial left scan & 19 & $\{2,26\}$ & $d:\triple{63}{d}{L}$ \\
63 & Third digit of initial left scan & 19 & $\{2,23\}$ & $0:\triple{64}{0}{L}\quad 1:\triple{65}{1}{L}$ \\
64 & Left scan after reading zero & 19 & $\{23\}$ & $0:\triple{64}{0}{L}\quad 1:\triple{65}{1}{L}\quad \blank:\triple{66}{\blank}{D}$ \\
65 & Left scan after reading one & 19 & $\{23\}$ & $0:\triple{64}{0}{L}\quad 1:\triple{65}{1}{L}$ \\
66 & Inspect fresh lower leading cell & 20 & $\{12,36\}$ & $0:\triple{67}{0}{R}\quad \blank:\triple{92}{\blank}{U}$ \\
67 & First digit after lower leading zero & 21 & $\{0\}$ & $d:\triple{68}{d}{R}$ \\
68 & Scan the lower partial product & 21 & $\{24\}$ & $d:\triple{68}{d}{R}\quad \blank:\triple{69}{\blank}{U}$ \\
69 & First upper alignment digit & 28 & $\{17\}$ & $d:\triple{70}{d}{R}$ \\
70 & Next upper alignment cell & 28 & $\{8\}$ & $d:\triple{71}{d}{R}\quad \blank:\triple{72}{\blank}{L}$ \\
71 & Finish aligning the upper row & 28 & $\{4,8\}$ & $d:\triple{71}{d}{R}\quad \blank:\triple{72}{\blank}{L}$ \\
\bottomrule
\end{longtable}
\endgroup

\textbf{Initial Left-Scan Coverage.}
On the top row, the product's units digit has not changed between the right scan and the first left step of summation. If $m=1$, the full padded word equals $0d$: scanning left reads $d$, then $0$, then a blank (phases
$56$--$57$, $60$--$61$). If $m\ge2$, the padded word has at least three digits,
so two more digits must follow the first one (phases $62$--$63$).
In both cases the padding zero immediately precedes the left blank.

\textbf{Alignment Coverage.}
\Cref{mult2:lem:merge} gives either a blank in the inspected lower cell
(no remaining row) or the next row's leading $0$. After that zero comes
at least one digit. The upward move from the lower right boundary reaches
an upper digit, so alignment begins inside the upper word. The expert scans to the right blank and then returns left to the units digit. The first lower units cell contains a blank. These facts justify every alignment branch here and
the first three phases in the next certificate subsection.

\paragraph{Column Addition, Carries and Stopping}\label{a_mult2:sec:part6}
Separate phases track carry-zero returns from lower blanks and lower digits. This distinction prevents an
impossible early end of the upper word from entering the proof protocol.

\begingroup

\footnotesize\renewcommand{\arraystretch}{1.12}
\setlength{\tabcolsep}{2pt}
\begin{longtable}{r l c l l}
\toprule$i$ & Phase & $g_i$ & $Q_i$ & Read $:$ (next phase, write, move)\\\midrule
\endhead
72 & Read upper units digit & 22 & $\{32,41\}$ & $0:\triple{73}{\blank}{D}\quad 1:\triple{74}{\blank}{D}$ \\
73 & Write units digit 0 below & 24 & $\{27,46\}$ & $\blank:\triple{88}{0}{U}$ \\
74 & Write units digit 1 below & 25 & $\{22\}$ & $\blank:\triple{88}{1}{U}$ \\
75 & Return above, carry zero & 23 & $\{50\}$ & $\blank:\triple{76}{\blank}{L}$ \\
76 & Read upper bit, carry zero & 22 & $\{37\}$ & $0:\triple{77}{\blank}{D}\quad 1:\triple{78}{\blank}{D}\quad \blank:\triple{83}{\blank}{D}$ \\
77 & Add value zero below & 24 & $\{30\}$ & $d:\triple{75}{d}{U}\quad \blank:\triple{90}{0}{U}$ \\
78 & Add value one below & 25 & $\{29\}$ & $\blank:\triple{90}{1}{U}\quad 0:\triple{75}{1}{U}\quad 1:\triple{79}{0}{U}$ \\
79 & Return above, carry one & 27 & $\{10\}$ & $\blank:\triple{80}{\blank}{L}$ \\
80 & Read upper bit, carry one & 26 & $\{38\}$ & $0:\triple{81}{\blank}{D}\quad 1:\triple{82}{\blank}{D}\quad \blank:\triple{85}{\blank}{D}$ \\
81 & Add one inside lower digit block & 25 & $\{29\}$ & $0:\triple{75}{1}{U}\quad 1:\triple{79}{0}{U}$ \\
82 & Add two inside lower digit block & 29 & $\{9\}$ & $d:\triple{79}{d}{U}$ \\
83 & Preserve unneeded leading zero & 19 & $\{20\}$ & $0:\triple{84}{0}{L}$ \\
84 & Blank to the left of merged row & 19 & $\{31,34\}$ & $\blank:\triple{66}{\blank}{D}$ \\
85 & Apply final carry to leading zero & 25 & $\{42\}$ & $0:\triple{86}{1}{U}$ \\
86 & Return after final carry & 23 & $\{47\}$ & $\blank:\triple{87}{\blank}{L}$ \\
87 & Step down left of merged row & 22 & $\{33\}$ & $\blank:\triple{84}{\blank}{D}$ \\
88 & Return after units digit & 23 & $\{18,21\}$ & $\blank:\triple{89}{\blank}{L}$ \\
89 & Read the second upper digit & 22 & $\{21\}$ & $0:\triple{77}{\blank}{D}\quad 1:\triple{78}{\blank}{D}$ \\
90 & Return from the lower blank gap & 23 & $\{21\}$ & $\blank:\triple{91}{\blank}{L}$ \\
91 & Upper digit after lower blank gap & 22 & $\{21\}$ & $0:\triple{77}{\blank}{D}\quad 1:\triple{78}{\blank}{D}$ \\
92 & Emit final stop & 30 & $\{35,40\}$ & $\blank:\triple{93}{\blank}{S}$ \\
93 & Halted & 31 & $\{51\}$ & $\text{--}$ \\
\bottomrule
\end{longtable}
\endgroup

\textbf{The Lower Blank Gap.}
While merging row $k$ into row $k+1$, the upper word begins at $-k-1$;
the original lower word ends at $m-k-2\ge-k-1$. A blank below an upper digit
must therefore lie to the right of that lower endpoint. The next column
left still has index at least $-k-1$, so the next upper reading contains a digit.
This proves the restriction in phase $91$. Independently, the upper word
has length $m+k+1\ge2$, proving the second-digit restriction in phase $89$.

\textbf{Carry and Boundary Coverage.}
\Cref{mult2:lem:carry} excludes lower blanks in the carry-one digit phases
and forces $0$ for the final carry extension. Returning above a written
lower digit reads the upper blank just erased. When the upper word ends,
the next lower cell contains the original leading zero: carry zero
preserves this bit, whereas carry one changes the bit to $1$. Both branches reach the blank to the
left of the merged row. When no next row remains, the upward move reaches
a blank, and the expert emits the final $S$. Phase $93$ has no further PTO actions.

\textbf{Local Transition Identity.}
For every listed input branch $x:(j,w,M)$ and every $q\in Q_i$, direct
comparison with the two transition maps gives
\begin{equation}
\delta_\EXPERT(g_i,x)=(g_j,w,M),\qquad
\delta_\CTRL(q,x)=(q',w,M)\quad\text{for some }q'\in Q_j. \label{eq:proof:a-mult2.3}
\end{equation}
The certificate contains $109$ phase--state pairs and $201$ local checks,
covering $133$ distinct nonterminal controller entries. A transition reaches
a halt state exactly when $j=93$, with $M=S$. Initially $g_0=0$ and $Q_0=\{0\}$.
Protocol coverage excludes the remaining controller entries on valid inputs.

%% file: Appendix/Proof_Tables/add10.tex
\label{m_add10:sec:common2}
\paragraph{Accumulator and Marker Invariant}\label{m_add10:sec:part1}
Write $B=\sum_{j=0}^{n-1}b_j10^j$, with $b_0$ the rightmost digit.
The expert scans to the separator at column $m$, steps left and down,
writes a working $+$ below column $m-1$, and enters state $16$ above the marker.

\begin{lemma}[Round Invariant]\label{m_add10:lem:round}
After processing $k$ right-operand digits, $0\le k\le n$, the expert enters
state $16$ at $(p_k,0)$, where $p_k=m-1-k$. At this point:
\begin{itemize}
\item Row $1$ contains exactly one nonblank symbol: a $+$ at column $p_k$.
\item Row $0$ contains a contiguous accumulator on $[\ell_k,m-1]$, with value
$C_k=A+\sum_{j<k}b_j10^j$ and left edge $\ell_k\le p_k+1$.
Column $m-1$ carries the units digit.
\item The separator remains at column $m$. The unprocessed prefix of $b$
occupies columns $m+1,\ldots,m+n-k$. All remaining cells contain blanks.
\end{itemize}
\end{lemma}
\begin{proof}
Initialization gives $k=0$, $\ell_0=0$, $C_0=A$, and $p_0=m-1$.
State $16$ preserves the digit at $p_k$, or writes $0$ on a blank.
A blank at $p_k$ forces $\ell_k=p_k+1$, so zero extension preserves contiguity.

For $k<n$, the right scan reaches the blank after the unprocessed prefix,
steps left to digit $b_k$ at $r_k=m+n-k$, erases $b_k$, and descends.
The expert transports $b_k$ left along row $1$ to the marker. The distance satisfies
\begin{equation}
r_k-p_k=n+1\ge2. \label{eq:proof:m-add10.1}
\end{equation}
Consequently, the first two transport steps read blanks. At the marker, the
expert ascends and adds $b_k$ to the accumulator digit $e$ at column $p_k$.
If $e+b_k<10$, the expert writes $e+b_k$ and descends. Otherwise, the expert
writes $e+b_k-10$, moves left, and propagates one carry. Each digit $9$
changes to $0$; the first smaller digit increases by one, or a blank changes
to $1$. The update adds exactly $b_k10^k$ and preserves contiguity.

The expert returns along row $1$, erases the marker, steps left, writes the
next marker at $p_{k+1}=p_k-1$, and ascends into state $16$.
The earlier zero extension gives $\ell_{k+1}\le p_k=p_{k+1}+1$.
The accumulator now has value $C_{k+1}$, establishing the next round.
\end{proof}

\begin{lemma}[Termination and Final Configuration]\label{m_add10:lem:finish}
The expert terminates with only the padded base-$10$ representation of $A+B$
on row $0$, ending at column $m-1$. The final head stops at $(m-n-1,0)$.

Writing $\ell_{10}(s)$ for the decimal digit count, with $\ell_{10}(0)=1$,
the output width equals $L=\max\{m,n+1,\ell_{10}(A+B)\}$.
\end{lemma}
\begin{proof}
Each round consumes one of the $n$ right-operand digits. Every scan crosses
a finite interval, and every carry reaches a digit below $9$ or a blank.
After round $n$, state $16$ ensures a digit at $p_n$. The next right scan
returns to the separator. State $3$ erases the separator and descends.
State $17$ walks left, erases the final marker, ascends, preserves the
accumulator digit, and stops. The invariant gives $C_n=A+B$, and cleanup
removes all work symbols.

The original $m$ columns persist, and marker extension reaches $p_n$,
ensuring at least $n+1$ digits. Every additional carry extension contributes
a digit required by the current numerical value. Since partial sums never
decrease, the final width equals $L$.
\end{proof}

\label{m_add10:sec:common3}
The observation protocol refines expert states by geometric context. Each phase records the permitted read symbols and successor phases. \Cref{m_add10:sec:common4} gives the complete protocol together with the controller-state relation; the shared tables avoid repeating the same phase definitions.

\begin{lemma}[Protocol Coverage]\label{m_add10:lem:coverage}
Every expert execution on a valid decimal addition grid follows the $59$
phases in \Cref{m_add10:sec:common4}, starting at phase $0$ and ending at phase $58$.
\end{lemma}
\begin{proof}
Initialization satisfies the first phases because both operands contain a digit.
\Cref{m_add10:lem:round} gives the contiguous accumulator, the ensured digit,
the single working marker and the remaining right-operand prefix in every
round. The explanations in \Cref{m_add10:sec:part2} and~\Cref{m_add10:sec:part3} establish all first-cell,
boundary, transport and return readings. The carry rule processes exactly
the digits permitted in phase $50$. Induction over the transitions of each
round preserves these conditions. \Cref{m_add10:lem:finish} establishes finite
cleanup and the final phase.
\end{proof}

\label{m_add10:sec:common4}
\paragraph{Scanning, Carrying, and Cleanup}\label{m_add10:sec:part2}
A proof phase records geometric context. Phase $i$ pairs expert state
$g_i$ with controller state $q\in Q_i$, with identical grids and heads. An entry $x:(j,w,M)$ means read $x$, write $w$, move $M$, and
advance to phase $j$. The index $j$ denotes a phase. Each $Q_i$ lists the related controller states.

Here $e\in D$ and $h\in\{0,\ldots,8\}$. A digit denotes the corresponding symbol when read or written. Phases $10$--$49$ comprise ten digit-transport families in the next certificate subsection, with $F_d=10+4d$.

\begingroup

\footnotesize\renewcommand{\arraystretch}{1.12}
\setlength{\tabcolsep}{2pt}
\begin{longtable}{r l c l l}
\toprule$i$ & Phase & $g_i$ & $Q_i$ & Read $:$ (next phase, write, move)\\\midrule
\endhead
0 & First left-operand digit & 0 & $\{0\}$ & $e:\triple{1}{e}{R}$\\
1 & Find the input separator & 0 & $\{0\}$ & $e:\triple{1}{e}{R}\quad +:\triple{2}{+}{L}$\\
2 & Descend below units digit & 1 & $\{1\}$ & $e:\triple{3}{e}{D}$\\
3 & Place the working marker & 1 & $\{2\}$ & $\blank:\triple{4}{+}{U}$\\
4 & Ensure a digit above marker & 16 & $\{3\}$ & $e:\triple{6}{e}{R}\quad\blank:\triple{5}{0}{R}$\\
5 & Adjacent existing digit & 2 & $\{4\}$ & $e:\triple{6}{e}{R}$\\
6 & Scan accumulator to separator & 2 & $\{4\}$ & $e:\triple{6}{e}{R}\quad +:\triple{7}{+}{R}$\\
7 & Inspect remaining right operand & 2 & $\{5\}$ & $e:\triple{8}{e}{R}\quad\blank:\triple{54}{\blank}{L}$\\
8 & Scan remaining right operand & 2 & $\{5\}$ & $e:\triple{8}{e}{R}\quad\blank:\triple{9}{\blank}{L}$\\
9 & Read rightmost factor digit & 3 & $\{6\}$ & $d:\triple{F_d}{\blank}{D}\quad(d\in D)$\\
\midrule
50 & Propagate decimal carry & 14 & $\{11\}$ & $9:\triple{50}{0}{L}\quad h:\triple{51}{h+1}{D}$\quad  $\blank:\triple{51}{1}{D}$\\
51 & First return cell after carry & 15 & $\{15\}$ & $\blank:\triple{52}{\blank}{R}$\\
52 & Return to working marker & 15 & $\{15\}$ & $\blank:\triple{52}{\blank}{R}\quad +:\triple{3}{\blank}{L}$\\
53 & Erase and shift marker & 15 & $\{11,15\}$ & $+:\triple{3}{\blank}{L}$\\
54 & Erase input separator & 3 & $\{6\}$ & $+:\triple{55}{\blank}{D}$\\
55 & Begin cleanup below separator & 17 & $\{24\}$ & $\blank:\triple{56}{\blank}{L}$\\
56 & Erase final working marker & 17 & $\{25\}$ & $\blank:\triple{56}{\blank}{L}\quad +:\triple{57}{\blank}{U}$\\
57 & Preserve final digit and stop & 17 & $\{26\}$ & $e:\triple{58}{e}{S}$\\
58 & Halted & 18 & $\{51\}$ & --\\
\bottomrule
\end{longtable}
\endgroup

\textbf{Why These Read Symbols Cover Every Valid Run.}
Both operands contain at least one digit. Initialization therefore reads a digit before
the separator, descends below a digit and places the marker on a blank.
\Cref{m_add10:lem:round} places an existing digit immediately after each zero extension. The scan crosses the contiguous accumulator to the separator.
Immediately to the separator's right, a blank means the next left step
reads that separator; after any digit, the next left step instead reads a digit.

A carry moves strictly left of the marker, so the first return cell on
row $1$ contains a blank. A noncarry addition descends directly onto the marker.
Erasing and shifting the marker reaches a blank. Cleanup
starts below the separator on a blank, walks to the marker, then ascends
to a digit. These facts establish the table's restrictions.

\paragraph{Digit-Transport Phases}\label{m_add10:sec:part3}
For each $d\in D$, four phases carry the selected digit to the accumulator:
\begin{equation*}
F_d=10+4d,\qquad H_d=F_d+1,\qquad W_d=F_d+2,\qquad A_d=F_d+3.
\end{equation*}
All four phases use expert state $4+d$, paired with singleton controller
sets containing $f_d,h_d,w_d,a_d$, respectively. The protocol follows:

\begin{center}
\small\renewcommand{\arraystretch}{1.12}
\begin{tabular}{lll}
\toprule
Phase & Meaning & Read $:$ (next phase, write, move)\\\midrule
$F_d$ & First transport blank & $\blank:\triple{H_d}{\blank}{L}$\\
$H_d$ & Second transport blank & $\blank:\triple{W_d}{\blank}{L}$\\
$W_d$ & Walk left to the marker & $\blank:\triple{W_d}{\blank}{L}\quad +:\triple{A_d}{+}{U}$\\
$A_d$ & Add at the marked column & $e:\triple{53}{d+e}{D}$ if $d+e<10$\\
 & & $e:\triple{50}{d+e-10}{L}$ if $d+e\ge10$\\
\bottomrule
\end{tabular}
\end{center}

\textbf{The Actual Controller-State Numbers.}
 This table expands the entire $40$-phase family.

\begin{center}
\small\renewcommand{\arraystretch}{1.12}\setlength{\tabcolsep}{2pt}
\begin{tabular}{rrrrrr}
\toprule$d$ & Expert $4+d$ & $f_d$ & $h_d$ & $w_d$ & $a_d$\\\midrule
0 & 4 & 27 & 28 & 29 & 30 \\
1 & 5 & 12 & 13 & 14 & 11 \\
2 & 6 & 39 & 40 & 41 & 42 \\
3 & 7 & 31 & 32 & 33 & 34 \\
4 & 8 & 43 & 44 & 45 & 46 \\
5 & 9 & 47 & 48 & 49 & 50 \\
6 & 10 & 7 & 8 & 9 & 10 \\
7 & 11 & 20 & 21 & 22 & 23 \\
8 & 12 & 16 & 17 & 18 & 19 \\
9 & 13 & 35 & 36 & 37 & 38 \\
\bottomrule
\end{tabular}
\end{center}

\textbf{Transport Coverage.}
Equation~\eqref{eq:proof:m-add10.1} gives $r_k-p_k=n+1\ge2$. The first two transport cells therefore contain blanks. After two left
steps, the head reaches either the marker or a blank to the marker's right. The walk reaches the
marker and ascends to a digit ensured at the start of the round.
The two branches implement decimal addition with at most one carry:
\begin{equation*}
d+e=s+10c,\qquad s\in D,\quad c\in\{0,1\}.
\end{equation*}
This accounts for every pair $(d,e)$, including $0$, $9$ and all carry cases.

\textbf{Multiple Roles of Controller State $11$.}
State $11$ performs addition of $1$ when reading a digit, propagates carries when
entered to the left of an overflowing column, and erases the marker when
reading $+$. With $d=1$ and no carry, the controller enters state $15$; every other noncarry branch enters state $11$. Both belong to
$Q_{53}=\{11,15\}$ and have the same required action on $+$.
Every overflow enters state $11\in Q_{50}$.
The phase relation distinguishes these geometric contexts and therefore
does not require a one-to-one correspondence between controller states.

\textbf{Local Transition Identity.}
For every phase $i$, controller state $q\in Q_i$ and listed branch
$x:(j,w,M)$, direct comparison with the transition maps gives
\begin{equation}
\delta_\EXPERT(g_i,x)=(g_j,w,M),\qquad
\delta_\CTRL(q,x)=(q',w,M),\quad q'\in Q_j. \label{eq:proof:m-add10.2}
\end{equation}
Every required entry exists. A transition reaches a halt state exactly
when $j=58$, with $M=S$. The certificate contains $60$ phase--state pairs,
including the terminal pair, and $266$ local checks of \eqref{eq:proof:m-add10.2}. The checks cover
all $223$ defined nonterminal controller entries. The remaining $12$ entries
contain halt loops, excluded from PTOs.

%% file: Appendix/Proof_Tables/mult10.tex
\label{m_mult10:sec:common2}
\paragraph{Copying and Partial Products}
Write $b=b_{n-1}\cdots b_0$, so $B=\sum_{i=0}^{n-1}b_i10^i$.
Assign global weight $10^{m-1-z}$ to column $z$.
Column $m-1$ carries the units digit of the final product.

\begin{lemma}[Copy Staircase]\label{m_mult10:lem:copy}
Before product formation, the expert constructs exactly $n$ occupied rows.
For $0\le i<n$, row $i$ contains
\begin{equation}
a\,x\,b_{n-1}\cdots b_i
\quad\text{starting in column }-i. \label{eq:proof:m-mult10.1}
\end{equation}
All remaining cells contain blanks.
\end{lemma}
\begin{proof}
Row $0$ contains the input. On row $i$, the expert scans to the right blank,
steps left over $b_i$, and inspects the preceding cell. A digit at this
position implies at least two remaining factor digits. The expert copies
every symbol except $b_i$ one row down and one column left, from right to left.
Each destination contains a blank before writing; the source row remains unchanged.

The first descent reaches a blank. Subsequent descents reach the copied
right neighbor: a digit, except when copying the last digit of $a$, where
the lower cell contains the copied $x$. The descent from the source $x$
reaches a digit. After copying the first digit of $a$, the expert reads the
left blank and descends onto the first digit of the new row.
The new row satisfies \eqref{eq:proof:m-mult10.1}, with one fewer factor digit.

When one factor digit remains, the second left step reaches $x$ and starts
product formation. Thus $n-1$ copies produce exactly the stated staircase.
\end{proof}

\begin{lemma}[Aligned Partial Products]\label{m_mult10:lem:products}
After product formation, row $i$ contains the $(m+1)$-digit representation
of $A b_i$ on columns $-i-1,\ldots,m-i-1$.
The leading digit belongs to $\{0,\ldots,8\}$.
In global column weights, this row represents $A b_i10^i$.
All remaining cells contain blanks.
\end{lemma}
\begin{proof}
The expert processes rows from $n-1$ upwards. On row $i$, the expert erases
the selected factor $f=b_i$, the remaining factor prefix, and $x$, then
walks left across $a$. For $f=0$, each digit becomes $0$, followed by one
padding $0$ on the left blank. For $1\le f\le9$, the control state stores
carry $c$, initially $0$, and applies
\begin{equation}
fd+c=w+10c',\qquad
w=(fd+c)\bmod10,\quad c'=\lfloor(fd+c)/10\rfloor. \label{eq:proof:m-mult10.2}
\end{equation}
The bound $0\le c<f$ persists because $fd+c\le9f+f-1=10f-1$.
The usual suffix invariant for \eqref{eq:proof:m-mult10.2} gives the product digits; the left blank
receives the final carry. Thus the leading digit never exceeds $f-1\le8$.

The right scan ends at column $m-i$ and moves up. For $i>0$, the upper cell
contains the last digit of $a$, followed by $x$ and the remaining factor prefix.
The expert scans this prefix, selects the last digit, and repeats.
Every higher row contains at least two factor digits, so at least one further
digit follows the selected factor during leftward erasure.
For $i=0$, the upper cell contains a blank, starting summation on row $0$.
\end{proof}

Every scan crosses a finite word. Copying shortens the factor prefix;
product formation visits each occupied row once. Both passes terminate.

\paragraph{Accumulation and Carry Confinement}
\begin{lemma}[Fixed-Width Accumulator]\label{m_mult10:lem:sum}
After merging through row $k$, $0\le k<n$, all rows above $k$ contain blanks.
Row $k$ contains a contiguous $(m+k+1)$-digit accumulator on columns
$-k-1,\ldots,m-1$, with value
\begin{equation*}
S_k=A\sum_{i=0}^{k}b_i10^i.
\end{equation*}
Rows below $k$ retain the partial products from \Cref{m_mult10:lem:products}.
At inspection of the next row, expert state $88$ has head $(-k-2,k+1)$.
\end{lemma}
\begin{proof}
For $k=0$, the expert scans left across the product on row $0$, reaches
column $-2$, and descends to inspect row $1$.
If row $k+1$ exists, inspection reads the fresh leading digit in
$\{0,\ldots,8\}$. The expert scans the lower product to the right blank at
$m-k-1$ and moves up. The upper cell contains an accumulator digit, since
\begin{equation*}
-k-1\le m-k-1\le m-1.
\end{equation*}
The expert scans to the upper right blank and steps left to column $m-1$.
The lower product ends at $m-k-2$, so the first lower cell contains a blank.

For upper digit $u$, lower digit $v$ (a blank contributes zero), and incoming
carry $c\in\{0,1\}$, the update applies
\begin{equation}
u+v+c=w+10c',\qquad
w\in D,\quad c'\in\{0,1\}. \label{eq:proof:m-mult10.3}
\end{equation}
The expert erases $u$, writes $w$ below, returns to the erased upper cell,
and moves left. The processed lower suffix contains the correct sum suffix;
the unprocessed cells retain the old digits; $c'$ records the contribution
to the next column. Equation~\eqref{eq:proof:m-mult10.3} preserves this invariant.

The lower word extends one column farther left than the upper word.
After the upper word ends, carry zero preserves this lower leading digit.
Carry one increments the leading digit and clears, since the digit never
exceeds $8$. The expert reaches column $-k-3$ on row $k+1$ and descends.
Row $k$ contains only blanks; row $k+1$ represents
$S_k+A b_{k+1}10^{k+1}=S_{k+1}$ on the claimed support.
\end{proof}

\begin{lemma}[Boundary Restrictions]\label{m_mult10:lem:boundary}
A carry-one update beneath an upper digit reads an original lower digit.
A blank lower update guarantees another upper digit on the next left step.
A carry beyond the upper word reads a lower digit in $\{0,\ldots,8\}$ and clears.
\end{lemma}
\begin{proof}
The blank gap lies strictly to the right of the original lower word.
The first update has carry zero, and $u+0\le9$, so no carry originates
inside this gap. Carry one can first arise only inside the lower digit block;
further leftward updates stay inside that block until the leading digit.

A blank lower cell has column $z\ge m-k-1$. The next upper column satisfies
$z-1\ge m-k-2\ge-k-1$, using $m\ge1$, and therefore still contains a digit.
Finally, the lower leading digit lies one column left of the upper word.
\Cref{m_mult10:lem:products} bounds this untouched digit by $8$; adding $1$ clears
any final carry without extending the fixed-width support.
\end{proof}

Each merge consumes finitely many upper digits and at most one extra carry
column. After $n-1$ merges, state $88$ reads the blank below the final row,
moves up to state $106$, and emits $S$ into halt state $107$.
\Cref{m_mult10:lem:sum} gives the final product, padding, and head position.

\label{m_mult10:sec:common3}
\Cref{m_mult10:sec:common4} gives $207$ phases. Phase $i$ pairs expert state $g_i$ with
controller state $q\in Q_i$, on an identical grid and head.
An edge $i\xrightarrow{x}j$ permits read symbol $x$ and advances to phase $j$.
The protocol begins at phase $0$ and ends at phase $206$.
All phase sets contain one controller state except
$Q_{201}=\{26,106\}$.

\begin{lemma}[Protocol Coverage]\label{m_mult10:lem:coverage}
Every expert execution in the stated domain follows the protocol in \Cref{m_mult10:sec:common4}
from phase $0$ to phase $206$.
\end{lemma}
\begin{proof}
The following geometric and arithmetic conditions justify every read restriction.

\textbf{Copying.}
The initial scan and every restart encounter a nonempty left operand,
then $x$, then a nonempty right operand before the right blank.
The second left step reads a factor digit or $x$.
\Cref{m_mult10:lem:copy} distinguishes the first descent (blank), ordinary descents
(digits), and the descent directly above the copied operator ($x$).
Every destination contains a blank. After the first copied symbol, the next
source symbol contains a digit or $x$; no left blank can occur yet.
After copying $x$, the next source symbol contains the last digit of $a$.
The ordinary copy loop eventually reaches the left blank and restarts.

\textbf{Products.}
The bottom row contains one factor digit; erasing that digit immediately
exposes $x$. Every higher row contains at least two factor digits;
erasing the selected factor therefore exposes another digit before $x$.
After erasing $x$, the first product update reads a digit because $m\ge1$.
Later updates read digits followed by the left blank.
Equation~\eqref{eq:proof:m-mult10.2} determines the next carry phase and preserves the carry bound.
The completed product scan reads digits followed by a blank.
\Cref{m_mult10:lem:products} places the head on an upper digit followed by $x$
and a nonempty factor prefix, or on a blank above row $0$.

\textbf{Alignment and Summation.}
Summation starts on the right blank of row $0$. The initial left scan crosses
a nonempty product. Inspection below reads a fresh leading digit from
$\{0,\ldots,8\}$ or a blank below the last row.
After a leading digit, at least one lower digit remains because $m\ge1$.
\Cref{m_mult10:lem:sum} places the first upper alignment step on a digit;
further alignment steps read digits until the right blank.
The first upper units digit has carry zero and a blank lower cell.

After every lower update, the upper return cell contains the erased blank.
A blank lower update leads to a nonblank upper read by
\Cref{m_mult10:lem:boundary}. An update inside the lower digit block leads to
an upper digit or the left blank. Carry-one digit updates remain inside
the lower digit block. At the end of the upper word, carry zero preserves
the untouched lower leading digit, while carry one increments that digit
and clears. Both paths reach the blank immediately left of the merged word.
The final stop reads a blank.

These conditions give exactly the permitted branches used by the expert.
Induction over individual moves, copied rows, and merged rows establishes
coverage. The finite-pass arguments in \Cref{m_mult10:lem:copy}--\Cref{m_mult10:lem:sum}
give termination at phase $206$.
\end{proof}

The protocol may permit additional observation sequences. Checking every
permitted branch therefore suffices for all valid grid executions.

\label{m_mult10:sec:common4}
\textbf{Finite Local Certificate.}
For every protocol edge $i\xrightarrow{x}j$ and every $q\in Q_i$,
direct comparison with the two transition maps gives
\begin{equation}
\delta_\EXPERT(g_i,x)=(g_j,w,M),\qquad
\delta_\CTRL(q,x)=(q',w,M),\qquad q'\in Q_j. \label{eq:proof:m-mult10.4}
\end{equation}
Every required entry exists. A transition reaches halt exactly when
$j=206$, with motion $S$. Direct substitution verifies all $1{,}509$
obligations across $208$ phase--state pairs. These obligations cover all
$1{,}259$ defined nonterminal controller entries. The remaining $12$ defined
entries contain halt loops, which never contribute to a PTO.

\input{Appendix/Proof_Tables/mult10_expert_map}
\paragraph{Phases and Controller-State Relation}
The table gives the phase index $i$, expert state $g_i$, controller set $Q_i$,
and permitted reads with successor phase indices.
For a digit vector $[j_0,\ldots,j_9]$, read $d$ advances to $j_d$.
The notation $D:j$ sends every decimal digit to phase $j$;
$0{:}8:j$ permits only digits $0,\ldots,8$.
Unlisted reads never occur in the corresponding phase of a valid input.

The expert transition supplies the write and motion for each edge;
Equation~\eqref{eq:proof:m-mult10.4} checks both transition maps against that action. The copying moves
follow \Cref{m_mult10:lem:copy}; product updates follow \eqref{eq:proof:m-mult10.2}; addition updates
follow \eqref{eq:proof:m-mult10.3}. Phase labels describe positions before the next action.
\begingroup
\footnotesize
\setlength{\tabcolsep}{2pt}
\renewcommand{\arraystretch}{1.12}
\begin{longtable}{r l c l l}
\toprule
$i$ & Phase & $g_i$ & $Q_i$ & Read $:$ next phase\\\midrule
\endfirsthead
\multicolumn{5}{l}{\small\sffamily Simulation Certificate (Continued)}\\[5pt]
\toprule
$i$ & Phase & $g_i$ & $Q_i$ & Read $:$ next phase\\\midrule
\endhead
\midrule\multicolumn{5}{r}{\scriptsize Continued on the next page}\\
\endfoot
\bottomrule
\endlastfoot
0 & First left-operand digit & 0 & $\{0\}$ & $D:1$ \\
1 & Scan left operand & 0 & $\{0\}$ & $D:1\quad x:2$ \\
2 & First factor digit & 0 & $\{1\}$ & $D:3$ \\
3 & Scan right operand & 0 & $\{1\}$ & $D:3\quad \blank:4$ \\
4 & Skip last factor digit & 1 & $\{2\}$ & $D:5$ \\
5 & Copy or start products & 2 & $\{3\}$ & $D:[6, 8, 10, 12, 14, 16, 18, 20, 22, 24]$\quad $x:62$ \\
6 & First descent, digit 0 & 3 & $\{172\}$ & $\blank:7$ \\
7 & First copied write, digit 0 & 14 & $\{173\}$ & $\blank:26$ \\
8 & First descent, digit 1 & 4 & $\{123\}$ & $\blank:9$ \\
9 & First copied write, digit 1 & 15 & $\{124\}$ & $\blank:26$ \\
10 & First descent, digit 2 & 5 & $\{64\}$ & $\blank:11$ \\
11 & First copied write, digit 2 & 16 & $\{65\}$ & $\blank:26$ \\
12 & First descent, digit 3 & 6 & $\{4\}$ & $\blank:13$ \\
13 & First copied write, digit 3 & 17 & $\{5\}$ & $\blank:26$ \\
14 & First descent, digit 4 & 7 & $\{87\}$ & $\blank:15$ \\
15 & First copied write, digit 4 & 18 & $\{88\}$ & $\blank:26$ \\
16 & First descent, digit 5 & 8 & $\{39\}$ & $\blank:17$ \\
17 & First copied write, digit 5 & 19 & $\{40\}$ & $\blank:26$ \\
18 & First descent, digit 6 & 9 & $\{163\}$ & $\blank:19$ \\
19 & First copied write, digit 6 & 20 & $\{164\}$ & $\blank:26$ \\
20 & First descent, digit 7 & 10 & $\{51\}$ & $\blank:21$ \\
21 & First copied write, digit 7 & 21 & $\{52\}$ & $\blank:26$ \\
22 & First descent, digit 8 & 11 & $\{128\}$ & $\blank:23$ \\
23 & First copied write, digit 8 & 22 & $\{129\}$ & $\blank:26$ \\
24 & First descent, digit 9 & 12 & $\{79\}$ & $\blank:25$ \\
25 & First copied write, digit 9 & 23 & $\{80\}$ & $\blank:26$ \\
26 & Source after first copy & 25 & $\{6\}$ & $D:[27, 29, 31, 33, 35, 37, 39, 41, 43, 45]$\quad $x:48$ \\
27 & Ordinary descent, digit 0 & 3 & $\{178\}$ & $D:28$ \\
28 & Ordinary copied write, digit 0 & 14 & $\{159\}$ & $\blank:47$ \\
29 & Ordinary descent, digit 1 & 4 & $\{134\}$ & $D:30$ \\
30 & Ordinary copied write, digit 1 & 15 & $\{135\}$ & $\blank:47$ \\
31 & Ordinary descent, digit 2 & 5 & $\{68\}$ & $D:32$ \\
32 & Ordinary copied write, digit 2 & 16 & $\{67\}$ & $\blank:47$ \\
33 & Ordinary descent, digit 3 & 6 & $\{125\}$ & $D:34$ \\
34 & Ordinary copied write, digit 3 & 17 & $\{11\}$ & $\blank:47$ \\
35 & Ordinary descent, digit 4 & 7 & $\{166\}$ & $D:36$ \\
36 & Ordinary copied write, digit 4 & 18 & $\{90\}$ & $\blank:47$ \\
37 & Ordinary descent, digit 5 & 8 & $\{165\}$ & $D:38$ \\
38 & Ordinary copied write, digit 5 & 19 & $\{131\}$ & $\blank:47$ \\
39 & Ordinary descent, digit 6 & 9 & $\{55\}$ & $D:40$ \\
40 & Ordinary copied write, digit 6 & 20 & $\{42\}$ & $\blank:47$ \\
41 & Ordinary descent, digit 7 & 10 & $\{170\}$ & $D:42$ \\
42 & Ordinary copied write, digit 7 & 21 & $\{148\}$ & $\blank:47$ \\
43 & Ordinary descent, digit 8 & 11 & $\{150\}$ & $D:44$ \\
44 & Ordinary copied write, digit 8 & 22 & $\{116\}$ & $\blank:47$ \\
45 & Ordinary descent, digit 9 & 12 & $\{167\}$ & $D:46$ \\
46 & Ordinary copied write, digit 9 & 23 & $\{54\}$ & $\blank:47$ \\
47 & Continue source copy & 25 & $\{12\}$ & $D:[27, 29, 31, 33, 35, 37, 39, 41, 43, 45]$\quad $x:48\quad \blank:61$ \\
48 & Descend below operator & 13 & $\{7\}$ & $D:49$ \\
49 & Write copied operator & 24 & $\{8\}$ & $\blank:50$ \\
50 & Copy last digit of $a$ & 25 & $\{9\}$ & $D:[51, 52, 53, 54, 55, 56, 57, 58, 59, 60]$ \\
51 & Descend above $x$, digit 0 & 3 & $\{158\}$ & $x:28$ \\
52 & Descend above $x$, digit 1 & 4 & $\{145\}$ & $x:30$ \\
53 & Descend above $x$, digit 2 & 5 & $\{66\}$ & $x:32$ \\
54 & Descend above $x$, digit 3 & 6 & $\{10\}$ & $x:34$ \\
55 & Descend above $x$, digit 4 & 7 & $\{89\}$ & $x:36$ \\
56 & Descend above $x$, digit 5 & 8 & $\{130\}$ & $x:38$ \\
57 & Descend above $x$, digit 6 & 9 & $\{41\}$ & $x:40$ \\
58 & Descend above $x$, digit 7 & 10 & $\{147\}$ & $x:42$ \\
59 & Descend above $x$, digit 8 & 11 & $\{115\}$ & $x:44$ \\
60 & Descend above $x$, digit 9 & 12 & $\{53\}$ & $x:46$ \\
61 & Restart on copied row & 0 & $\{13\}$ & $D:1$ \\
62 & Bottom-row factor & 26 & $\{14\}$ & $D:[63, 66, 69, 73, 78, 84, 91, 99, 108, 118]$ \\
63 & Erase $x$, bottom factor 0 & 27 & $\{111\}$ & $x:64$ \\
64 & First product digit, factor 0 & 37 & $\{62\}$ & $D:65$ \\
65 & Continue product by zero & 37 & $\{62\}$ & $D:65\quad \blank:129$ \\
66 & Erase $x$, bottom factor 1 & 28 & $\{78\}$ & $x:67$ \\
67 & First product digit, factor 1 & 38 & $\{24\}$ & $D:68$ \\
68 & Product by 1, carry 0 & 38 & $\{24\}$ & $D:68\quad \blank:129$ \\
69 & Erase $x$, bottom factor 2 & 29 & $\{69\}$ & $x:70$ \\
70 & First product digit, factor 2 & 39 & $\{70\}$ & $D:[71, 71, 71, 71, 71, 72, 72, 72, 72, 72]$ \\
71 & Product by 2, carry 0 & 39 & $\{70\}$ & $D:[71, 71, 71, 71, 71, 72, 72, 72, 72, 72]$\quad $\blank:129$ \\
72 & Product by 2, carry 1 & 40 & $\{143\}$ & $D:[71, 71, 71, 71, 71, 72, 72, 72, 72, 72]$\quad $\blank:129$ \\
73 & Erase $x$, bottom factor 3 & 30 & $\{15\}$ & $x:74$ \\
74 & First product digit, factor 3 & 41 & $\{16\}$ & $D:[75, 75, 75, 75, 76, 76, 76, 77, 77, 77]$ \\
75 & Product by 3, carry 0 & 41 & $\{16\}$ & $D:[75, 75, 75, 75, 76, 76, 76, 77, 77, 77]$\quad $\blank:129$ \\
76 & Product by 3, carry 1 & 42 & $\{48\}$ & $D:[75, 75, 75, 76, 76, 76, 76, 77, 77, 77]$\quad $\blank:129$ \\
77 & Product by 3, carry 2 & 43 & $\{155\}$ & $D:[75, 75, 75, 76, 76, 76, 77, 77, 77, 77]$\quad $\blank:129$ \\
78 & Erase $x$, bottom factor 4 & 31 & $\{75\}$ & $x:79$ \\
79 & First product digit, factor 4 & 44 & $\{76\}$ & $D:[80, 80, 80, 81, 81, 82, 82, 82, 83, 83]$ \\
80 & Product by 4, carry 0 & 44 & $\{76\}$ & $D:[80, 80, 80, 81, 81, 82, 82, 82, 83, 83]$\quad $\blank:129$ \\
81 & Product by 4, carry 1 & 45 & $\{91\}$ & $D:[80, 80, 80, 81, 81, 82, 82, 82, 83, 83]$\quad $\blank:129$ \\
82 & Product by 4, carry 2 & 46 & $\{77\}$ & $D:[80, 80, 81, 81, 81, 82, 82, 83, 83, 83]$\quad $\blank:129$ \\
83 & Product by 4, carry 3 & 47 & $\{117\}$ & $D:[80, 80, 81, 81, 81, 82, 82, 83, 83, 83]$\quad $\blank:129$ \\
84 & Erase $x$, bottom factor 5 & 32 & $\{43\}$ & $x:85$ \\
85 & First product digit, factor 5 & 48 & $\{44\}$ & $D:[86, 86, 87, 87, 88, 88, 89, 89, 90, 90]$ \\
86 & Product by 5, carry 0 & 48 & $\{44\}$ & $D:[86, 86, 87, 87, 88, 88, 89, 89, 90, 90]$\quad $\blank:129$ \\
87 & Product by 5, carry 1 & 49 & $\{153\}$ & $D:[86, 86, 87, 87, 88, 88, 89, 89, 90, 90]$\quad $\blank:129$ \\
88 & Product by 5, carry 2 & 50 & $\{144\}$ & $D:[86, 86, 87, 87, 88, 88, 89, 89, 90, 90]$\quad $\blank:129$ \\
89 & Product by 5, carry 3 & 51 & $\{45\}$ & $D:[86, 86, 87, 87, 88, 88, 89, 89, 90, 90]$\quad $\blank:129$ \\
90 & Product by 5, carry 4 & 52 & $\{95\}$ & $D:[86, 86, 87, 87, 88, 88, 89, 89, 90, 90]$\quad $\blank:129$ \\
91 & Erase $x$, bottom factor 6 & 33 & $\{107\}$ & $x:92$ \\
92 & First product digit, factor 6 & 53 & $\{108\}$ & $D:[93, 93, 94, 94, 95, 96, 96, 97, 97, 98]$ \\
93 & Product by 6, carry 0 & 53 & $\{108\}$ & $D:[93, 93, 94, 94, 95, 96, 96, 97, 97, 98]$\quad $\blank:129$ \\
94 & Product by 6, carry 1 & 54 & $\{171\}$ & $D:[93, 93, 94, 94, 95, 96, 96, 97, 97, 98]$\quad $\blank:129$ \\
95 & Product by 6, carry 2 & 55 & $\{110\}$ & $D:[93, 93, 94, 95, 95, 96, 96, 97, 98, 98]$\quad $\blank:129$ \\
96 & Product by 6, carry 3 & 56 & $\{174\}$ & $D:[93, 93, 94, 95, 95, 96, 96, 97, 98, 98]$\quad $\blank:129$ \\
97 & Product by 6, carry 4 & 57 & $\{142\}$ & $D:[93, 94, 94, 95, 95, 96, 97, 97, 98, 98]$\quad $\blank:129$ \\
98 & Product by 6, carry 5 & 58 & $\{109\}$ & $D:[93, 94, 94, 95, 95, 96, 97, 97, 98, 98]$\quad $\blank:129$ \\
99 & Erase $x$, bottom factor 7 & 34 & $\{56\}$ & $x:100$ \\
100 & First product digit, factor 7 & 59 & $\{57\}$ & $D:[101, 101, 102, 103, 103, 104, 105, 105, 106, 107]$ \\
101 & Product by 7, carry 0 & 59 & $\{57\}$ & $D:[101, 101, 102, 103, 103, 104, 105, 105, 106, 107]$\quad $\blank:129$ \\
102 & Product by 7, carry 1 & 60 & $\{168\}$ & $D:[101, 101, 102, 103, 103, 104, 105, 106, 106, 107]$\quad $\blank:129$ \\
103 & Product by 7, carry 2 & 61 & $\{127\}$ & $D:[101, 101, 102, 103, 104, 104, 105, 106, 106, 107]$\quad $\blank:129$ \\
104 & Product by 7, carry 3 & 62 & $\{179\}$ & $D:[101, 102, 102, 103, 104, 104, 105, 106, 106, 107]$\quad $\blank:129$ \\
105 & Product by 7, carry 4 & 63 & $\{59\}$ & $D:[101, 102, 102, 103, 104, 104, 105, 106, 107, 107]$\quad $\blank:129$ \\
106 & Product by 7, carry 5 & 64 & $\{169\}$ & $D:[101, 102, 102, 103, 104, 105, 105, 106, 107, 107]$\quad $\blank:129$ \\
107 & Product by 7, carry 6 & 65 & $\{58\}$ & $D:[101, 102, 103, 103, 104, 105, 105, 106, 107, 107]$\quad $\blank:129$ \\
108 & Erase $x$, bottom factor 8 & 35 & $\{112\}$ & $x:109$ \\
109 & First product digit, factor 8 & 66 & $\{113\}$ & $D:[110, 110, 111, 112, 113, 114, 114, 115, 116, 117]$ \\
110 & Product by 8, carry 0 & 66 & $\{113\}$ & $D:[110, 110, 111, 112, 113, 114, 114, 115, 116, 117]$\quad $\blank:129$ \\
111 & Product by 8, carry 1 & 67 & $\{137\}$ & $D:[110, 110, 111, 112, 113, 114, 114, 115, 116, 117]$\quad $\blank:129$ \\
112 & Product by 8, carry 2 & 68 & $\{121\}$ & $D:[110, 111, 111, 112, 113, 114, 115, 115, 116, 117]$\quad $\blank:129$ \\
113 & Product by 8, carry 3 & 69 & $\{114\}$ & $D:[110, 111, 111, 112, 113, 114, 115, 115, 116, 117]$\quad $\blank:129$ \\
114 & Product by 8, carry 4 & 70 & $\{132\}$ & $D:[110, 111, 112, 112, 113, 114, 115, 116, 116, 117]$\quad $\blank:129$ \\
115 & Product by 8, carry 5 & 71 & $\{149\}$ & $D:[110, 111, 112, 112, 113, 114, 115, 116, 116, 117]$\quad $\blank:129$ \\
116 & Product by 8, carry 6 & 72 & $\{120\}$ & $D:[110, 111, 112, 113, 113, 114, 115, 116, 117, 117]$\quad $\blank:129$ \\
117 & Product by 8, carry 7 & 73 & $\{136\}$ & $D:[110, 111, 112, 113, 113, 114, 115, 116, 117, 117]$\quad $\blank:129$ \\
118 & Erase $x$, bottom factor 9 & 36 & $\{81\}$ & $x:119$ \\
119 & First product digit, factor 9 & 74 & $\{82\}$ & $D:[120, 120, 121, 122, 123, 124, 125, 126, 127, 128]$ \\
120 & Product by 9, carry 0 & 74 & $\{82\}$ & $D:[120, 120, 121, 122, 123, 124, 125, 126, 127, 128]$\quad $\blank:129$ \\
121 & Product by 9, carry 1 & 75 & $\{151\}$ & $D:[120, 121, 121, 122, 123, 124, 125, 126, 127, 128]$\quad $\blank:129$ \\
122 & Product by 9, carry 2 & 76 & $\{177\}$ & $D:[120, 121, 122, 122, 123, 124, 125, 126, 127, 128]$\quad $\blank:129$ \\
123 & Product by 9, carry 3 & 77 & $\{154\}$ & $D:[120, 121, 122, 123, 123, 124, 125, 126, 127, 128]$\quad $\blank:129$ \\
124 & Product by 9, carry 4 & 78 & $\{176\}$ & $D:[120, 121, 122, 123, 124, 124, 125, 126, 127, 128]$\quad $\blank:129$ \\
125 & Product by 9, carry 5 & 79 & $\{83\}$ & $D:[120, 121, 122, 123, 124, 125, 125, 126, 127, 128]$\quad $\blank:129$ \\
126 & Product by 9, carry 6 & 80 & $\{156\}$ & $D:[120, 121, 122, 123, 124, 125, 126, 126, 127, 128]$\quad $\blank:129$ \\
127 & Product by 9, carry 7 & 81 & $\{152\}$ & $D:[120, 121, 122, 123, 124, 125, 126, 127, 127, 128]$\quad $\blank:129$ \\
128 & Product by 9, carry 8 & 82 & $\{98\}$ & $D:[120, 121, 122, 123, 124, 125, 126, 127, 128, 128]$\quad $\blank:129$ \\
129 & Scan completed product & 83 & $\{17\}$ & $D:129\quad \blank:130$ \\
130 & Inspect row above & 84 & $\{18\}$ & $D:131\quad \blank:155$ \\
131 & Cross upper operator & 85 & $\{19\}$ & $x:132$ \\
132 & First upper factor digit & 85 & $\{20\}$ & $D:133$ \\
133 & Scan upper factor prefix & 85 & $\{20\}$ & $D:133\quad \blank:134$ \\
134 & Select upper factor & 26 & $\{21\}$ & $D:[135, 137, 139, 141, 143, 145, 147, 149, 151, 153]$ \\
135 & First prefix erasure, factor 0 & 27 & $\{60\}$ & $D:136$ \\
136 & Later prefix erasure, factor 0 & 27 & $\{61\}$ & $D:136\quad x:64$ \\
137 & First prefix erasure, factor 1 & 28 & $\{22\}$ & $D:138$ \\
138 & Later prefix erasure, factor 1 & 28 & $\{23\}$ & $D:138\quad x:67$ \\
139 & First prefix erasure, factor 2 & 29 & $\{71\}$ & $D:140$ \\
140 & Later prefix erasure, factor 2 & 29 & $\{72\}$ & $D:140\quad x:70$ \\
141 & First prefix erasure, factor 3 & 30 & $\{46\}$ & $D:142$ \\
142 & Later prefix erasure, factor 3 & 30 & $\{47\}$ & $D:142\quad x:74$ \\
143 & First prefix erasure, factor 4 & 31 & $\{92\}$ & $D:144$ \\
144 & Later prefix erasure, factor 4 & 31 & $\{93\}$ & $D:144\quad x:79$ \\
145 & First prefix erasure, factor 5 & 32 & $\{84\}$ & $D:146$ \\
146 & Later prefix erasure, factor 5 & 32 & $\{85\}$ & $D:146\quad x:85$ \\
147 & First prefix erasure, factor 6 & 33 & $\{140\}$ & $D:148$ \\
148 & Later prefix erasure, factor 6 & 33 & $\{141\}$ & $D:148\quad x:92$ \\
149 & First prefix erasure, factor 7 & 34 & $\{160\}$ & $D:150$ \\
150 & Later prefix erasure, factor 7 & 34 & $\{161\}$ & $D:150\quad x:100$ \\
151 & First prefix erasure, factor 8 & 35 & $\{118\}$ & $D:152$ \\
152 & Later prefix erasure, factor 8 & 35 & $\{119\}$ & $D:152\quad x:109$ \\
153 & First prefix erasure, factor 9 & 36 & $\{96\}$ & $D:154$ \\
154 & Later prefix erasure, factor 9 & 36 & $\{97\}$ & $D:154\quad x:119$ \\
155 & Start summation & 86 & $\{25\}$ & $\blank:156$ \\
156 & First initial left digit & 87 & $\{26\}$ & $D:157$ \\
157 & Continue initial left scan & 87 & $\{26\}$ & $D:157\quad \blank:158$ \\
158 & Inspect fresh lower row & 88 & $\{27\}$ & $0{:}8:159\quad \blank:205$ \\
159 & First digit after leading cell & 89 & $\{28\}$ & $D:160$ \\
160 & Scan lower product & 89 & $\{28\}$ & $D:160\quad \blank:161$ \\
161 & First upper alignment digit & 104 & $\{29\}$ & $D:162$ \\
162 & Continue upper alignment & 104 & $\{30\}$ & $D:162\quad \blank:163$ \\
163 & Read upper units digit & 90 & $\{31\}$ & $D:[164, 165, 166, 167, 168, 169, 170, 171, 172, 173]$ \\
164 & First lower update, value 0 & 92 & $\{63\}$ & $\blank:174$ \\
165 & First lower update, value 1 & 93 & $\{99\}$ & $\blank:174$ \\
166 & First lower update, value 2 & 94 & $\{138\}$ & $\blank:174$ \\
167 & First lower update, value 3 & 95 & $\{32\}$ & $\blank:174$ \\
168 & First lower update, value 4 & 96 & $\{73\}$ & $\blank:174$ \\
169 & First lower update, value 5 & 97 & $\{133\}$ & $\blank:174$ \\
170 & First lower update, value 6 & 98 & $\{94\}$ & $\blank:174$ \\
171 & First lower update, value 7 & 99 & $\{162\}$ & $\blank:174$ \\
172 & First lower update, value 8 & 100 & $\{49\}$ & $\blank:174$ \\
173 & First lower update, value 9 & 101 & $\{146\}$ & $\blank:174$ \\
174 & Return after lower blank & 91 & $\{33\}$ & $\blank:175$ \\
175 & Read next upper digit & 90 & $\{34\}$ & $D:[178, 179, 180, 181, 182, 183, 184, 185, 186, 187]$ \\
176 & Return with carry zero & 91 & $\{36\}$ & $\blank:177$ \\
177 & Read upper, carry zero & 90 & $\{37\}$ & $D:[178, 179, 180, 181, 182, 183, 184, 185, 186, 187]$\quad $\blank:200$ \\
178 & Add 0 below & 92 & $\{35\}$ & $D:176\quad \blank:174$ \\
179 & Add 1 below & 93 & $\{50\}$ & $D:[176, 176, 176, 176, 176, 176, 176, 176, 176, 188]$\quad $\blank:174$ \\
180 & Add 2 below & 94 & $\{122\}$ & $D:[176, 176, 176, 176, 176, 176, 176, 176, 188, 188]$\quad $\blank:174$ \\
181 & Add 3 below & 95 & $\{86\}$ & $D:[176, 176, 176, 176, 176, 176, 176, 188, 188, 188]$\quad $\blank:174$ \\
182 & Add 4 below & 96 & $\{74\}$ & $D:[176, 176, 176, 176, 176, 176, 188, 188, 188, 188]$\quad $\blank:174$ \\
183 & Add 5 below & 97 & $\{139\}$ & $D:[176, 176, 176, 176, 176, 188, 188, 188, 188, 188]$\quad $\blank:174$ \\
184 & Add 6 below & 98 & $\{157\}$ & $D:[176, 176, 176, 176, 188, 188, 188, 188, 188, 188]$\quad $\blank:174$ \\
185 & Add 7 below & 99 & $\{175\}$ & $D:[176, 176, 176, 188, 188, 188, 188, 188, 188, 188]$\quad $\blank:174$ \\
186 & Add 8 below & 100 & $\{100\}$ & $D:[176, 176, 188, 188, 188, 188, 188, 188, 188, 188]$\quad $\blank:174$ \\
187 & Add 9 below & 101 & $\{126\}$ & $D:[176, 188, 188, 188, 188, 188, 188, 188, 188, 188]$\quad $\blank:174$ \\
188 & Return with carry one & 103 & $\{101\}$ & $\blank:189$ \\
189 & Read upper, carry one & 102 & $\{102\}$ & $D:[190, 191, 192, 193, 194, 195, 196, 197, 198, 199]$\quad $\blank:202$ \\
190 & Add 1 inside lower block & 93 & $\{50\}$ & $D:[176, 176, 176, 176, 176, 176, 176, 176, 176, 188]$ \\
191 & Add 2 inside lower block & 94 & $\{122\}$ & $D:[176, 176, 176, 176, 176, 176, 176, 176, 188, 188]$ \\
192 & Add 3 inside lower block & 95 & $\{86\}$ & $D:[176, 176, 176, 176, 176, 176, 176, 188, 188, 188]$ \\
193 & Add 4 inside lower block & 96 & $\{74\}$ & $D:[176, 176, 176, 176, 176, 176, 188, 188, 188, 188]$ \\
194 & Add 5 inside lower block & 97 & $\{139\}$ & $D:[176, 176, 176, 176, 176, 188, 188, 188, 188, 188]$ \\
195 & Add 6 inside lower block & 98 & $\{157\}$ & $D:[176, 176, 176, 176, 188, 188, 188, 188, 188, 188]$ \\
196 & Add 7 inside lower block & 99 & $\{175\}$ & $D:[176, 176, 176, 188, 188, 188, 188, 188, 188, 188]$ \\
197 & Add 8 inside lower block & 100 & $\{100\}$ & $D:[176, 176, 188, 188, 188, 188, 188, 188, 188, 188]$ \\
198 & Add 9 inside lower block & 101 & $\{126\}$ & $D:[176, 188, 188, 188, 188, 188, 188, 188, 188, 188]$ \\
199 & Add 10 inside lower block & 105 & $\{180\}$ & $D:188$ \\
200 & Preserve leading digit & 87 & $\{26\}$ & $0{:}8:201$ \\
201 & Blank left of merged row & 87 & $\{26,106\}$ & $\blank:158$ \\
202 & Increment leading digit & 93 & $\{103\}$ & $0{:}8:203$ \\
203 & Return after final carry & 91 & $\{104\}$ & $\blank:204$ \\
204 & Step down left of sum & 90 & $\{105\}$ & $\blank:201$ \\
205 & Emit final stop & 106 & $\{38\}$ & $\blank:206$ \\
206 & Halted & 107 & $\{181\}$ & $\text{--}$ \\
\end{longtable}
\endgroup

%% file: Appendix/Proof_Tables/mult10_expert_map.tex
\paragraph{The Expert Transition Map.}
The following formulas supply every action needed by the decimal certificate. Let $d,f\in\{0,\ldots,9\}$ and $z\in\{\blank,x,0,\ldots,9\}$. Define
\begin{equation*}
P_{f,c}=38+\frac{f(f-1)}2+c\quad(1\le f\le9,\ 0\le c<f),
\qquad V_v=\begin{cases}92+v,&0\le v<10,\\105,&v=10.\end{cases}
\end{equation*}
An entry $(q,x)\mapsto(q',w,M)$ specifies an actual state transition. A symbol $d$ in a written field denotes the corresponding decimal digit.
\begingroup
\footnotesize\setlength{\tabcolsep}{5pt}\renewcommand{\arraystretch}{1.1}
\begin{longtable}{llp{8.4cm}}
\toprule
State & Read & $(\text{successor},\text{write},\text{motion})$\\\midrule
\endfirsthead
\toprule State & Read & $(\text{successor},\text{write},\text{motion})$\\\midrule
\endhead
\bottomrule\endfoot
$0$ & $d,x$ & $(0,\text{read},R)$\\
$0$ & $\blank$ & $(1,\blank,L)$\\
$1$ & $d$ & $(2,d,L)$\\
$2$ & $d$ & $(3+d,d,D)$\\
$2$ & $x$ & $(26,x,R)$\\
$3+d$ & $z$ & $(14+d,z,L)$\\
$13$ & $d$ & $(24,d,L)$\\
$14+d$ & $\blank$ & $(25,d,U)$\\
$24$ & $\blank$ & $(25,x,U)$\\
$25$ & $d$ & $(3+d,d,D)$\\
$25$ & $x$ & $(13,x,D)$\\
$25$ & $\blank$ & $(0,\blank,D)$\\
$26$ & $f$ & $(27+f,\blank,L)$\\
$27+f$ & $d$ & $(27+f,\blank,L)$\\
$27$ & $x$ & $(37,\blank,L)$\\
$27+f$, $f>0$ & $x$ & $(P_{f,0},\blank,L)$\\
$37$ & $d$ & $(37,0,L)$\\
$37$ & $\blank$ & $(83,0,R)$\\
$P_{f,c}$ & $d$ & $(P_{f,c'},w,L)$, where $fd+c=10c'+w$, $0\le w<10$\\
$P_{f,c}$ & $\blank$ & $(83,c,R)$\\
$83$ & $d$ & $(83,d,R)$\\
$83$ & $\blank$ & $(84,\blank,U)$\\
$84$ & $d$ & $(85,d,R)$\\
$84$ & $\blank$ & $(86,\blank,D)$\\
$85$ & $d,x$ & $(85,\text{read},R)$\\
$85$ & $\blank$ & $(26,\blank,L)$\\
$86$ & $\blank$ & $(87,\blank,L)$\\
$87$ & $d$ & $(87,d,L)$\\
$87$ & $\blank$ & $(88,\blank,D)$\\
$88$ & $d$ & $(89,d,R)$\\
$88$ & $\blank$ & $(106,\blank,U)$\\
$89$ & $d$ & $(89,d,R)$\\
$89$ & $\blank$ & $(104,\blank,U)$\\
$104$ & $d$ & $(104,d,R)$\\
$104$ & $\blank$ & $(90,\blank,L)$\\
$90$ & $d$ & $(V_d,\blank,D)$\\
$90$ & $\blank$ & $(87,\blank,D)$\\
$102$ & $d$ & $(V_{d+1},\blank,D)$\\
$102$ & $\blank$ & $(93,\blank,D)$\\
$91$ & $\blank$ & $(90,\blank,L)$\\
$103$ & $\blank$ & $(102,\blank,L)$\\
$V_v$ & $d$ or $\blank$ & $(91+12c',w,U)$, where $v+d=10c'+w$ and a blank contributes $d=0$\\
$106$ & $z$ & $(107,z,S)$\\
\end{longtable}
\endgroup
The phase restrictions below exclude every omitted expert transition. In particular, a final carry only encounters a lower leading digit at most $8$.

%% file: Appendix/Proof_Tables/universal_shared.tex
\label{universal:sec:common2}
\paragraph{Encoding and Rune Semantics}\label{universal:sec:encoding}
Let $\kappa=\max(1,\lceil\log_3 N\rceil)$. Encode each state with a fixed-width,
most-significant-first ternary word using $L,R,S$ for $0,1,2$ respectively.
Write this word as $c(q)$. The trit symbol $S$ encodes the value $2$; the motion $S$ stops execution.

Order tape symbols as $\sigma_0=\blank<\sigma_1=0<\sigma_2=1$.
For $\delta(q,\sigma_i)=(q',w,M)$, program row $r=3q+i$ contains
\begin{equation}
\bigl[\sigma_i,\blank,c(q),0,c(q'),0,w,M\bigr]. \label{eq:proof:universal.1}
\end{equation}
The source and successor codes each occupy $\kappa$ separate cells. Thus the
program has $3N$ rows, ordered by state and then by read symbol.

\begin{center}
\renewcommand{\arraystretch}{1.12}\small
\begin{tabular}{lll}
\toprule
Columns & Field & Purpose\\\midrule
$0$ & $\sigma_i$ & Simulated tape read symbol\\
$1$ & $\blank$ & Bookmark column for rune $X$\\
$2,\ldots,\kappa+1$ & $c(q)$ & Source-state code\\
$\kappa+2$ & $0$ & First delimiter\\
$\kappa+3,\ldots,2\kappa+2$ & $c(q')$ & Successor-state code\\
$2\kappa+3$ & $0$ & Second delimiter\\
$2\kappa+4$ & $w$ & Symbol to write\\
$2\kappa+5$ & $M$ & Simulated move $L,R$ or $S$\\
\bottomrule
\end{tabular}
\end{center}

\textbf{Initial Placement.}
Row $r$ occupies grid height $y=-r$, so $D$ advances down the table and
$U$ moves up the table. Initially $X$ covers the blank at $(1,0)$; no cell
contains $Y$ or $Z$.
The input occupies columns $1,\ldots,|v|$ of row $3N$, at height $-3N$.
The controller starts at $(1,-3N)$, including for empty input.
The tape row extends infinitely in both horizontal directions.

\textbf{Rune Semantics.}
Each of $X,Y,Z$ has one position and a hidden symbol. Writing a rune first
restores the hidden symbol at the old rune position, then installs the rune over
the symbol at the new position. Rewriting a rune in place preserves the hidden symbol. Writing an ordinary symbol at a rune removes that rune.
A motion $X$, $Y$ or $Z$ jumps to the corresponding rune after the write.
Ordinary motions move one cell; $S$ stops without moving.

The \emph{logical program} means the program with every rune replaced by the corresponding hidden symbol. Correctness depends on preserving this logical program,
not on treating a rune as a destructive overwrite.

\paragraph{Endpoint and Prefix Invariant}\label{universal:sec:part2}
\begin{lemma}[Endpoint and Prefix Invariant]\label{universal:lem:geometry}
At the start of every simulated step, the expert occupies state $0$ at the
simulated tape head. Rune $X$ occupies column $1$ on the first or last row
of the current simulated state's three-row block. After matching $j$
successor trits, $X$ occupies column $1+j$ on a row with the correct
length-$j$ successor prefix. The row index remains congruent to $0$ or $2$
modulo $3$. The logical program remains unchanged.
\end{lemma}
\begin{proof}
Initially $X$ occupies the first row of state $0$. Assume the invariant at
the start of a simulated step. The expert reads the tape symbol, installs
$Z$, jumps to $X$, and steps left. The symbol search selects the matching
row within the current state's three-row block. The following right scan
crosses column $1$ and the source code to the first delimiter, leaving $X$
in place.

For successor trit $t$, the expert places $Y$ on the trit, jumps to $X$, and
steps right to the next source-code column. All rows sharing the matched
prefix form one interval. Within that interval, next-trit values increase
monotonically down the table. The successor state exists, so a nonempty
subinterval contains next trit $t$. Moving down on a smaller trit, or up
on a larger trit, reaches $t$ before leaving the matched-prefix interval.

An immediate match preserves the endpoint. A downward search enters the
target subinterval at the first row, with read symbol $\blank$; an upward
search enters at the last row, with read symbol $1$. Both rows occupy
endpoints of three-row state blocks. Writing $X$ advances the matched
prefix by one trit. Jumping to $Y$ resumes instruction decoding.

After all $\kappa$ trits, $X$ occupies a row in the successor state's block.
The second delimiter triggers a leftward rewind of $X$ to column $1$ on
the same row. The expert then applies the prescribed write and motion at
$Z$. Rune relocation restores every covered program symbol, preserving
the logical program. Each nonstopping motion establishes the next simulated step.
\end{proof}

\begin{lemma}[No Missing Middle Trit]\label{universal:lem:middle}
Within a fixed source-code prefix block, a vertical search cannot pass
directly between an $L$ region and an $S$ region. A nonempty $R$ region
separates the two regions.
\end{lemma}
\begin{proof}
The state numbers fill the contiguous interval $0,\ldots,N-1$, with a common
encoding width. A prefix block contains a possibly truncated interval of
consecutive integers. A block containing next-trit values $L$ and $S$ also
contains every state in the intervening $R$ subinterval. Each state
contributes all three read-symbol rows.
\end{proof}

\begin{corollary}[Faithful Expert Simulation]\label{universal:cor:simulation}
Each simulated TM transition takes finitely many controller steps and
performs the write and motion prescribed by $\delta$. The expert halts
exactly when the simulated TM issues $S$.
\end{corollary}
\begin{proof}
\Cref{universal:lem:geometry} selects the correct instruction and successor.
Every scan stays within a finite program interval, and every trit search
reaches a nonempty target interval. The final two instruction fields supply
the prescribed write and motion at $Z$. Only a simulated $S$ triggers a
stopping transition.
\end{proof}

%% file: Appendix/Proof_Tables/m_universal.tex
\label{m_universal:sec:common3}
\label{m_universal:sec:protocol}
The protocol records finite geometric context. Let $e\in\{0,2\}$ identify the endpoint row occupied by $X$.
Use $r,j\in\{0,1,2\}$ to index tape symbols and $t,a\in\{L,R,S\}$ to denote trits. A vertical search records the last
observed trit. Each phase and parameter tuple determines an expert state.

\begingroup

\footnotesize\renewcommand{\arraystretch}{1.12}
\begin{longtable}{l c p{6.25cm}}
\toprule
Phase & Expert state & Allowed reading and continuation\\\midrule
\endhead
Initial & $0$ & Read $0$ or $1$; select the matching row from endpoint $0$.\\
Dispatch$(e)$ & $0$ & Read $\sigma_r$; jump to selection of $r$.\\
Selection jump$(e,r)$ & $1+r$ & Read $X$; step left to row index $j=e$.\\
Select$(e,r,j)$ & $1+r$ & Read $\sigma_j$; if $j=r$, cross column $1$; otherwise change $j$ by $\operatorname{sgn}(r-j)$.\\
Cross$(e,r)$ & $4$ & Read $X$ if $e=r$, and $\blank$ otherwise; enter first source trit.\\
First source trit$(e)$ & $4$ & Read any trit; enter remaining source-code scan.\\
Source scan$(e)$ & $4$ & Trit: repeat. Delimiter $0$: enter first successor trit.\\
First successor trit$(e)$ & $5$ & Read a trit $t$; jump to $X$ for matching $t$.\\
Successor scan$(e)$ & $5$ & Trit $t$: match $t$. Delimiter $0$: start rewind.\\
Match jump$(e,t)$ & $g_t$ & Read $X$; step right and begin comparison.\\
First comparison$(e,t)$ & $g_t$ & Read $a$. If $a=t$, resume with endpoint $e$; if $a<t$, search down; if $a>t$, search up.\\
Search down$(t,a)$ & $g_t$ & Read $a$ or the immediate successor of $a$. Repeat until $t$; then resume with endpoint $0$.\\
Search up$(t,a)$ & $g_t$ & Read $a$ or the immediate predecessor of $a$. Repeat until $t$; then resume with endpoint $2$.\\
Resume$(e,t)$ & $g_t$ & Read $Y$; advance to successor scan with endpoint $e$.\\
Rewind jump$(e)$ & $9$ & Read $X$; step left.\\
Rewind$(e)$ & $9$ & Trit: continue left. Blank: move $X$ here and jump to $Y$.\\
Rewind return$(e)$ & $9$ & Read $Y$; step right to the write field.\\
Read write$(e)$ & $10$ & Read $\sigma_r$; remember this write symbol.\\
Read motion$(e,r)$ & $11+r$ & Read $M\in\{L,R,S\}$; jump to $Z$.\\
Apply$(e,r,M)$ & $h_{r,M}$ & Read $Z$; write $\sigma_r$ and move $M$. Dispatch$(e)$ if $M\ne S$; otherwise halt.\\
Halt & $14$ & No outgoing action.\\
\bottomrule
\end{longtable}
\endgroup
Here $g_L=6$, $g_R=7$, $g_S=8$, and
\begin{equation*}
h_{r,L}=1+2r,\qquad h_{r,R}=2+2r,\qquad h_{r,S}=11+r.
\end{equation*}
The protocol starts at Initial; later simulated steps start at Dispatch$(e)$.
Searches update the remembered trit
after each change. Thus a downward search for $S$ that has seen $L$ must see
$R$ before $S$. The protocol permits unbounded repetition while retaining finite annotations.

\begin{lemma}[Protocol Coverage]\label{m_universal:lem:coverage}
Every expert execution on a correct encoding with nonempty input follows
this protocol.
\end{lemma}
\begin{proof}
Initially the tape head reads $0$ or $1$, and $X$ occupies endpoint $0$.
Endpoint selection and the following column-$1$ reading follow from \Cref{universal:lem:geometry}.
Both state codes have positive width. Prefix-block monotonicity and
\Cref{universal:lem:middle} give the allowed vertical-search readings and endpoint
updates. Rewind reaches the blank in column $1$, then $Y$, the write field,
the motion field and $Z$. These exhaust the phases of one simulated step;
induction gives coverage for every finite prefix of the execution.
\end{proof}

The protocol permits additional observation sequences, including unequal
lengths for the two code scans. The certificate checks every permitted
transition, so this overapproximation preserves the equivalence argument.

\label{m_universal:sec:common4}
\paragraph{Initial Entry, Row Selection and Code Scanning}\label{m_universal:sec:part4}
For phase $i$, write $Q_i$ for the controller-state set specified in the
following certificate subsections. Relate configurations with identical grids, heads
and rune records, with expert state $g_i$ and controller state in $Q_i$.

\textbf{Initial and Subsequent Dispatches.}
The Initial phase has $Q_i=\{0\}$ and permits $0,1$ only, with successors
$1,35$, respectively. Dispatch$(e)$ has $Q_i=\{17\}$ for either endpoint
$e$, with successors $18,24,26$ on $\blank,0,1$. Every permitted transition
writes $Z$ and teleports to $X$, matching expert state $0$.
The certificate distinguishes the two initial selection routes.

\textbf{The Complete Row-Selection Routes.}
Each state list gives the controller state \emph{before each reading}; after
the final reading, the controller enters state $5$ at the first source-code
trit. The table represents each endpoint by the corresponding read symbol,
$\blank$ or $1$.

\begin{center}
\small\renewcommand{\arraystretch}{1.12}\setlength{\tabcolsep}{3pt}
\begin{tabular}{l c c l l}
\toprule
Entry & Endpoint & Wanted & Readings in order & States in order\\\midrule
Initial & $\blank$ & $0$ & $X,\blank,0,\blank$ & $1,2,3,4$\\
Initial & $\blank$ & $1$ & $X,\blank,0,1,\blank$ & $35,36,37,38,4$\\\midrule
Later & $\blank$ & $\blank$ & $X,\blank,X$ & $18,19,20$\\
Later & $\blank$ & $0$ & $X,\blank,0,\blank$ & $24,25,3,4$\\
Later & $\blank$ & $1$ & $X,\blank,0,1,\blank$ & $26,27,37,38,4$\\
Later & $1$ & $\blank$ & $X,1,0,\blank,\blank$ & $18,19,32,33,4$\\
Later & $1$ & $0$ & $X,1,0,\blank$ & $24,25,3,4$\\
Later & $1$ & $1$ & $X,1,X$ & $26,27,20$\\
\bottomrule
\end{tabular}
\end{center}
The expert uses state $1+r$ during selection of $\sigma_r$, then state
$4$ for the final column-$1$ reading. The $X$ step moves left; comparisons move up
or down toward the wanted symbol; the matching and crossing steps move
right. Every step preserves the read symbol. Substitution in the transition maps
checks every route and successor state $5$.

Taking the union where Initial and Later routes share a phase gives exactly
the needed multivalued sets. For endpoint $0$ and wanted symbol $0$, the selection-jump and first-comparison sets equal $\{1,24\}$ and
$\{2,25\}$, respectively. For endpoint $0$ and wanted symbol $1$, the same
sets equal $\{26,35\}$ and $\{27,36\}$. Every other phase/parameter set
contains one controller state.

\textbf{Source and Successor Code Scans.}
The first and subsequent source-trit phases both have $Q_i=\{5\}$.
State $5$ preserves each trit, moves right, and remains $5$.
On delimiter $0$, state $5$ preserves $0$, moves right, and enters state $6$.
Both the first successor-trit phase and the subsequent successor scan
have $Q_i=\{6\}$, with the following transitions:
\begin{center}
\small\renewcommand{\arraystretch}{1.12}
\begin{tabular}{c cccc}
\toprule
Controller state & Read $L$ & Read $R$ & Read $S$ & Read $0$\\\midrule
$6$ & $7$ & $10$ & $28$ & $11$\\
\bottomrule
\end{tabular}
\end{center}
Entries list successor states; each action writes $Y$ and teleports to $X$.
The delimiter branch follows at least one successor trit.
Expert state $5$ enters $6,7,8$ for trits $L,R,S$, and $9$ for the delimiter.
This gives the match-jump states $7,10,28$ and rewind-jump state $11$.

\paragraph{Matching Trits, Rewinding and Updating the Tape}\label{m_universal:sec:part5}
\textbf{Match One Successor Trit.}
The match-jump states $7,10,28$ preserve $X$, move right and enter $8,10,29$,
respectively, for targets $L,R,S$. The following table gives the comparison actions.

\begin{center}
\small\renewcommand{\arraystretch}{1.12}
\begin{tabular}{c c ccc}
\toprule
Target & State & Read $L$ & Read $R$ & Read $S$\\\midrule
$L$ & $8$ & $\triple{9}{X}{Y}$ & $\triple{8}{R}{U}$ & $\triple{8}{S}{U}$\\
$R$ & $10$ & $\triple{10}{L}{D}$ & $\triple{9}{X}{Y}$ & $\triple{10}{S}{U}$\\
$S$ & $29$ & $\triple{29}{L}{D}$ & $\triple{29}{R}{D}$ & $\triple{9}{X}{Y}$\\
\bottomrule
\end{tabular}
\end{center}
These rows cover every permitted first-comparison and vertical-search
phase, paired with expert states $6,7,8$.
On a match, writing $X$ advances the bookmark and teleporting to $Y$
returns to the instruction. Resume$(e,t)$ has $Q_i=\{9\}$ for every endpoint
and target: state $9$ preserves $Y$, moves right, and enters state $6$.

\textbf{Rewind $X$ and Reach the Write Field.}
All rewind phases use expert state $9$ and follow this sequence:
\begin{center}
\small\renewcommand{\arraystretch}{1.12}
\begin{tabular}{l c c l}
\toprule
Phase & Controller state & Reading & Transition\\\midrule
Rewind jump & $11$ & $X$ & $\triple{12}{X}{L}$\\
Rewind through code & $12$ & $t\in\{L,R,S\}$ & $\triple{12}{t}{L}$\\
Rewind reaches column $1$ & $12$ & $\blank$ & $\triple{13}{X}{Y}$\\
Rewind return & $13$ & $Y$ & $\triple{14}{Y}{R}$\\
\bottomrule
\end{tabular}
\end{center}
The final step enters expert state $10$, so ReadWrite$(e)$ has $Q_i=\{14\}$.
On reading $\blank,0,1$, state $14$ preserves that symbol, moves right and
enters $15,21,30$, respectively. The corresponding expert transitions enter states $11,12,13$.

\textbf{Motion Decoding and Tape Update.}
Each motion-reading transition writes $Y$, teleports to $Z$, and enters
the apply state below. That state reads $Z$, writes the row's symbol and
performs exactly the chosen motion.
\begin{center}
\small\renewcommand{\arraystretch}{1.12}\setlength{\tabcolsep}{3pt}
\begin{tabular}{c c ccc}
\toprule
Write symbol & Motion-reading state & Apply $L$ & Apply $R$ & Apply $S$\\\midrule
$\blank$ & $15$ & $16$ & $23$ & $34$\\
$0$ & $21$ & $40$ & $22$ & $42$\\
$1$ & $30$ & $39$ & $31$ & $41$\\
\bottomrule
\end{tabular}
\end{center}
The $L$ and $R$ apply transitions return to controller state $17$, paired
with expert state $0$ in Dispatch$(e)$. Either motion may reach a blank tape cell;
state $17$ handles that case correctly. The $S$ transitions enter controller
halt state $43$, paired with expert halt state $14$.

\textbf{Certificate Size.}
These certificate subsections specify every $Q_i$, including the four doubletons caused
by initial versus later selection. The $92$ reachable parameterized phases
produce $96$ related phase--state pairs. The $160$ local checks cover
all $73$ defined nonterminal entries of the extracted controller.
The remaining nine defined entries contain halt loops, excluded from
PTOs. Every permitted successor belongs to the next phase's controller-state set.

For each permitted protocol edge $i\xrightarrow{x}j$ and each $q\in Q_i$,
the preceding certificate establishes
\begin{equation}
\delta_\EXPERT(g_i,x)=(g_j,w,M),\qquad
\delta_\CTRL(q,x)=(q',w,M),\qquad q'\in Q_j. \label{eq:proof:m-universal.2}
\end{equation}
Every required entry exists. A transition reaches a halt state exactly
on an edge to the terminal phase, with motion $S$.

%% file: Appendix/Proof_Tables/a_universal.tex
\label{a_universal:sec:common3}
\label{a_universal:sec:protocol}
The protocol records finite geometric context. Let $e\in\{0,2\}$ identify the endpoint row occupied by $X$.
Use $r,j\in\{0,1,2\}$ to index tape symbols and $t,a\in\{L,R,S\}$ to denote trits. A vertical search records the last
observed trit. Each phase and parameter tuple determines an expert state.

\begingroup

\footnotesize\renewcommand{\arraystretch}{1.12}
\begin{longtable}{l c p{6.25cm}}
\toprule
Phase & Expert state & Allowed reading and continuation\\\midrule
\endhead
Dispatch$(e)$ & $0$ & Read $\sigma_r$; jump to selection of $r$.\\
Selection jump$(e,r)$ & $1+r$ & Read $X$; step left to row index $j=e$.\\
Select$(e,r,j)$ & $1+r$ & Read $\sigma_j$; if $j=r$, cross column $1$; otherwise change $j$ by $\operatorname{sgn}(r-j)$.\\
Cross$(e,r)$ & $4$ & Read $X$ if $e=r$, and $\blank$ otherwise; enter first source trit.\\
First source trit$(e)$ & $4$ & Read any trit; enter remaining source-code scan.\\
Source scan$(e)$ & $4$ & Trit: repeat. Delimiter $0$: enter first successor trit.\\
First successor trit$(e)$ & $5$ & Read a trit $t$; jump to $X$ for matching $t$.\\
Successor scan$(e)$ & $5$ & Trit $t$: match $t$. Delimiter $0$: start rewind.\\
Match jump$(e,t)$ & $g_t$ & Read $X$; step right and begin comparison.\\
First comparison$(e,t)$ & $g_t$ & Read $a$. If $a=t$, resume with endpoint $e$; if $a<t$, search down; if $a>t$, search up.\\
Search down$(t,a)$ & $g_t$ & Read $a$ or the immediate successor of $a$. Repeat until $t$; then resume with endpoint $0$.\\
Search up$(t,a)$ & $g_t$ & Read $a$ or the immediate predecessor of $a$. Repeat until $t$; then resume with endpoint $2$.\\
Resume$(e,t)$ & $g_t$ & Read $Y$; advance to successor scan with endpoint $e$.\\
Rewind jump$(e)$ & $9$ & Read $X$; step left.\\
Rewind$(e)$ & $9$ & Trit: continue left. Blank: move $X$ here and jump to $Y$.\\
Rewind return$(e)$ & $9$ & Read $Y$; step right to the write field.\\
Read write$(e)$ & $10$ & Read $\sigma_r$; remember this write symbol.\\
Read motion$(e,r)$ & $11+r$ & Read $M\in\{L,R,S\}$; jump to $Z$.\\
Apply$(e,r,M)$ & $h_{r,M}$ & Read $Z$; write $\sigma_r$ and move $M$. Dispatch$(e)$ if $M\ne S$; otherwise halt.\\
Halt & $14$ & No outgoing action.\\
\bottomrule
\end{longtable}
\endgroup
Here $g_L=6$, $g_R=7$, $g_S=8$, and
\begin{equation*}
h_{r,L}=1+2r,\qquad h_{r,R}=2+2r,\qquad h_{r,S}=11+r.
\end{equation*}
The protocol starts at Dispatch$(0)$. Searches update the remembered trit
after each change. Thus a downward search for $S$ that has seen $L$ must see
$R$ before $S$. The protocol permits unbounded repetition while retaining finite annotations.

\begin{lemma}[Protocol Coverage]\label{a_universal:lem:coverage}
Every expert execution on a correct encoding follows this protocol.
\end{lemma}
\begin{proof}
Endpoint selection and the following column-$1$ reading follow from \Cref{universal:lem:geometry}.
Both state codes have positive width. Prefix-block monotonicity and
\Cref{universal:lem:middle} give the allowed vertical-search readings and endpoint
updates. Rewind reaches the blank in column $1$, then $Y$, the write field,
the motion field and $Z$. These exhaust the phases of one simulated step;
induction gives coverage for every finite prefix of the execution.
\end{proof}

The protocol permits additional observation sequences, including unequal
lengths for the two code scans. The certificate checks every permitted
transition, so this overapproximation preserves the equivalence argument.

\label{a_universal:sec:common4}
\paragraph{Dispatch, Row Selection and Code Scanning}\label{a_universal:sec:part4}
For phase $i$, write $Q_i$ for the controller-state set specified in the
following certificate subsections. The simulation relation pairs expert state $g_i$ with controller state
$q\in Q_i$, with identical grids, heads, and rune records.

\textbf{Dispatch.}
For either endpoint $e$, phase Dispatch$(e)$ has $Q_i=\{0\}$.
On reading $\blank,0,1$, state $0$ writes $Z$, jumps to $X$, and enters
controller state $23,0,22$, respectively. These actions match expert state $0$.

\textbf{All Six Row-Selection Cases.}
The following lists the readings from arrival at $X$ through crossing
column $1$. The controller-state list gives the state \emph{before each
reading}. Each route ends in controller state $10$ at the first source-code
trit. The table represents endpoint $e$ by the corresponding read symbol, $\blank$ or $1$.

\begin{center}
\small\renewcommand{\arraystretch}{1.12}
\begin{tabular}{c c l l}
\toprule
Endpoint & Wanted & Readings in order & Controller states in order\\\midrule
$\blank$ & $\blank$ & $X,\blank,X$ & $23,10,19$\\
$\blank$ & $0$ & $X,\blank,0,\blank$ & $0,20,7,25$\\
$\blank$ & $1$ & $X,\blank,0,1,\blank$ & $22,7,19,25,25$\\
$1$ & $\blank$ & $X,1,0,\blank,\blank$ & $23,10,10,9,22$\\
$1$ & $0$ & $X,1,0,\blank$ & $0,20,4,25$\\
$1$ & $1$ & $X,1,X$ & $22,7,19$\\
\bottomrule
\end{tabular}
\end{center}
Each route uses expert state $1+r$ during selection and state $4$ for
the final column-$1$ reading. The $X$ step moves left; comparisons
move up or down toward the wanted symbol; the match and crossing steps
move right. Every step preserves the read symbol. Substitution in the transition map
verifies each route, including successor state $10$.

\textbf{Scan the Source-State Code.}
The first trit uses controller state $10$; subsequent trits use states
from $\{4,18\}$. All of the following writes preserve the read symbol and move right:
\begin{center}
\small\renewcommand{\arraystretch}{1.12}
\begin{tabular}{c cccc}
\toprule
Controller state & Read $L$ & Read $R$ & Read $S$ & Read delimiter $0$\\\midrule
$10$ & $18$ & $4$ & $4$ & excluded at first trit\\
$4$ & $18$ & $4$ & $4$ & $25$\\
$18$ & $18$ & $4$ & $18$ & $25$\\
\bottomrule
\end{tabular}
\end{center}
Entries list successor states. Expert state $4$ remains $4$ on trits and
enters $5$ on the delimiter. Thus $Q_i=\{10\}$ at the first source trit, $Q_i=\{4,18\}$ during the
remaining source scan, and $Q_i=\{25\}$ during the successor scan, including
the first successor trit.

\textbf{Successor-Code Dispatch.}
Controller state $25$ has the four relevant transitions
\begin{equation*}
\begin{array}{c|cccc}
\text{read}&L&R&S&0\\\hline
\text{successor}&7&26&15&9
\end{array}
\end{equation*}
and in every case writes $Y$ and jumps to $X$. The corresponding expert transitions enter states $6,7,8,9$. The delimiter case occurs only after at least one
successor trit. The match jumps therefore use states $7,26,15$, and the rewind jump
uses state $9$, independently of the endpoint.

\paragraph{Trit Matching, Rewind and Tape Updates}\label{a_universal:sec:part5}
\textbf{Match One Successor Trit.}
The match-jump states $7,26,15$ preserve $X$ and move right, entering
$19,17,16$ respectively. The following table gives all comparison cases.
Each entry lists $(\text{successor},\text{written symbol},\text{motion})$.

\begin{center}
\small\renewcommand{\arraystretch}{1.12}
\begin{tabular}{l c ccc}
\toprule
Context & State & Read $L$ & Read $R$ & Read $S$\\\midrule
Target $L$ & $19$ & $\triple{18}{X}{Y}$ & $\triple{19}{R}{U}$ & $\triple{19}{S}{U}$\\
Target $R$ & $17$ & $\triple{17}{L}{D}$ & $\triple{2}{X}{Y}$ & $\triple{17}{S}{U}$\\
Target $S$: first comparison & $16$ & $\triple{3}{L}{D}$ & $\triple{5}{R}{D}$ & $\triple{3}{X}{Y}$\\
Target $S$: previous trit $L$ & $3$ & $\triple{3}{L}{D}$ & $\triple{5}{R}{D}$ & excluded\\
Target $S$: previous trit $R$ & $5$ & excluded & $\triple{5}{R}{D}$ & $\triple{3}{X}{Y}$\\
\bottomrule
\end{tabular}
\end{center}
The corresponding expert states have indices $6,7,8$. The exclusions follow from the
protocol, especially \Cref{universal:lem:middle}. After equality, states $18,2,3$
respectively read and preserve $Y$, then move right to state $25$.
These states represent Resume$(e,L)$, Resume$(e,R)$, and Resume$(e,S)$.
An immediate match retains the endpoint; a downward search selects endpoint
$0$, and an upward search selects endpoint $2$.

\textbf{Rewind $X$ and Reach the Write Field.}
Expert state $9$ corresponds to the following exact sequence:
\begin{center}
\small\renewcommand{\arraystretch}{1.12}
\begin{tabular}{l c c l}
\toprule
Phase & Controller state & Reading & Transition\\\midrule
Rewind jump & $9$ & $X$ & $\triple{14}{X}{L}$\\
Rewind through code & $14$ & $t\in\{L,R,S\}$ & $\triple{14}{t}{L}$\\
Rewind reaches column $1$ & $14$ & $\blank$ & $\triple{19}{X}{Y}$\\
Rewind return & $19$ & $Y$ & $\triple{8}{Y}{R}$\\
\bottomrule
\end{tabular}
\end{center}
The last step enters expert state $10$, so $Q_i=\{8\}$ in phase ReadWrite$(e)$.
On reading $\blank,0,1$, state $8$ preserves the symbol, moves right and
enters $21,9,8$ respectively; the expert transitions enter states $11,12,13$.

\textbf{Motion Decoding and Tape Update.}
For each row below, reading the selected motion writes $Y$, teleports to
$Z$, and enters the listed controller state. That state then reads $Z$,
writes the indicated symbol and performs exactly the selected move.
\begin{center}
\small\renewcommand{\arraystretch}{1.12}
\begin{tabular}{c c ccc}
\toprule
Write symbol & Motion-reading state & Apply $L$ & Apply $R$ & Apply $S$\\\midrule
$\blank$ & $21$ & $7$ & $4$ & $14$\\
$0$ & $9$ & $3$ & $26$ & $18$\\
$1$ & $8$ & $11$ & $1$ & $15$\\
\bottomrule
\end{tabular}
\end{center}
The $L$ and $R$ cases return to controller state $0$, matching expert state
$0$. The $S$ cases enter controller halt state $27$, matching expert halt
state $14$. These establish all Apply phases and the terminal pair.

Together, the certificate subsections specify $Q_i$ for every reachable
protocol phase and verify every permitted transition. Only source-code scanning requires multivalued sets: $\{4,18\}$ for each endpoint.
All other phase--parameter combinations have a unique controller state.

The certificate contains $91$ reachable parameterized phases and $93$ related
phase--controller pairs, including the terminal pair. For every allowed protocol edge $i\xrightarrow{x}j$ and every controller state $q\in Q_i$, the preceding certificate gives
\begin{equation}
\delta_\EXPERT(g_i,x)=(g_j,w,M),\qquad
\delta_\CTRL(q,x)=(q',w,M),\qquad q'\in Q_j. \label{eq:proof:a-universal.2}
\end{equation}
The $162$ local checks cover $81$ distinct nonterminal entries of the controller table.
A transition reaches a halt state exactly on an edge to the terminal phase.

%% file: Appendix/Spurious_Controllers.tex
\section{Short-Input Agreement Does Not Identify an Algorithm}\label{app:spurious}

Small training datasets do not force the expert arithmetic rule, even within compact controller classes. We construct two counterexamples using the same alphabets, grid, initial placement, and action vocabulary as the binary experts. Both controllers reproduce every expert PTO within the training-length bounds, including all write and motion decisions. Both nevertheless stop prematurely on longer inputs. Thus the ambiguity persists even with exhaustive supervision over the short-input domain.

Throughout this section, $a,b\in\bits^+$ denote nonempty binary operand words, with leading zeros permitted. The input $a+b$ or $a\mathtt{x}b$ occupies row zero from column zero; the head starts at the first left-operand digit and every other cell starts blank. Each table entry gives the next state, written symbol, and motion. Motion $S$ writes once and stops; the stopping-state loops specify a total transition map but add no further PTO steps. All state counts include the stopping state.

\input{Appendix/Spurious_Tables/add2}
\input{Appendix/Spurious_Tables/mult2}

\subsection{Capacity Does Not Exclude the Counterexamples}\label{app:spurious-capacity}

The constructions add only three states to the eleven-state addition expert and six states to the thirty-two-state multiplication expert. Every training subset within the stated length bounds supplies identical PTO targets for the expert and the corresponding spurious controller.

\Cref{thm-universal} represents the two controllers using ordinary DFST dimensions 14 and 38. For Softmax DFSTs with finite logits in exact real arithmetic, dimensions 15 and 39 suffice for exact argmax behavior. These bounds provide sufficient capacities, without asserting minimality. Padding the discrete controllers with unreachable states extends the construction to every larger dimension. In particular, the addition sweeps at dimensions 22 and 33 and the multiplication sweeps at dimensions 64 and 96 can express the counterexamples.

Action agreement and fitting the one-hot training targets require separate statements. The exact-real softmax construction in \Cref{app:dfsts} guarantees the correct argmax actions, but need not produce one-hot readouts. For any PTO dataset within the stated training-length bounds, softmax logits approaching the deterministic vertices instead drive the MSE arbitrarily close to zero: finite-horizon recurrence and readout maps depend continuously on the effective parameters. Under rounded arithmetic that saturates softmax at vertices and preserves vertex operations, the one-hot realization gives exactly zero training MSE. One explicit choice assigns logit $0$ to each selected transition, output, and initial-state coordinate, and $-128$ to every other coordinate under correctly rounded binary32 exponentiation.

\begin{table}[htbp]
\centering
\caption{\textbf{Generalization despite Capacity for Short-Input Counterexamples.} The wider Softmax DFST sweeps each contain six configurations. Five reach the robust testing cap in each task.}
\label{tab:spurious-capacity}
\small
\begin{tabular}{lrrrr}
\toprule
Task & Expert states & Spurious states & Tested dimensions & Cap-reaching runs \\
\midrule
\texttt{add2} & 11 & 14 & $22,33$ & $5/6$ \\
\texttt{mult2} & 32 & 38 & $64,96$ & $5/6$ \\
\bottomrule
\end{tabular}
\end{table}

The wider sweeps exclude insufficient capacity as the explanation for avoiding these two spurious rules (\Cref{tab:spurious-capacity,fig:softmax-arithmetic}). The experiments support a useful implicit bias of parametrization, initialization, and gradient-based optimization toward length-independent behavior. However, no claim identifies the failed runs with the constructed counterexamples, and the results do not establish that gradient descent always selects the simplest controller. Extracted state counts alone need not capture the relevant notion of simplicity.

\subsection{A Simple Model for Studying Algorithmic Learning}

The ordinary DFST uses three input-indexed tensors and an initial-state vector. Each recurrence and readout consists of a matrix-vector product, without nonlinear activations. This stepwise multilinearity offers a transparent starting point for analyzing which solutions gradient-based optimization favors. Repeated use of shared transition parameters across a trajectory still produces a nonconvex learning problem; neither the complete trajectory map nor the MSE objective generally remains multilinear in the shared parameters. The softmax parametrization adds another nonlinearity through the logits.

%% file: Appendix/Spurious_Tables/add2.tex
\subsection{A Fourteen-State Spurious Addition Controller}\label{app:spurious-add2}
The binary-addition expert uses eleven states, including the stopping state.
Replace the initial scanning state by a counter with states $0,11,12,13$,
representing zero, one, two, and three scanned left-operand digits.
The resulting controller contains fourteen states.

\begin{proposition}[Short-Input Agreement without Exact Addition]\label{prop:spurious-add2}
The controller in \Cref{tab:spurious-add2} produces the expert PTO on every
input $a+b$ with $1\leq |a|\leq3$ and $b\in\bits^+$, including leading zeros.
For every $|a|\geq4$ and $b\in\bits^+$, execution stops on the fourth left-operand
digit and leaves the input unchanged.
\end{proposition}
\begin{proof}
Before reaching the separator, the expert remains in state $0$ and writes each
read digit unchanged while moving right. The counter states produce exactly
these writes and motions for the first three digits. On reading $+$, every
counter state enters expert state $1$ with the same write and left move.
No expert transition from states $1,\ldots,10$ returns to state $0$.
Consequently, the controllers have identical states, grids, and heads after
the separator transition and follow the same remaining execution, including
the stopping action. If a fourth left-operand digit occurs, state $13$ instead
writes the digit unchanged and stops, whereas the expert moves right.
The unchanged input still contains $+$ and therefore fails the output contract.
\end{proof}

For example, $1111+1$ triggers the discrepancy at step four:
the expert emits $(1,\mathsf R)$, while the spurious controller emits
$(1,\mathsf S)$. The expert eventually leaves $10000$ on the grid.
Thus even all $14^2=196$ operand pairs with lengths at most three cannot
distinguish these controllers through PTO supervision.

\begin{table}[htbp]
\centering
\caption{\textbf{Complete Spurious-Addition Transition Map.}
Initial state $0$; stopping state $10$. Each entry gives
$(\text{next state},\text{write},\text{motion})$ for the column's read symbol.
Motion $S$ writes once and terminates execution.}
\label{tab:spurious-add2}
\begingroup
\small\renewcommand{\arraystretch}{1.12}
\setlength{\tabcolsep}{5pt}
\begin{tabular}{rcccc}
\toprule
State & $\lambda$ & $+$ & $0$ & $1$ \\
\midrule
0 & $(10,\lambda,\mathsf{S})$ & $(1,+,\mathsf{L})$ & $(11,0,\mathsf{R})$ & $(11,1,\mathsf{R})$ \\
1 & $(8,+,\mathsf{U})$ & $(10,+,\mathsf{S})$ & $(1,0,\mathsf{D})$ & $(1,1,\mathsf{D})$ \\
2 & $(3,\lambda,\mathsf{L})$ & $(2,+,\mathsf{R})$ & $(2,0,\mathsf{R})$ & $(2,1,\mathsf{R})$ \\
3 & $(10,\lambda,\mathsf{S})$ & $(9,\lambda,\mathsf{D})$ & $(4,\lambda,\mathsf{D})$ & $(5,\lambda,\mathsf{D})$ \\
4 & $(4,\lambda,\mathsf{L})$ & $(4,+,\mathsf{U})$ & $(7,0,\mathsf{D})$ & $(7,1,\mathsf{D})$ \\
5 & $(5,\lambda,\mathsf{L})$ & $(5,+,\mathsf{U})$ & $(7,1,\mathsf{D})$ & $(6,0,\mathsf{L})$ \\
6 & $(7,1,\mathsf{D})$ & $(10,+,\mathsf{S})$ & $(7,1,\mathsf{D})$ & $(6,0,\mathsf{L})$ \\
7 & $(7,\lambda,\mathsf{R})$ & $(1,\lambda,\mathsf{L})$ & $(10,0,\mathsf{S})$ & $(10,1,\mathsf{S})$ \\
8 & $(2,0,\mathsf{R})$ & $(10,+,\mathsf{S})$ & $(2,0,\mathsf{R})$ & $(2,1,\mathsf{R})$ \\
9 & $(9,\lambda,\mathsf{L})$ & $(9,\lambda,\mathsf{U})$ & $(10,0,\mathsf{S})$ & $(10,1,\mathsf{S})$ \\
10 & $(10,\lambda,\mathsf{S})$ & $(10,+,\mathsf{S})$ & $(10,0,\mathsf{S})$ & $(10,1,\mathsf{S})$ \\
11 & $(10,\lambda,\mathsf{S})$ & $(1,+,\mathsf{L})$ & $(12,0,\mathsf{R})$ & $(12,1,\mathsf{R})$ \\
12 & $(10,\lambda,\mathsf{S})$ & $(1,+,\mathsf{L})$ & $(13,0,\mathsf{R})$ & $(13,1,\mathsf{R})$ \\
13 & $(10,\lambda,\mathsf{S})$ & $(1,+,\mathsf{L})$ & $(10,0,\mathsf{S})$ & $(10,1,\mathsf{S})$ \\
\bottomrule
\end{tabular}
\endgroup
\end{table}

%% file: Appendix/Spurious_Tables/mult2.tex
\subsection{A Thirty-Eight-State Spurious Multiplication Controller}\label{app:spurious-mult2}
Let $T_\times$ denote the 32-state binary-multiplication expert. Define $\widetilde T_\times$ by retaining all expert transitions and adding six states $32,\ldots,37$ for the initial scan of the left operand. State $32+j$ records $j$ digits already scanned. The initial state becomes $32$; the halt state remains $31$. The added states preserve each scanned symbol and move right, except that reading a sixth digit triggers an immediate stop. Reading the separator $\mathtt{x}$ enters expert state $0$ with the expert's write and motion.

\begin{proposition}\label{prop:spurious-mult2}
For every $a\in\bits^m$ with $1\leq m\leq5$ and every nonempty binary word $b$, the complete PTOs of $\widetilde T_\times$ and $T_\times$ on $a\mathtt{x}b$ agree. For every left operand of at least six digits and every nonempty right operand, the spurious controller stops after six actions and leaves the input expression unchanged.
\end{proposition}
\begin{proof}
For the first $m\leq5$ digit readings, both controllers preserve the read symbol and move right. The expert remains in state $0$, while the spurious controller reaches state $32+m$. On the separator, both controllers write $\mathtt{x}$, move right, and enter state $0$. The grid, head, and control state then coincide, and all remaining transitions agree. For $m\geq6$, the first five actions agree, but state $37$ stops on the sixth digit while the expert moves right. Every preceding action preserves the input, so the final grid still contains the separator and both operands, violating the arithmetic output specification.
\end{proof}

For example, $111111\mathtt{x}1$ requires the product $111111$ (decimal $63$), allowing leading-zero padding. The spurious controller instead leaves $111111\mathtt{x}1$ on the grid and halts at the sixth digit. Exhaustive enumeration of all $62^2=3844$ pairs with operand lengths from one through five confirms complete PTO agreement; the argument above additionally covers arbitrary right-operand lengths.

\begingroup
\footnotesize
\renewcommand{\arraystretch}{1.12}
\setlength{\tabcolsep}{5pt}
\begin{longtable}{rcccc}
\caption{Complete Transition Map of the 38-State Spurious Multiplication Controller. Each entry gives $(\text{next state},\text{written symbol},\text{motion})$. Initial state $32$; halt state $31$. Motion $S$ writes once and stops.}\label{tab:spurious-mult2}\\
\toprule
State & $\lambda$ & $0$ & $1$ & $\mathtt{x}$\\
\midrule
\endfirsthead
\toprule
State & $\lambda$ & $0$ & $1$ & $\mathtt{x}$\\
\midrule
\endhead
\midrule
\multicolumn{5}{r}{Continued on the next page.}\\
\endfoot
\bottomrule
\endlastfoot
0 & $(1,\lambda,\mathsf{L})$ & $(0,0,\mathsf{R})$ & $(0,1,\mathsf{R})$ & $(0,\mathtt{x},\mathsf{R})$ \\
1 & $(31,\lambda,\mathsf{S})$ & $(2,0,\mathsf{L})$ & $(2,1,\mathsf{L})$ & $(31,\mathtt{x},\mathsf{S})$ \\
2 & $(31,\lambda,\mathsf{S})$ & $(3,0,\mathsf{D})$ & $(4,1,\mathsf{D})$ & $(10,\mathtt{x},\mathsf{R})$ \\
3 & $(6,\lambda,\mathsf{L})$ & $(6,0,\mathsf{L})$ & $(6,1,\mathsf{L})$ & $(6,\mathtt{x},\mathsf{L})$ \\
4 & $(7,\lambda,\mathsf{L})$ & $(7,0,\mathsf{L})$ & $(7,1,\mathsf{L})$ & $(7,\mathtt{x},\mathsf{L})$ \\
5 & $(31,\lambda,\mathsf{S})$ & $(8,0,\mathsf{L})$ & $(8,1,\mathsf{L})$ & $(31,\mathtt{x},\mathsf{S})$ \\
6 & $(9,0,\mathsf{U})$ & $(31,0,\mathsf{S})$ & $(31,1,\mathsf{S})$ & $(31,\mathtt{x},\mathsf{S})$ \\
7 & $(9,1,\mathsf{U})$ & $(31,0,\mathsf{S})$ & $(31,1,\mathsf{S})$ & $(31,\mathtt{x},\mathsf{S})$ \\
8 & $(9,\mathtt{x},\mathsf{U})$ & $(31,0,\mathsf{S})$ & $(31,1,\mathsf{S})$ & $(31,\mathtt{x},\mathsf{S})$ \\
9 & $(0,\lambda,\mathsf{D})$ & $(3,0,\mathsf{D})$ & $(4,1,\mathsf{D})$ & $(5,\mathtt{x},\mathsf{D})$ \\
10 & $(31,\lambda,\mathsf{S})$ & $(11,\lambda,\mathsf{L})$ & $(12,\lambda,\mathsf{L})$ & $(31,\mathtt{x},\mathsf{S})$ \\
11 & $(31,\lambda,\mathsf{S})$ & $(11,\lambda,\mathsf{L})$ & $(11,\lambda,\mathsf{L})$ & $(13,\lambda,\mathsf{L})$ \\
12 & $(31,\lambda,\mathsf{S})$ & $(12,\lambda,\mathsf{L})$ & $(12,\lambda,\mathsf{L})$ & $(14,\lambda,\mathsf{L})$ \\
13 & $(15,0,\mathsf{R})$ & $(13,0,\mathsf{L})$ & $(13,0,\mathsf{L})$ & $(31,\mathtt{x},\mathsf{S})$ \\
14 & $(15,0,\mathsf{R})$ & $(14,0,\mathsf{L})$ & $(14,1,\mathsf{L})$ & $(31,\mathtt{x},\mathsf{S})$ \\
15 & $(16,\lambda,\mathsf{U})$ & $(15,0,\mathsf{R})$ & $(15,1,\mathsf{R})$ & $(31,\mathtt{x},\mathsf{S})$ \\
16 & $(18,\lambda,\mathsf{D})$ & $(17,0,\mathsf{R})$ & $(17,1,\mathsf{R})$ & $(31,\mathtt{x},\mathsf{S})$ \\
17 & $(10,\lambda,\mathsf{L})$ & $(17,0,\mathsf{R})$ & $(17,1,\mathsf{R})$ & $(17,\mathtt{x},\mathsf{R})$ \\
18 & $(19,\lambda,\mathsf{L})$ & $(31,0,\mathsf{S})$ & $(31,1,\mathsf{S})$ & $(31,\mathtt{x},\mathsf{S})$ \\
19 & $(20,\lambda,\mathsf{D})$ & $(19,0,\mathsf{L})$ & $(19,1,\mathsf{L})$ & $(31,\mathtt{x},\mathsf{S})$ \\
20 & $(30,\lambda,\mathsf{U})$ & $(21,0,\mathsf{R})$ & $(21,1,\mathsf{R})$ & $(31,\mathtt{x},\mathsf{S})$ \\
21 & $(28,\lambda,\mathsf{U})$ & $(21,0,\mathsf{R})$ & $(21,1,\mathsf{R})$ & $(31,\mathtt{x},\mathsf{S})$ \\
22 & $(19,\lambda,\mathsf{D})$ & $(24,\lambda,\mathsf{D})$ & $(25,\lambda,\mathsf{D})$ & $(31,\mathtt{x},\mathsf{S})$ \\
23 & $(22,\lambda,\mathsf{L})$ & $(31,0,\mathsf{S})$ & $(31,1,\mathsf{S})$ & $(31,\mathtt{x},\mathsf{S})$ \\
24 & $(23,0,\mathsf{U})$ & $(23,0,\mathsf{U})$ & $(23,1,\mathsf{U})$ & $(31,\mathtt{x},\mathsf{S})$ \\
25 & $(23,1,\mathsf{U})$ & $(23,1,\mathsf{U})$ & $(27,0,\mathsf{U})$ & $(31,\mathtt{x},\mathsf{S})$ \\
26 & $(25,\lambda,\mathsf{D})$ & $(25,\lambda,\mathsf{D})$ & $(29,\lambda,\mathsf{D})$ & $(31,\mathtt{x},\mathsf{S})$ \\
27 & $(26,\lambda,\mathsf{L})$ & $(31,0,\mathsf{S})$ & $(31,1,\mathsf{S})$ & $(31,\mathtt{x},\mathsf{S})$ \\
28 & $(22,\lambda,\mathsf{L})$ & $(28,0,\mathsf{R})$ & $(28,1,\mathsf{R})$ & $(31,\mathtt{x},\mathsf{S})$ \\
29 & $(27,0,\mathsf{U})$ & $(27,0,\mathsf{U})$ & $(27,1,\mathsf{U})$ & $(31,\mathtt{x},\mathsf{S})$ \\
30 & $(31,\lambda,\mathsf{S})$ & $(31,0,\mathsf{S})$ & $(31,1,\mathsf{S})$ & $(31,\mathtt{x},\mathsf{S})$ \\
31 & $(31,\lambda,\mathsf{S})$ & $(31,0,\mathsf{S})$ & $(31,1,\mathsf{S})$ & $(31,\mathtt{x},\mathsf{S})$ \\
\midrule
32 & $(1,\lambda,\mathsf{L})$ & $(33,0,\mathsf{R})$ & $(33,1,\mathsf{R})$ & $(0,\mathtt{x},\mathsf{R})$ \\
33 & $(1,\lambda,\mathsf{L})$ & $(34,0,\mathsf{R})$ & $(34,1,\mathsf{R})$ & $(0,\mathtt{x},\mathsf{R})$ \\
34 & $(1,\lambda,\mathsf{L})$ & $(35,0,\mathsf{R})$ & $(35,1,\mathsf{R})$ & $(0,\mathtt{x},\mathsf{R})$ \\
35 & $(1,\lambda,\mathsf{L})$ & $(36,0,\mathsf{R})$ & $(36,1,\mathsf{R})$ & $(0,\mathtt{x},\mathsf{R})$ \\
36 & $(1,\lambda,\mathsf{L})$ & $(37,0,\mathsf{R})$ & $(37,1,\mathsf{R})$ & $(0,\mathtt{x},\mathsf{R})$ \\
37 & $(1,\lambda,\mathsf{L})$ & $(31,0,\mathsf{S})$ & $(31,1,\mathsf{S})$ & $(0,\mathtt{x},\mathsf{R})$ \\
\end{longtable}
\endgroup

%% file: checklist.tex
\section*{NeurIPS Paper Checklist}

\begin{enumerate}

\item {\bf Claims}
    \item[] Question: Do the main claims made in the abstract and introduction accurately reflect the paper's contributions and scope?
    \item[] Answer: \answerYes{} 
    \item[] Justification: In the abstract and introduction, we claim that we introduce a simple learning framework capable of achieving extreme length generalization and exact learning on arithmetic and universal simulation tasks. We substantiate these claims with experimental results and mathematical proofs.
    \item[] Guidelines:
    \begin{itemize}
        \item The answer \answerNA{} means that the abstract and introduction do not include the claims made in the paper.
        \item The abstract and/or introduction should clearly state the claims made, including the contributions made in the paper and important assumptions and limitations. A \answerNo{} or \answerNA{} answer to this question will not be perceived well by the reviewers. 
        \item The claims made should match theoretical and experimental results, and reflect how much the results can be expected to generalize to other settings. 
        \item It is fine to include aspirational goals as motivation as long as it is clear that these goals are not attained by the paper. 
    \end{itemize}

\item {\bf Limitations}
    \item[] Question: Does the paper discuss the limitations of the work performed by the authors?
    \item[] Answer: \answerYes{} 
    \item[] Justification: We provide evidence of hyperparameter configurations where our differentiable controllers cannot achieve extreme length generalization and provably do not exactly learn the ground-truth procedure.
    \item[] Guidelines:
    \begin{itemize}
        \item The answer \answerNA{} means that the paper has no limitation while the answer \answerNo{} means that the paper has limitations, but those are not discussed in the paper. 
        \item The authors are encouraged to create a separate ``Limitations'' section in their paper.
        \item The paper should point out any strong assumptions and how robust the results are to violations of these assumptions (e.g., independence assumptions, noiseless settings, model well-specification, asymptotic approximations only holding locally). The authors should reflect on how these assumptions might be violated in practice and what the implications would be.
        \item The authors should reflect on the scope of the claims made, e.g., if the approach was only tested on a few datasets or with a few runs. In general, empirical results often depend on implicit assumptions, which should be articulated.
        \item The authors should reflect on the factors that influence the performance of the approach. For example, a facial recognition algorithm may perform poorly when image resolution is low or images are taken in low lighting. Or a speech-to-text system might not be used reliably to provide closed captions for online lectures because it fails to handle technical jargon.
        \item The authors should discuss the computational efficiency of the proposed algorithms and how they scale with dataset size.
        \item If applicable, the authors should discuss possible limitations of their approach to address problems of privacy and fairness.
        \item While the authors might fear that complete honesty about limitations might be used by reviewers as grounds for rejection, a worse outcome might be that reviewers discover limitations that aren't acknowledged in the paper. The authors should use their best judgment and recognize that individual actions in favor of transparency play an important role in developing norms that preserve the integrity of the community. Reviewers will be specifically instructed to not penalize honesty concerning limitations.
    \end{itemize}

\item {\bf Theory assumptions and proofs}
    \item[] Question: For each theoretical result, does the paper provide the full set of assumptions and a complete (and correct) proof?
    \item[] Answer: \answerYes{} 
    \item[] Justification: We provide the full set of empirically falsifiable assumptions for our theorems. All theorems are carefully proved in the appendix sections.
    \item[] Guidelines:
    \begin{itemize}
        \item The answer \answerNA{} means that the paper does not include theoretical results. 
        \item All the theorems, formulas, and proofs in the paper should be numbered and cross-referenced.
        \item All assumptions should be clearly stated or referenced in the statement of any theorems.
        \item The proofs can either appear in the main paper or the supplemental material, but if they appear in the supplemental material, the authors are encouraged to provide a short proof sketch to provide intuition. 
        \item Inversely, any informal proof provided in the core of the paper should be complemented by formal proofs provided in appendix or supplemental material.
        \item Theorems and Lemmas that the proof relies upon should be properly referenced. 
    \end{itemize}

    \item {\bf Experimental result reproducibility}
    \item[] Question: Does the paper fully disclose all the information needed to reproduce the main experimental results of the paper to the extent that it affects the main claims and/or conclusions of the paper (regardless of whether the code and data are provided or not)?
    \item[] Answer: \answerYes{} 
    \item[] Justification: In our appendix section about implementation details, we provide full details on our training procedure. Furthermore, we make our code publicly available (alongside the weights of the runs discussed in the paper) so that other researchers can experiment with and visualize learning from Turing-machine computation.
    \item[] Guidelines:
    \begin{itemize}
        \item The answer \answerNA{} means that the paper does not include experiments.
        \item If the paper includes experiments, a \answerNo{} answer to this question will not be perceived well by the reviewers: Making the paper reproducible is important, regardless of whether the code and data are provided or not.
        \item If the contribution is a dataset and\slash or model, the authors should describe the steps taken to make their results reproducible or verifiable. 
        \item Depending on the contribution, reproducibility can be accomplished in various ways. For example, if the contribution is a novel architecture, describing the architecture fully might suffice, or if the contribution is a specific model and empirical evaluation, it may be necessary to either make it possible for others to replicate the model with the same dataset, or provide access to the model. In general. releasing code and data is often one good way to accomplish this, but reproducibility can also be provided via detailed instructions for how to replicate the results, access to a hosted model (e.g., in the case of a large language model), releasing of a model checkpoint, or other means that are appropriate to the research performed.
        \item While NeurIPS does not require releasing code, the conference does require all submissions to provide some reasonable avenue for reproducibility, which may depend on the nature of the contribution. For example
        \begin{enumerate}
            \item If the contribution is primarily a new algorithm, the paper should make it clear how to reproduce that algorithm.
            \item If the contribution is primarily a new model architecture, the paper should describe the architecture clearly and fully.
            \item If the contribution is a new model (e.g., a large language model), then there should either be a way to access this model for reproducing the results or a way to reproduce the model (e.g., with an open-source dataset or instructions for how to construct the dataset).
            \item We recognize that reproducibility may be tricky in some cases, in which case authors are welcome to describe the particular way they provide for reproducibility. In the case of closed-source models, it may be that access to the model is limited in some way (e.g., to registered users), but it should be possible for other researchers to have some path to reproducing or verifying the results.
        \end{enumerate}
    \end{itemize}

\item {\bf Open access to data and code}
    \item[] Question: Does the paper provide open access to the data and code, with sufficient instructions to faithfully reproduce the main experimental results, as described in supplemental material?
    \item[] Answer: \answerYes{} 
    \item[] Justification: We provide a well documented codebase with sufficient instructions.
    \item[] Guidelines:
    \begin{itemize}
        \item The answer \answerNA{} means that paper does not include experiments requiring code.
        \item Please see the NeurIPS code and data submission guidelines (\url{https://neurips.cc/public/guides/CodeSubmissionPolicy}) for more details.
        \item While we encourage the release of code and data, we understand that this might not be possible, so \answerNo{} is an acceptable answer. Papers cannot be rejected simply for not including code, unless this is central to the contribution (e.g., for a new open-source benchmark).
        \item The instructions should contain the exact command and environment needed to run to reproduce the results. See the NeurIPS code and data submission guidelines (\url{https://neurips.cc/public/guides/CodeSubmissionPolicy}) for more details.
        \item The authors should provide instructions on data access and preparation, including how to access the raw data, preprocessed data, intermediate data, and generated data, etc.
        \item The authors should provide scripts to reproduce all experimental results for the new proposed method and baselines. If only a subset of experiments are reproducible, they should state which ones are omitted from the script and why.
        \item At submission time, to preserve anonymity, the authors should release anonymized versions (if applicable).
        \item Providing as much information as possible in supplemental material (appended to the paper) is recommended, but including URLs to data and code is permitted.
    \end{itemize}

\item {\bf Experimental setting/details}
    \item[] Question: Does the paper specify all the training and test details (e.g., data splits, hyperparameters, how they were chosen, type of optimizer) necessary to understand the results?
    \item[] Answer: \answerYes{} 
    \item[] Justification: We completely specify all details in the appendix section about implementations. We also provide the code, hyperparameter configurations, and trained final weights in our GitHub repository.
    \item[] Guidelines:
    \begin{itemize}
        \item The answer \answerNA{} means that the paper does not include experiments.
        \item The experimental setting should be presented in the core of the paper to a level of detail that is necessary to appreciate the results and make sense of them.
        \item The full details can be provided either with the code, in appendix, or as supplemental material.
    \end{itemize}

\item {\bf Experiment statistical significance}
    \item[] Question: Does the paper report error bars suitably and correctly defined or other appropriate information about the statistical significance of the experiments?
    \item[] Answer: \answerYes{} 
    \item[] Justification: Our DSFT models are deterministically initialized so the only source of randomness comes from the dataset generation. We achieve our length-generalization results across all tasks on many different random seeds (details in the appendix).
    \item[] Guidelines:
    \begin{itemize}
        \item The answer \answerNA{} means that the paper does not include experiments.
        \item The authors should answer \answerYes{} if the results are accompanied by error bars, confidence intervals, or statistical significance tests, at least for the experiments that support the main claims of the paper.
        \item The factors of variability that the error bars are capturing should be clearly stated (for example, train/test split, initialization, random drawing of some parameter, or overall run with given experimental conditions).
        \item The method for calculating the error bars should be explained (closed form formula, call to a library function, bootstrap, etc.)
        \item The assumptions made should be given (e.g., Normally distributed errors).
        \item It should be clear whether the error bar is the standard deviation or the standard error of the mean.
        \item It is OK to report 1-sigma error bars, but one should state it. The authors should preferably report a 2-sigma error bar than state that they have a 96\% CI, if the hypothesis of Normality of errors is not verified.
        \item For asymmetric distributions, the authors should be careful not to show in tables or figures symmetric error bars that would yield results that are out of range (e.g., negative error rates).
        \item If error bars are reported in tables or plots, the authors should explain in the text how they were calculated and reference the corresponding figures or tables in the text.
    \end{itemize}

\item {\bf Experiments compute resources}
    \item[] Question: For each experiment, does the paper provide sufficient information on the computer resources (type of compute workers, memory, time of execution) needed to reproduce the experiments?
    \item[] Answer: \answerYes{} 
    \item[] Justification: As we mention in the appendix, we trained on single NVIDIA A100-SXM4-40GB and Tesla V100-SXM2-32GB GPUs.
    \item[] Guidelines:
    \begin{itemize}
        \item The answer \answerNA{} means that the paper does not include experiments.
        \item The paper should indicate the type of compute workers CPU or GPU, internal cluster, or cloud provider, including relevant memory and storage.
        \item The paper should provide the amount of compute required for each of the individual experimental runs as well as estimate the total compute. 
        \item The paper should disclose whether the full research project required more compute than the experiments reported in the paper (e.g., preliminary or failed experiments that didn't make it into the paper). 
    \end{itemize}
    
\item {\bf Code of ethics}
    \item[] Question: Does the research conducted in the paper conform, in every respect, with the NeurIPS Code of Ethics \url{https://neurips.cc/public/EthicsGuidelines}?
    \item[] Answer: \answerYes{} 
    \item[] Justification: We carefully reviewed the ethics guidelines and confirm that our research respects them.
    \item[] Guidelines:
    \begin{itemize}
        \item The answer \answerNA{} means that the authors have not reviewed the NeurIPS Code of Ethics.
        \item If the authors answer \answerNo, they should explain the special circumstances that require a deviation from the Code of Ethics.
        \item The authors should make sure to preserve anonymity (e.g., if there is a special consideration due to laws or regulations in their jurisdiction).
    \end{itemize}

\item {\bf Broader impacts}
    \item[] Question: Does the paper discuss both potential positive societal impacts and negative societal impacts of the work performed?
    \item[] Answer: \answerNA{} 
    \item[] Justification: This paper presents work whose goal is to advance the field of machine learning. There are many potential societal consequences of our work, none of which we feel must be specifically highlighted here.
    \item[] Guidelines:
    \begin{itemize}
        \item The answer \answerNA{} means that there is no societal impact of the work performed.
        \item If the authors answer \answerNA{} or \answerNo, they should explain why their work has no societal impact or why the paper does not address societal impact.
        \item Examples of negative societal impacts include potential malicious or unintended uses (e.g., disinformation, generating fake profiles, surveillance), fairness considerations (e.g., deployment of technologies that could make decisions that unfairly impact specific groups), privacy considerations, and security considerations.
        \item The conference expects that many papers will be foundational research and not tied to particular applications, let alone deployments. However, if there is a direct path to any negative applications, the authors should point it out. For example, it is legitimate to point out that an improvement in the quality of generative models could be used to generate Deepfakes for disinformation. On the other hand, it is not needed to point out that a generic algorithm for optimizing neural networks could enable people to train models that generate Deepfakes faster.
        \item The authors should consider possible harms that could arise when the technology is being used as intended and functioning correctly, harms that could arise when the technology is being used as intended but gives incorrect results, and harms following from (intentional or unintentional) misuse of the technology.
        \item If there are negative societal impacts, the authors could also discuss possible mitigation strategies (e.g., gated release of models, providing defenses in addition to attacks, mechanisms for monitoring misuse, mechanisms to monitor how a system learns from feedback over time, improving the efficiency and accessibility of ML).
    \end{itemize}
    
\item {\bf Safeguards}
    \item[] Question: Does the paper describe safeguards that have been put in place for responsible release of data or models that have a high risk for misuse (e.g., pre-trained language models, image generators, or scraped datasets)?
    \item[] Answer: \answerNA{} 
    \item[] Justification: We don't believe our research holds such risks.
    \item[] Guidelines:
    \begin{itemize}
        \item The answer \answerNA{} means that the paper poses no such risks.
        \item Released models that have a high risk for misuse or dual-use should be released with necessary safeguards to allow for controlled use of the model, for example by requiring that users adhere to usage guidelines or restrictions to access the model or implementing safety filters. 
        \item Datasets that have been scraped from the Internet could pose safety risks. The authors should describe how they avoided releasing unsafe images.
        \item We recognize that providing effective safeguards is challenging, and many papers do not require this, but we encourage authors to take this into account and make a best faith effort.
    \end{itemize}

\item {\bf Licenses for existing assets}
    \item[] Question: Are the creators or original owners of assets (e.g., code, data, models), used in the paper, properly credited and are the license and terms of use explicitly mentioned and properly respected?
    \item[] Answer: \answerYes{} 
    \item[] Justification: We, the authors, are the original owners of the assets used in the paper.
    \item[] Guidelines:
    \begin{itemize}
        \item The answer \answerNA{} means that the paper does not use existing assets.
        \item The authors should cite the original paper that produced the code package or dataset.
        \item The authors should state which version of the asset is used and, if possible, include a URL.
        \item The name of the license (e.g., CC-BY 4.0) should be included for each asset.
        \item For scraped data from a particular source (e.g., website), the copyright and terms of service of that source should be provided.
        \item If assets are released, the license, copyright information, and terms of use in the package should be provided. For popular datasets, \url{paperswithcode.com/datasets} has curated licenses for some datasets. Their licensing guide can help determine the license of a dataset.
        \item For existing datasets that are re-packaged, both the original license and the license of the derived asset (if it has changed) should be provided.
        \item If this information is not available online, the authors are encouraged to reach out to the asset's creators.
    \end{itemize}

\item {\bf New assets}
    \item[] Question: Are new assets introduced in the paper well documented and is the documentation provided alongside the assets?
    \item[] Answer: \answerYes{} 
    \item[] Justification: We believe that the codebase for our paper is well documented.
    \item[] Guidelines:
    \begin{itemize}
        \item The answer \answerNA{} means that the paper does not release new assets.
        \item Researchers should communicate the details of the dataset\slash code\slash model as part of their submissions via structured templates. This includes details about training, license, limitations, etc. 
        \item The paper should discuss whether and how consent was obtained from people whose asset is used.
        \item At submission time, remember to anonymize your assets (if applicable). You can either create an anonymized URL or include an anonymized zip file.
    \end{itemize}

\item {\bf Crowdsourcing and research with human subjects}
    \item[] Question: For crowdsourcing experiments and research with human subjects, does the paper include the full text of instructions given to participants and screenshots, if applicable, as well as details about compensation (if any)? 
    \item[] Answer: \answerNA{} 
    \item[] Justification: The paper does not involve crowdsourcing nor research with human subjects.
    \item[] Guidelines:
    \begin{itemize}
        \item The answer \answerNA{} means that the paper does not involve crowdsourcing nor research with human subjects.
        \item Including this information in the supplemental material is fine, but if the main contribution of the paper involves human subjects, then as much detail as possible should be included in the main paper. 
        \item According to the NeurIPS Code of Ethics, workers involved in data collection, curation, or other labor should be paid at least the minimum wage in the country of the data collector. 
    \end{itemize}

\item {\bf Institutional review board (IRB) approvals or equivalent for research with human subjects}
    \item[] Question: Does the paper describe potential risks incurred by study participants, whether such risks were disclosed to the subjects, and whether Institutional Review Board (IRB) approvals (or an equivalent approval/review based on the requirements of your country or institution) were obtained?
    \item[] Answer: \answerNA{} 
    \item[] Justification: The paper does not involve crowdsourcing nor research with human subjects.
    \item[] Guidelines:
    \begin{itemize}
        \item The answer \answerNA{} means that the paper does not involve crowdsourcing nor research with human subjects.
        \item Depending on the country in which research is conducted, IRB approval (or equivalent) may be required for any human subjects research. If you obtained IRB approval, you should clearly state this in the paper. 
        \item We recognize that the procedures for this may vary significantly between institutions and locations, and we expect authors to adhere to the NeurIPS Code of Ethics and the guidelines for their institution. 
        \item For initial submissions, do not include any information that would break anonymity (if applicable), such as the institution conducting the review.
    \end{itemize}

\item {\bf Declaration of LLM usage}
    \item[] Question: Does the paper describe the usage of LLMs if it is an important, original, or non-standard component of the core methods in this research? Note that if the LLM is used only for writing, editing, or formatting purposes and does \emph{not} impact the core methodology, scientific rigor, or originality of the research, declaration is not required.
    \item[] Answer: \answerYes{} 
    \item[] Justification: We used LLMs for code writing and for the generated controller equivalence proofs. The proofs have been carefully reviewed by the authors of the paper.
    \item[] Guidelines:
    \begin{itemize}
        \item The answer \answerNA{} means that the core method development in this research does not involve LLMs as any important, original, or non-standard components.
        \item Please refer to our LLM policy in the NeurIPS handbook for what should or should not be described.
    \end{itemize}

\end{enumerate}